%% file: main.tex
\documentclass{article}

\usepackage[main, final]{neurips_2026}

\usepackage[utf8]{inputenc} %
\usepackage[T1]{fontenc}    %
\usepackage{hyperref}
\hypersetup{
    colorlinks=true,
    linkcolor=linkindigo,  %
    citecolor=linkindigo,  %
    filecolor=linkindigo,  %
    urlcolor=linkindigo    %
}
\usepackage{url}            %
\usepackage{booktabs}       %
\usepackage{amsfonts}       %
\usepackage{nicefrac}       %
\usepackage{microtype}      %
\usepackage{xcolor}
\definecolor{linkindigo}{HTML}{515293} %

\usepackage{graphicx}
\usepackage{subcaption}
\usepackage{amsmath}
\usepackage{amssymb}
\usepackage{mathtools}
\usepackage{amsthm}
\usepackage{bm}
\usepackage[capitalize,noabbrev]{cleveref}
\usepackage[inline]{enumitem}
\usepackage{multirow}
\usepackage{array}
\usepackage{makecell}
\usepackage{xspace}
\usepackage[textsize=tiny]{todonotes}
\usepackage{wrapfig}
\usepackage{caption}
\usepackage{colortbl}
\usepackage{twemojis}
\definecolor{bondgray}{gray}{0.93}
\newcommand{\yes}{\twemoji{check mark button}}
\newcommand{\no}{\twemoji{cross mark}}

\newcommand{\oursacro}{\textsc{MotiFlow}\xspace}

\theoremstyle{plain}

\theoremstyle{definition}

\theoremstyle{remark}

\title{3D Molecule Generation from Rigid Motifs \\ via $\mathrm{SE}(3)$ Flows}

\author{
  \hspace*{-\tabcolsep}Roman Poletukhin$^{1,4}$\thanks{Correspondence to: \texttt{r.poletukhin26@imperial.ac.uk}.}\hspace{\dimexpr0.15em+7.08pt\relax}
  Marcel Kollovieh$^{1,2,3}$ \hspace{0.15em}
  Eike Eberhard$^{1,2,3}$ \hspace{0.15em}
  Stephan G\"{u}nnemann$^{1,2,3}$\hspace*{-\tabcolsep} \\[0.4em]
  $^{1}$Technical University of Munich (TUM), Germany \\
  $^{2}$Munich Data Science Institute (MDSI), Germany \\
  $^{3}$Munich Center for Machine Learning (MCML), Germany \\
  $^{4}$Imperial College London, UK}
\begin{document}
\maketitle
\begin{abstract}
Three-dimensional molecular structure generation is typically performed at the level of individual atoms, yet molecular graph generation techniques often consider \textit{fragments} as their structural units. Building on the advances in \textit{frame-based} protein structure generation, we extend these fragmentation ideas to 3D, treating general molecules as sets of \textit{rigid-body motifs}. Utilising this representation, we employ $\mathrm{SE}(3)$-equivariant generative modelling for \textit{de novo} 3D molecule generation from rigid motifs. In our evaluations, we observe results competitive with state-of-the-art bond-implicit methods, surpassing them in atom stability on \textsc{GEOM-Drugs}, while yielding a $2\times$ to $10\times$ reduction in generation steps and $3.5\times$ compression in molecular representations compared to standard atom-based methods. We release our code at \href{https://github.com/rpoletukhin/MotiFlow}{\texttt{github.com/rpoletukhin/MotiFlow}}.
\end{abstract}

\input{contents/intro}
\input{contents/background}
\input{contents/method}
\input{contents/related_work}
\input{contents/experiments}
\input{contents/conclusion}

\begin{ack}
This work was carried out as part of R.P.'s master's thesis at the Technical University of Munich. The authors received no specific third-party funding for this work and declare no competing interests.
\end{ack}

\bibliography{references}
\bibliographystyle{unsrtnat}

\appendix
\input{contents/appendix}

\end{document}

%% file: contents/intro.tex
\section{Introduction}
\label{Introduction}
\begin{wrapfigure}{r}{0.45\textwidth}
  \vspace{-2em}
  \centering
  \includegraphics[width=\linewidth]{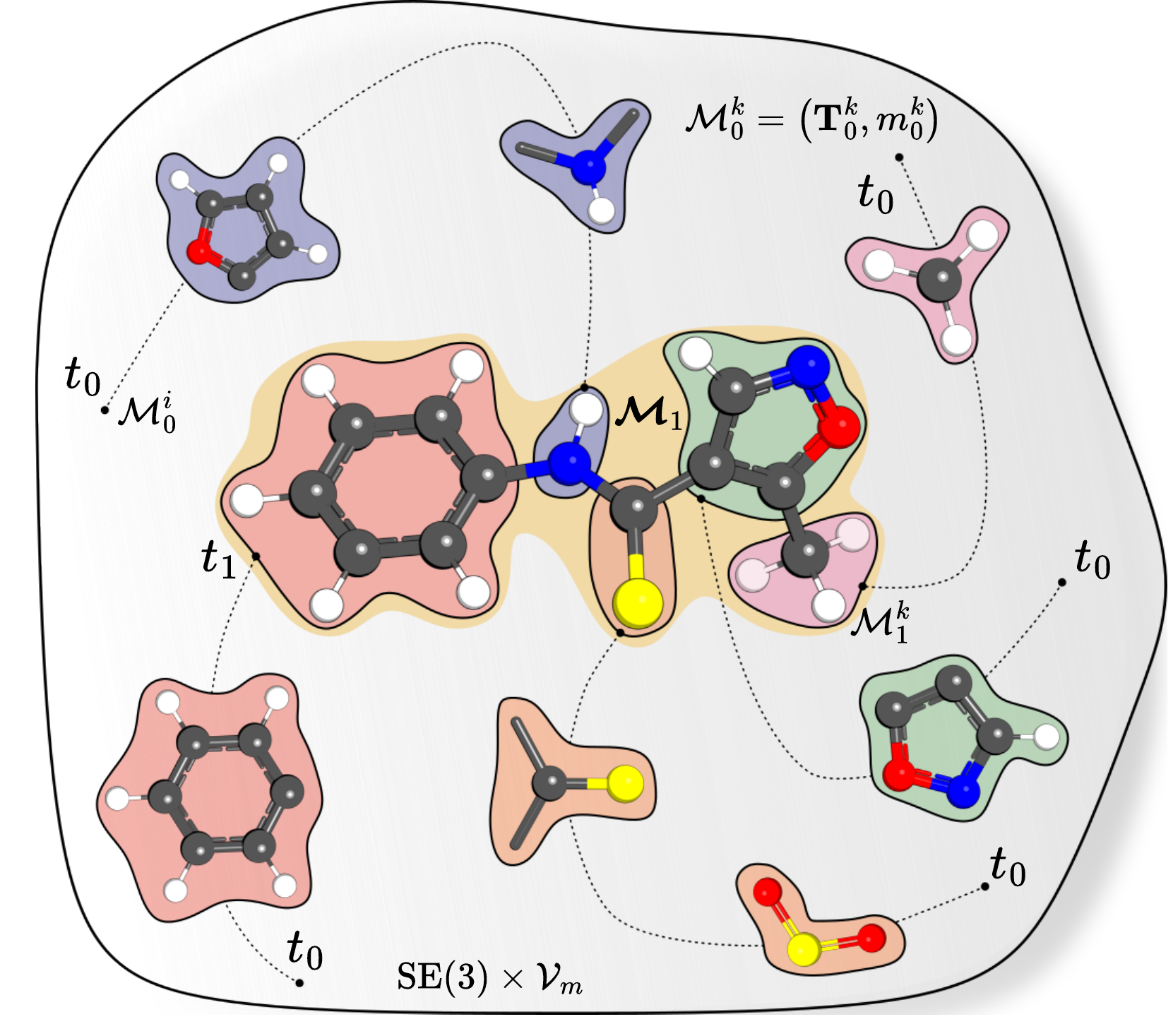}
  \caption{Molecule $\bm{\mathcal{M}}$ is generated from motifs $\mathcal{M}$ in a joint flow on the product space of rigid frames $\mathbf{T}$ and motif classes $m$. Atoms are recovered deterministically from motif frames and canonical poses.}
  \label{title_figure}
  \vspace{-0.5em}
\end{wrapfigure}
The generation of 3D molecular structures is a cornerstone of \textit{in silico} drug discovery and material design. Recent advances in deep learning have enabled the development of powerful generative models that treat molecules as point clouds of atoms, utilising $\mathrm{E(3)}$- and $\mathrm{SE(3)}$-equivariant diffusion-based frameworks to determine atomic coordinates \citep{edm, geoldm}. While these atom-based approaches achieve impressive performance, they operate at a low level of abstraction, discarding the rich chemical modularity inherent to molecular structures. In the realm of molecular graph generation, however, this hierarchical nature is widely exploited by fragment-based methods \citep{jtvae, magnet}, which assemble molecules from chemically meaningful motifs or scaffolds to capture high-level semantics and ensure validity. In the domain of 3D protein structure generation, such a modular perspective has also proven highly effective. Seminal works such as \textsc{AlphaFold2} \citep{alphafold2} and \textsc{FrameDiff} \citep{se3diffusion} demonstrated the efficacy of abstracting amino acid residues as \textit{rigid frames} in $\mathrm{SE(3)}$, decoupling backbone geometry from local side-chain details. However, lifting these ideas to general drug-like molecule generation in 3D presents significant challenges: unlike proteins, which are linear chains of a fixed set of residues, small molecules exhibit arbitrary branching, diverse topologies, and a vastly larger vocabulary of chemical motifs. 

In this work, we propose to bridge this gap by treating general molecules as collections of \textit{rigid motifs}. By decomposing molecules into chemically meaningful rigid fragments, we lift the generative task from the $\mathbb{R}^3$ atom space to the $\mathrm{SE(3)}$ manifold of fragments. We adopt a multimodal flow matching framework that jointly learns the discrete distribution of motif types and their continuous spatial configuration, illustrated in Figure \ref{title_figure}. This formulation enables us to generate high-fidelity molecular structures with significantly fewer sampling steps, leveraging chemically inspired representations that are substantially more concise than those of standard all-atom approaches.
\vspace{-3mm} 
\paragraph{Contributions} \looseness=-1 Our primary contributions are as follows. We propose \oursacro, a framework for 3D molecule generation that parametrises molecules as sets of rigid-body motifs in $\mathrm{SE}(3)$ rather than individual atoms. We develop a data-driven fragmentation and canonicalisation strategy that handles arbitrary molecular topologies and accounts for motif symmetries.
Further, we adapt the established multimodal flow \citep{discrete_flows} to the task of \textit{de novo} 3D structure generation of drug-like molecules. This formulation allows us to natively handle the joint generation of discrete fragment identities and their continuous geometric configurations without relying on autoregressive steps or learned decoding onto the atom-level structure. Finally, we empirically demonstrate that our method achieves superior performance compared to state-of-the-art bond-implicit atom-based baselines on medium- and large-sized molecules in \textsc{GEOM-Drugs} \citep{geom} and scales to larger molecules of \textsc{QMugs} \citep{qmugs}. \oursacro produces high-quality molecular structures with an order of magnitude fewer generation steps and excels at conditional generation tasks.

%% file: contents/background.tex
\section{Background}
\label{Background}
\subsection{\texorpdfstring{Flow Matching on $\mathrm{SE}(3)$ Manifold}{Flow Matching on SE(3) Manifold}}
\vspace{-1mm} 
\label{SE(3) Flow}
\paragraph{Geometric Preliminaries}
The group of rigid motions $\mathrm{SE}(3) \cong \mathbb{R}^3 \rtimes \mathrm{SO}(3)$ describes the configuration of rigid bodies in 3D space. While the translational component $\mathbb{R}^3$ is Euclidean (see Appendix \ref{Euclidean Flow}), the rotational one $\mathrm{SO}(3)$ is a \textit{compact Lie group} -- a smooth manifold equipped with a group structure. To define flows on this curved space, we rely on Riemannian geometry. At any point $\mathbf{R} \in \mathrm{SO}(3)$, the \textit{tangent space} $\mathcal{T}_{\mathbf{R}}\mathrm{SO}(3)$ is the vector space containing all possible velocity vectors passing through $\mathbf{R}$. We equip the manifold with the canonical bi-invariant \textit{Riemannian metric} defined by the inner product $\langle \mathbf{U}, \mathbf{V} \rangle_{\mathbf{R}} = \frac{1}{2}\mathrm{Tr}\left(\mathbf{U}^\top \mathbf{V}\right)$ for tangent vectors $\mathbf{U}, \mathbf{V} \in \mathcal{T}_{\mathbf{R}}\mathrm{SO}(3)$. This metric induces the norm $\|\mathbf{U}\|_{\mathrm{SO}(3)} = \sqrt{\langle \mathbf{U}, \mathbf{U} \rangle_{\mathbf{R}}}$. The mapping between the tangent space and the manifold is handled by the \textit{exponential map} $\mathbf{Q} = \exp_{\mathbf{R}}\left(\mathbf{V}\right)$, which projects a tangent vector $\mathbf{V}$ along a geodesic to a point $\mathbf{Q}$ on the manifold, and its inverse, the \textit{logarithmic map} $\mathbf{V} = \log_{\mathbf{R}}\left(\mathbf{Q}\right)$, which recovers the tangent vector, connecting $\mathbf{R}$ to $\mathbf{Q}$.
\vspace{-2mm} 
\paragraph{$\mathrm{SE}(3)$ Flow Matching} 
We decouple the generative process on $\mathrm{SE}(3)$ into independent translational and rotational components \citep{se3diffusion}.  
\looseness=-1 For the rotational component, we follow the \textsc{FoldFlow-Base} framework of \citet{foldflow}. This method extends \textit{Riemannian flow matching} \citep{rfm} (see Appendix \ref{Riemannian Flow}) by deriving a closed-form expression for the ground-truth conditional vector field, thereby significantly improving training speed and stability. We define a probability path $\mathbf{R}_t$ that interpolates along the geodesic connecting a sample from the uniform prior $\mathbf{R}_0 \sim \mathcal{U}_{\mathrm{SO}(3)}$ to a data sample $\mathbf{R}_1$. The conditional vector field $u^{\mathbf{R}}_t\left(\mathbf{R}_t \mid \mathbf{R}_1\right) \in \mathcal{T}_{\mathbf{R}_t}\mathrm{SO}(3)$ generating this path is:
\begin{equation}\label{eq:rot_vf}
    u_t^{\mathbf{R}}\left(\mathbf{R}_t \mid \mathbf{R}_1\right) = \frac{1}{1-t} \log_{\mathbf{R}_t}(\mathbf{R}_1).
\end{equation}
Computing $\log_{\mathbf{R}_t}(\mathbf{R}_1)$ in \eqref{eq:rot_vf} naively requires evaluating an infinite matrix power series, which is computationally expensive and numerically unstable. To circumvent this, \citet{foldflow} exploit the Lie group structure. Instead of computing the logarithm directly at $\mathbf{R}_t$, they first compute the relative rotation $\mathbf{R}_{\text{rel}} = \mathbf{R}_t^\top \mathbf{R}_1$. This maps the problem to the identity-tangent space, the Lie algebra $\mathfrak{so}(3)$, where the logarithm admits a fast, closed-form solution via Rodrigues' formula. The resulting vector is then transported back to the tangent space at $\mathbf{R}_t$ via left-multiplication.

To train the model, one samples intermediate noisy frames $\mathbf{T}_t = \left(\mathbf{R}_t, \mathbf{x}_t\right)$. The rotation evolves according to the geodesic formula $\mathbf{R}_t = \exp_{\mathbf{R}_0}\left(t \log_{\mathbf{R}_0}\left(\mathbf{R}_1\right)\right)$. For translation, the conditional probability path $p_t\left(\mathbf{x}_t \mid \mathbf{x}_0, \mathbf{x}_1\right)$ is defined by the linear interpolation $\mathbf{x}_t = (1-t)\mathbf{x}_0 + t\mathbf{x}_1$, which corresponds to the constant conditional vector field $u_t^\mathbf{x}\left(\mathbf{x}_t \mid \mathbf{x}_0, \mathbf{x}_1\right) = \mathbf{x}_1 - \mathbf{x}_0$ on $\mathbb{R}^3$.

The final objective is to regress the neural vector field $v_\theta = \left(v_\theta^{\mathbf{x}}, v_\theta^{\mathbf{R}}\right)$ to these target fields. Assuming an independent coupling between the prior $p_0$ and the data $p^*$, the loss is:
\begin{equation}\label{eq:loss_se3}
\resizebox{0.94\columnwidth}{!}{$\displaystyle
    \mathcal{L}_{\mathrm{SE}(3)}(\theta)
    = \mathbb{E}_{\substack{\mathbf{T}_0 \sim p_0, \mathbf{T}_1 \sim p^* \\ t \sim \mathcal{U}(0,1)}} \Big[\,
        \| v_\theta^{\mathbf{x}}\left(\mathbf{T}_t, t\right) - u_t^\mathbf{x}\left(\mathbf{x}_t \mid \mathbf{x}_0, \mathbf{x}_1\right) \|^2 +\; \| v_\theta^{\mathbf{R}}\left(\mathbf{T}_t, t\right) - u_t^{\mathbf{R}}\left(\mathbf{R}_t \mid \mathbf{R}_1\right) \|_{\mathrm{SO}(3)}^2
    \Big].
$}
\end{equation}
\looseness=-1 At inference, one samples an initial frame $\mathbf{T}_0 = \left(\mathbf{R}_0, \mathbf{x}_0\right)$ from the prior $p_0 = \mathcal{U}_{\mathrm{SO}(3)} \times \mathcal{N}(\mathbf{0}, \mathbf{I})$ and integrates the learned joint vector field $v_\theta$ to generate the final rigid frame $\mathbf{T}_1 = \left(\mathbf{R}_1, \mathbf{x}_1\right)$.
\subsection{Discrete Flows}
\label{Discrete Flows}
While continuous approaches to flow matching on the categorical simplex exist (e.g., \citealp{catflow, fisher_fm}), we follow \citet{discrete_flows} and handle generative modelling of discrete data with \textit{continuous-time Markov chains} (CTMCs; see Appendix \ref{app:discrete_flows}). Let $m$ denote a discrete random variable taking values in a finite vocabulary $\mathcal{V}_m$. Analogous to the continuous case, we aim to transform a sample $m_0$ from a tractable prior distribution $p_0$, e.g., a uniform distribution or a $[\text{MASK}]$ state, to a data sample $m_1 \sim p^*$ via a probability path $p_t$ for $t \in [0, 1]$.
\vspace{-2mm} 
\paragraph{CTMC Dynamics}
The evolution of the marginal probability mass function $p_t$ over $\mathcal{V}_m$ is governed by the \textit{Kolmogorov forward equation}. For a time-dependent transition \textit{rate matrix} $\mathbf{Q}_t \in \mathbb{R}^{|\mathcal{V}_m| \times |\mathcal{V}_m|}$:
\begin{equation}\label{eq:kolmogorov}
    \partial_t p_t(k) = \sum_{j \neq k} p_t(j) \mathbf{Q}_t(j, k) - p_t(k) \sum_{j \neq k} \mathbf{Q}_t(k, j),
\end{equation}
where $\mathbf{Q}_t(j, k) \geq 0$ for $j \neq k$ represents the instantaneous rate of jumping from state $j$ to state $k$. This linear system in \eqref{eq:kolmogorov} serves as the discrete analogue to the continuity equation in continuous flow matching, with $\mathbf{Q}_t$ playing the role of the vector field $u_t$.
\vspace{-2mm} 
\paragraph{Discrete Flow Matching}
Constructing a generative model requires finding a rate matrix $\mathbf{Q}_t$ that generates a desired probability path $p_t$. Following the conditional flow matching paradigm, we define the marginal path as an expectation over conditional paths $p_t(m_t \mid m_1)$ anchored at the sample $m_1$: \(p_t(m_t) = \mathbb{E}_{m_1 \sim p^*}\left[p_t(m_t \mid m_1)\right]\).
The conditional flow $p_t(\cdot \vert m_1)$ is typically chosen as a linear interpolation in probability space, such that $p_0(\cdot \mid m_1) = p_0(\cdot)$ and $p_1(\cdot \mid m_1) = \delta_{m_1}(\cdot)$, where $\delta$ is the Dirac delta. \citet{discrete_flows} demonstrate that the marginal rate matrix $\mathbf{Q}_t$ can be realised as the expectation of a \textit{conditional rate matrix} $\mathbf{Q}_t(\cdot \mid m_1)$ which generates $p_t(\cdot \mid m_1)$:
\begin{equation}\label{eq:marginal_rate}
    \mathbf{Q}_t(j, k) = \mathbb{E}_{m_1 \sim p(m_1 \mid m_t=j)} \left[ \mathbf{Q}_t(j, k \mid m_1) \right].
\end{equation}
Here, $\mathbf{Q}_t(j, k \mid m_1)$ in \eqref{eq:marginal_rate} is a closed-form rate matrix derived analytically to satisfy the conditional Kolmogorov equation for the chosen interpolant. Unlike the continuous case, where we regress a vector field directly, learning the marginal rate matrix requires access to the posterior $p(m_1 \mid m_t=j)$. Consequently, we parametrise a denoising neural network $p_\theta(m_1 \mid m_t)$ to approximate the clean data distribution given a noisy state. The training objective is thus the cross-entropy loss:
\begin{equation}\label{eq:loss_dfm}
    \mathcal{L}_{\text{DFM}}(\theta) = \mathbb{E}_{\substack{m_1 \sim p^*, \, m_t \sim p_t(\cdot \mid m_1) \\ t \sim \mathcal{U}(0, 1)}} \left[ - \log p_\theta(m_1 \mid m_t) \right].
\end{equation}
At inference time, we construct the generative rate matrix $\mathbf{Q}_t^\theta$ using the learned denoiser: $\mathbf{Q}_t^\theta(j, k) = \mathbb{E}_{m_1 \sim p_\theta(\cdot \mid m_t=j)} [ \mathbf{Q}_t(j, k \mid m_1) ]$, sample $m_0 \sim p_0$ and simulate the CTMC trajectory defined by $\mathbf{Q}_t^\theta$.

%% file: contents/method.tex
\section{Method} \label{Method}
In this section, we start with a common representation of a molecule  $\{(\mathbf{y}_j, h_j)\}_{j=1}^{N}$ as a point cloud of $N$ atoms, each with coordinates $\mathbf{y}_j \in \mathbb{R}^3$ and the atom type $h_j \in \mathcal{V}_{a}$. Here, $\mathcal{V}_{a}$ is the vocabulary of all atom types, including ions. 

\looseness=-1 In Sections \ref{Rigid-Motif Decomposition} and \ref{Canonicalisation and Vocabulary}, we describe the \textit{frame fragmentation}; its purpose is to reparametrise the molecule as $K$ rigid motifs $\{\mathcal{M}_i\}_{i=1}^K = \{(\mathbf{T}_i, m_i)\}_{i=1}^K$ where a frame $\mathbf{T}_i = \left(\mathbf{R}_i, \mathbf{x}_i\right) \in \mathrm{SE}(3)$ has a rotation matrix $\mathbf{R}_i \in \mathrm{SO}(3)$ and a translation vector $\mathbf{x}_i \in \mathbb{R}^3$ from the origin to the geometric centre of the motif. Each frame's rotation is defined relative to this motif's \textit{exemplar fragment} $m_i \in \mathcal{V}_m$ from the \textit{motif vocabulary} $\mathcal{V}_m$. 

A fragment $m_i$ with $N_i$ atoms in this vocabulary describes the atom-level structure of the rigid motif, and is thus a tuple $m_i = (\mathbf{P}_i, \mathbf{h}_i, \mathcal{S}_i)$. Here, the fixed set of 3D coordinates of intra-fragment atoms $\mathbf{P}_i \in \mathbb{R}^{N_i \times 3}$ defines the motif's \textit{canonical pose} centred at the origin. The corresponding types of intra-fragment atoms are denoted by $\mathbf{h}_i \in \mathcal{V}_a^{N_i}$. The third element, $\mathcal{S}_i \subset \mathrm{SO}(3)$, represents the \textit{discrete symmetry group of the motif}, i.e., the set of all rotations that, if applied to the pose $\mathbf{P}_i$, result in an indistinguishable molecular motif in 3D space.

We note that such frame-based representation is \textit{invertible}: to recover the atom-level representation $\mathbf{Y}_i \in \mathbb{R}^{N_i \times 3}$ of atoms that corresponds to the rigid fragment $m_i$ with the frame $\mathbf{T}_i = \left(\mathbf{R}_i, \mathbf{x}_i\right)$ in a molecule in 3D space, one has to apply the rigid transformation to the motif's canonical pose:
\begin{equation}\label{eq:reconstruction}
\mathbf{Y}_i = \mathbf{P}_i \mathbf{R}_i + \mathbf{1}_{N_i}\mathbf{x}_i^\top \equiv \mathbf{P}_i \mathbf{S} \mathbf{R}_i + \mathbf{1}_{N_i}\mathbf{x}_i^\top, \,\, \forall \mathbf{S} \in \mathcal{S}_i.
\end{equation}
Under this rigid-frame parametrisation with the deterministic lift in \eqref{eq:reconstruction}, generating \textit{de novo} molecules with $K$ fragments can be seen as a task of sampling from the distribution on  $\mathrm{SE}(3)^K \times \mathcal{V}_m^K$.  In Section \ref{Multimodal Flow}, we formulate our generative framework \oursacro.
\subsection{Rigid-Motif Decomposition} \label{Rigid-Motif Decomposition}
To establish a motif vocabulary $\mathcal{V}_m$, we need to define a fragmentation scheme. Such a scheme has to satisfy several requirements, namely,
\begin{enumerate*}[label=(\roman*)]
  \item \textit{rigidity}: fragments must be structurally rigid approximations  (i.e., lacking internal rotatable bonds) to accurately represent all instances of a motif throughout the data,
  \item \textit{non-degeneracy}: each fragment must possess at least three non-collinear points to define a frame $\mathbf{T} \in \mathrm{SE}(3)$ and
  \item \textit{tractability}, meaning that the frequency of each distinct class of $\mathcal{V}_m$ in the data has to be sufficiently high for learning the distribution over $\mathcal{V}_m^K$.
\end{enumerate*}

The fragmentation comprises two stages. We begin by identifying all sufficiently rigid structures: we preserve double and triple bonds, as well as all bonds within approximately \textit{planar} rings and fused ring systems. Further, while we do cut the bonds between heavy atoms that are acyclic but adjacent to a cycle, unlike common fragmentation techniques \citep{hiervae, moler}, we do not cut bonds to hydrogens. This helps introduce fewer ill-defined rigid frames at the expense of having more unique motifs. After the chosen bonds are cut and the preliminary set of rigid motifs is established, we proceed by \textit{pruning} them: analogous to common methods in molecular graph decomposition (e.g., \citealp{hiervae}), we further fragment rigid motifs whose total number of occurrences across the dataset is less than $\alpha$ of the dataset size. We ablate different values of the hyperparameter $\alpha$ and its influence on the generation in Section \ref{Ablations}.  By default, we use $\alpha=0.1\%$.

At this point, the only ill-defined frames are those that are collinear motifs (e.g., alkynes) or isolated atoms (e.g., $\text{Cl}$ atom cut from a $\text{C}_6\text{H}_5\text{Cl}$ chlorobenzene ring). To be able to uniquely define their rotation matrix $\mathbf{R} \in \mathrm{SO}(3)$, similar to \citet{sigmadock}, we start adding dummy atoms placed at a unit distance to the nearest non-collinear neighbours of this fragment in the original molecule until the orientation of the frame is locked and the rigid body is well-defined.
\vspace{-1mm} 
\subsection{Canonicalisation and Vocabulary} \label{Canonicalisation and Vocabulary}
\vspace{-1mm} 
\begin{wrapfigure}[14]{r}{0.45\textwidth}
  \vspace{\dimexpr-12mm-2pt\relax}
  \centering
  \setlength{\abovecaptionskip}{5.65pt}
  \includegraphics[width=\linewidth]{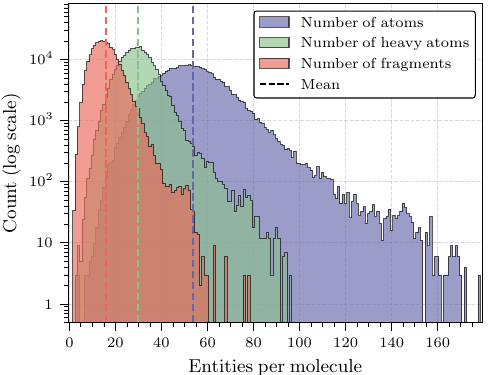}
  \caption{Comparison of molecular representations on \textsc{QMugs} \citep{qmugs}. Fragmentation is reported for $\alpha = 0.1\%$.}
  \label{qmugs_distr}
  \vspace{-3mm}
\end{wrapfigure}

Once the rigid motifs are defined, we proceed with constructing a vocabulary $\mathcal{V}_m$ where each element is a canonical descriptor of a chemical motif, formally defined as a tuple $m_i = (\mathbf{P}_i, \mathbf{h}_i, \mathcal{S}_i)$. For a given motif class $i$, the canonical fragment with its centred pose $\mathbf{P}_i \in \mathbb{R}^{N_i \times 3}$ and atom types $\mathbf{h}_i \in \mathcal{V}_a^{N_i}$ is chosen arbitrarily as the rigid motif's first occurrence during the dataset preprocessing and fixed. We compute $\mathcal{S}_i$ by identifying the set of all graph automorphisms $\Pi_i$, i.e., node permutations that preserve chemical connectivity and element types in the motif. For each automorphism $\pi \in \Pi_i$, we derive the corresponding rotation matrix $\mathbf{R}^{\pi}_i$ that maps the exemplar onto its permuted self, i.e., $\mathbf{P}_{i} \approx \pi(\mathbf{P}_{i}) \mathbf{R}^{\pi}_i$. This explicitly encodes the rotational invariance of symmetric motifs, e.g., the indistinguishable orientations of cyclopropane $(\text{CH}_2)_3$, into the vocabulary.

We further assign a ground truth pose $\mathbf{T}_j =\left(\mathbf{R}_j, \mathbf{x}_j\right) \in \mathrm{SE}(3)$ to each fragment instance $\mathcal{M}_j$ of type $m_i$ found in the data. The translation vector $\mathbf{x}_j$ is defined as the vector from the origin to the geometric centre of $\mathcal{M}_j$. We then compute a representative rotation $\mathbf{R}_j$ via the Kabsch algorithm \citep{kabsch} given any valid automorphism $\pi \in \Pi_i$ establishing the atom-wise correspondence:
{
\setlength{\abovedisplayskip}{3pt}
\setlength{\belowdisplayskip}{2pt}
\begin{equation}\label{eq:kabsch}
    \mathbf{R}_j = \mathop{\mathrm{argmin}}_{\mathbf{R} \in \mathrm{SO}(3)} \sum_{a=1}^{N_i} \left\| \mathbf{P}_{i, a} \mathbf{R} -\mathbf{Y}^{\pi}_{j, a} \right\|^2,
\end{equation}
}%
where $\mathbf{P}_{i,a}$ in \eqref{eq:kabsch} denotes the position of the $a$-th atom in $m_i$.
This method allows one to represent an arbitrary non-collinear molecule as a collection of well-defined rigid motifs from a tractable vocabulary $\mathcal{V}_m$. Figure \ref{qmugs_distr} shows that for \textsc{QMugs}, such a fragment-based molecular parametrisation compresses the common all-atom and heavy-atom representations by factors of $3.4$ and $1.9$, respectively, without resorting to a learnable latent space or a computationally expensive reconstruction.

\subsection{\texorpdfstring{Multimodal Flow on $\mathrm{SE}(3)^K \times \mathcal{V}_m^K$}{Multimodal Flow on SE(3)K x VmK}}
\label{Multimodal Flow}
Given the larger size of the motif vocabulary $\mathcal{V}_m$ compared to the common atom vocabulary $\mathcal{V}_a$, we opt not to approximate the discrete fragment classes in a one-hot continuous fashion, as commonly done in the literature \citep{edm, end}. Instead, we propose to handle the discrete and continuous supports of $m \in \mathcal{V}_m$ and $\mathbf{T} \in \mathrm{SE}(3)$, respectively, in a \textit{multimodal flow} \citep{discrete_flows}. For a molecule with $K$ rigid motifs $\bm{\mathcal{M}} = \{\mathcal{M}^k\}_{k=1}^K$, \oursacro models the following conditional flow\footnote{For brevity, $p_{t}(\cdot \mid \cdot)$ refers to probability density and probability mass functions for continuous and discrete variables.}, factorised over modalities and individual rigid-motif frames:
{
\begin{equation}\label{eq:factorisation}
\setlength{\abovedisplayskip}{3pt}
\setlength{\belowdisplayskip}{3pt}
p_{t}(\bm{\mathcal{M}}_t  \mid \bm{\mathcal{M}}_1) := \prod_{k=1}^{K} \underbrace{p_{t}(m_t^k  \mid m_1^k)}_{\text{Discrete Flow}} \underbrace{p_{t}(\mathbf{T}_t^k  \mid \mathbf{T}_1^k)}_{\text{SE}(3) \text{ Flow}}.
\end{equation}
}%
This factorisation in \eqref{eq:factorisation} implies that the generative process for each rigid motif is independent \textit{conditional} on the sample $\bm{\mathcal{M}}_1$, allowing us to train the joint model with a sum of modality-specific losses. We note that this factorisation is a modelling choice defining the conditional probability path, following the multimodal formulation of \citet{discrete_flows}: all vector fields and categorical predictions are parametrised by a shared network conditioned on the complete noisy molecule $\bm{\mathcal{M}}_t$, which captures dependencies across motifs, their identities, and geometry (see \textit{Architecture} below).
\vspace{-2mm}
\paragraph{Continuous Dynamics} 
For the geometric component, we independently apply the $\mathrm{SE}(3)$ flow matching framework described in Section \ref{SE(3) Flow} to each of the $K$ rigid frames. The conditional probability path $p_t(\mathbf{T}_t^k \mid \mathbf{T}_1^k)$ is constructed via the product of the Euclidean interpolant for translations and the geodesic one for rotations. The training objective $\mathcal{L}_{\mathrm{SE}(3)}$ in \eqref{eq:loss_se3} is the average over $K$ motifs of the regression loss between the network outputs and the target vector fields $u_t^\mathbf{x}$ and $u_t^\mathbf{R}$.
\vspace{-2mm}
\paragraph{Discrete Dynamics} 
For the motif types, we adopt the discrete flow from Section \ref{Discrete Flows} using a \textit{masking} prior. Specifically, we set the prior $p_0$ to be a Dirac delta on a special token, $m_0^k = [\text{MASK}]$ for all $k$. The conditional probability path interpolates linearly between this mask state and the true motif type $m_1^k$: \(p_t(m_t^k \mid m_1^k) = (1-t)\delta_{[\text{MASK}]}(m_t^k) + t\delta_{m_1^k}(m_t^k)\).
To generate this path via a CTMC, we require the conditional rate matrix $\mathbf{Q}_t(j, l \mid m_1^k)$. For the masking interpolant, this matrix takes a simple analytic form where probability mass is transferred solely from $[\text{MASK}]$ to the target class $m_1^k$ \citep{discrete_flows}: \(\mathbf{Q}_t(j, l \mid m_1^k) = \mathbb{I}(j = [\text{MASK}]) \cdot \mathbb{I}(l = m_1^k) \cdot \frac{1}{1-t}\).
Intuitively, at any time $t \in (0, 1)$, a masked motif has a rate of $(1-t)^{-1}$ to unmask to its true value $m_1^k$, while unmasked motifs remain fixed. To train the model, we employ a denoising network $p_\theta(m_1^k \mid \bm{\mathcal{M}}_t)$ that predicts the categorical distribution over $\mathcal{V}_m$ given the noisy state of the entire molecule. The discrete objective $\mathcal{L}_{\text{DFM}}$ in \eqref{eq:loss_dfm} is the cross-entropy between the predicted logits and the true motif type $m_1^k$. Denoting by $\mathcal{L}_{\text{trans}}(t)$, $\mathcal{L}_{\text{rot}}(t)$ and $\mathcal{L}_{\text{DFM}}(t)$ the translational and rotational terms of \eqref{eq:loss_se3} and the cross-entropy in \eqref{eq:loss_dfm} at time $t$, the full objective is 
\begin{equation}\label{eq:loss_total}
    \mathcal{L}(\theta) = \mathbb{E}_{t \sim \mathcal{U}(0, 1)}\left[\mathcal{L}_{\text{trans}}(t) + \mathcal{L}_{\text{rot}}(t) + \lambda_{\text{cat}}(t)\,\mathcal{L}_{\text{DFM}}(t)\right], \quad \lambda_{\text{cat}}(t) = \frac{1}{1-t}.
\end{equation}
\vspace{-3mm}
\paragraph{Architecture}
\looseness=-1 We parametrise the time-dependent vector fields and the discrete denoiser using a single unified neural network $v_\theta(\bm{\mathcal{M}}_t, t)$. Our architecture builds upon the \textsc{FoldFlow-Base} backbone \citep{foldflow}, which utilises invariant point attention (IPA) \citep{alphafold2} to process 3D rigid frames (see Appendix \ref{app:scalability} for a discussion of its scalability). The network takes as input the noisy frames $\{\mathbf{T}_t^k\}_{k=1}^K$ and the embeddings of the partially masked motif tokens $\{m_t^k\}_{k=1}^K$. We introduce three modifications to adapt this architecture for multimodal molecular generation. First, we incorporate \textit{self-conditioning} \citep{self_cond, harm_self_cond} for the discrete modality: during training, with a probability $0.5$, we feed the model's own estimated clean motif types $\hat{m}_1$ back as input, improving coherence between the geometric and semantic features. Second, we augment the network's layers with triangle multiplicative updates \citep{alphafold2} on the pair of fragment representations to better capture the geometric constraints between rigid motifs, which, unlike residues in the protein backbones, are ordered arbitrarily. Finally, we add a prediction head that outputs logits over $\mathcal{V}_m$ for the discrete flow.

\textbf{Symmetries \,}
This generative process has two kinds of physical symmetries. The first one, global $\mathrm{SE}(3)$ equivariance, is guaranteed by the IPA backbone, ensuring that if the entire molecule is rotated and translated, the generated vector fields rotate and translate accordingly. Additionally, individual motifs $m_i$ may also possess non-trivial finite groups of rotational symmetries $\mathcal{S}_i$, and the true distribution is invariant to the choice of canonical pose representation from its orbit $\{\mathbf{P}_i \mathbf{S}_1, \dots, \mathbf{P}_i \mathbf{S}_{|\mathcal{S}_i|}\}$. We set $\mathcal{S}_{[\text{MASK}]} \coloneqq \{\mathbf{I}\}$. To handle this, we adopt a \textsc{GeoDiff}-style alignment strategy \citep{geodiff} in the loss. Instead of regressing towards a fixed canonical frame, we dynamically select the target rotation $\tilde{\mathbf{R}}_1^k$ from the symmetry orbit that is closest to the current noisy frame $\mathbf{R}_t^k$:
\begin{equation}\label{eq:sym_align}
    \tilde{\mathbf{R}}_1^k = \mathbf{S}^*\mathbf{R}_1^k, \quad \text{where} \quad \mathbf{S}^* = \mathop{\mathrm{argmax}}_{\mathbf{S} \in \mathcal{S}_k} \mathrm{Tr}\left((\mathbf{R}_t^k)^\top \mathbf{S}\mathbf{R}_1^k\right).
\end{equation}
\looseness=-1 Minimising the geodesic distance on $\mathrm{SO}(3)$ in \eqref{eq:sym_align} corresponds to maximising the trace of the relative rotation matrix, which is computationally negligible as the finite symmetry groups of motifs are small. Empirically, we found that this alignment stabilises training at small times $t$, where the noisy state is close to the uninformed prior, preventing the flow from receiving conflicting gradients from equivalent but spatially distant symmetric targets. We ablate the effects of the design choices and compare them with alternatives in Appendix \ref{Architecture Ablations}. Implementation details are in Appendix \ref{Implementation Details}.

%% file: contents/related_work.tex
\section{Related Work}
\label{Related Work}

\begin{wraptable}{r}{0.56\textwidth} 
    \vspace{-5mm} %
    \centering
    \footnotesize 
    \setlength{\tabcolsep}{2.2pt}
    \renewcommand{\arraystretch}{0.7}
    \caption{Results of unconditional generation on \textsc{QM9}.}
    \label{qm9_uncond}
    \begin{tabular}{@{} l c c c c c @{}}
    \toprule
    \multirow{2}{*}{Method} & \multirow{2}{*}{Steps} & \multicolumn{2}{c}{Stability ($\uparrow$)} & \multicolumn{2}{c}{Val. / Uniq. ($\uparrow$)} \\ 
    \cmidrule(lr){3-4} \cmidrule(l){5-6}
     & & A, \% & M, \% & V, \% & V $\times$ U, \% \\
    \midrule
    \makecell[l]{Training Data} & -- & $99.0$ & $95.2$ & $97.7$ & $97.7$ \\
    \midrule
    \midrule
    
    \makecell[l]{\textsc{EDM} \\ \scriptsize\citeauthor{edm} [\scriptsize\citeyear{edm}]}
      & $10^3$ & $98.7$ & $82.0$ & $91.9$ & $90.7$ \\
    
    \makecell[l]{\textsc{EDM-Bridge} \\ \scriptsize\citet{edm_bridge}}
      & $10^3$ & $98.8$ & $84.6$ & $92.0$ & $90.7$ \\
    
    \makecell[l]{\textsc{GeoLDM} \\ \scriptsize\citet{geoldm}}
      & $10^3$ & $98.9_{\text{\tiny $\pm$.1}}$ & $89.4_{\text{\tiny $\pm$.5}}$ & $93.8_{\text{\tiny $\pm$.4}}$ & $92.7_{\text{\tiny $\pm$.5}}$ \\
    
    \makecell[l]{\textsc{EquiFM} \\ \scriptsize\citet{equifm}}
      & $-$ & 
      $98.9_{\text{\tiny $\pm$.1}}$ & $88.3_{\text{\tiny $\pm$.3}}$ & 
      $94.7_{\text{\tiny $\pm$.4}}$ & $93.5_{\text{\tiny $\pm$.3}}$ \\
    
    \makecell[l]{\textsc{END} \\ \scriptsize\citet{end}}
      & $10^3$ & $98.9_{\text{\tiny $\pm$.0}}$ & $89.1_{\text{\tiny $\pm$.1}}$ & $94.8_{\text{\tiny $\pm$.1}}$ & $92.6_{\text{\tiny $\pm$.2}}$ \\
    
    \makecell[l]{\textsc{EDM*} \\ \scriptsize\citet{end}}
    & $10^3$ & $98.4_{\text{\tiny $\pm$.0}}$ & $85.3_{\text{\tiny $\pm$.3}}$ & $93.5_{\text{\tiny $\pm$.1}}$ & $91.9_{\text{\tiny $\pm$.1}}$ \\
    
    \makecell[l]{\textsc{GeoBFN} \\ \scriptsize\citet{geobfn}}
    & $10^3$ & $99.1_{\text{\tiny $\pm$.1}}$ & $90.9_{\text{\tiny $\pm$.2}}$ & $95.3_{\text{\tiny $\pm$.1}}$ & $93.0_{\text{\tiny $\pm$.1}}$ \\
    
    \makecell[l]{\textsc{VEDA-E} \\ \scriptsize\citet{veda}}
    & $50$ & $\mathbf{99.4}$ & $\mathbf{93.7}$ & $\mathbf{98.1}$ & $\mathbf{97.9}$ \\
    \midrule
    
    \makecell[l]{\oursacro}
    & $10^2$  & $99.1_{\text{\tiny $\pm$.1}}$ & $92.6_{\text{\tiny $\pm$.5}}$ & $95.3_{\text{\tiny $\pm$.6}}$ & $86.3_{\text{\tiny $\pm$.9}}$ \\
    \bottomrule
    \end{tabular}
    \vspace{-3mm} %
\end{wraptable}
\paragraph{Fragment-Based Molecule Generation} \looseness=-1 In the task of molecular graph generation, several approaches to tokenising molecules into motifs exist, which can be classified into chemically inspired and data-driven ones. The former group \citep{hiervae, moler, jtvae, frag_fm} focuses on top-down separation of acyclic and cyclic parts in the molecule, while the latter group \citep{psvae, micam} adopts the bottom-up merging strategy starting from individual atoms. As an alternative to motif-based methods, \citet{magnet} proposed \textit{scaffold-based} fragmentation, in which the molecular assembly starts from basic geometric shapes and attributes the chemical properties in the subsequent stages of generation. To the best of our knowledge, neither motif-based nor scaffold-based fragmentation has been widely explored in 3D. A notable exception is \textsc{HierDiff} \citep{hierdiff}, which utilises a hierarchical diffusion framework to position coarse fragment nodes in 3D space. In contrast to our approach, which generates the full $\mathrm{SE}(3)$ configuration of rigid motifs without auxiliary mechanisms, \textsc{HierDiff} relies on a learnable decoding procedure to resolve the specific identities and geometries of the fragments, thereby taking a different route. We contrast \oursacro with further fragment-based approaches in Appendix \ref{app:comparison}.
\vspace{-3mm} 
\paragraph{3D Molecule Generation} \looseness=-1 Generation of spatial molecular structure is primarily tackled with diffusion- and flow-based \citep{enflows, enlinker, flowmol} and autoregressive \citep{autoreg1, autoreg2, quetzal} models on atom coordinates. A notable exception to these directions is the work of \citet{voxel}, where 3D molecules are represented as atomic densities on regular grids. Existing approaches to 3D molecular generation can also be roughly classified into two categories based on how they treat bonds. The seminal work of \citet{edm} uses a lookup table to infer bond types from pairwise atom distances, which is the approach we adopt. Recently, a line of work \citep{codesign, midi, semlaflow, moldiff} has proposed a joint modelling of the 2D molecular graph topology and 3D atom coordinates. While they improve upon point cloud approaches, we perform our experiments without modelling the graph structure beyond intra-motif connectivity, i.e., the bond structure within the fragments; therefore, we do not directly compare our method to these approaches, but list published results of bond-explicit methods in Tables \ref{qm9_uncond_extended} and \ref{geom_uncond_extended} for reference.
\vspace{-5mm} 
\paragraph{Rigid-body Generation} Parametrising protein residues as rigid frames has been originally introduced in the seminal work of \citet{alphafold2} and received widespread adoption in the subsequent methods for protein structure prediction and design \citep{rfdiffusion, se3diffusion, foldflow}. Generative modelling with rigid frames outside the protein application, however, is limited. A concurrent work \citep{sigmadock} explores rigid-fragment $\mathrm{SE}(3)$ diffusion for molecular docking, which is an orthogonal task to ours. Similarly, \citet{jing2022torsional} leverage rigidity by operating on the space of torsion angles, but focus on conformer generation. \textit{De novo} generation of 3D structure of general molecules from rigid fragments thus remains underexplored.

%% file: contents/experiments.tex
\section{Experiments} 
\label{Experiments}
\begin{table}[t]
    \centering
    \begin{minipage}[t]{0.45\textwidth}
        \vspace{0pt} %
        \centering
        \footnotesize 
        \setlength{\tabcolsep}{1.0pt}
        \renewcommand{\arraystretch}{0.95} 
        \caption{Results of unconditional generation on \textsc{GEOM-Drugs}. $^\ddagger$Results are obtained by \citet{end}.}
        \label{geom_uncond}
        \begin{tabular}{@{} l c c c @{}}
        \toprule
        Method & Steps & Stable A, \% ($\uparrow$) & V $\times$ C, \% ($\uparrow$) \\ 
        \midrule
        Training Data & -- & $86.5$ & $99.0$ \\
        
        \midrule
        \midrule
        
        \makecell[l]{\textsc{GeoLDM} \\{\scriptsize\citet{geoldm}}}
          & $1000$ & $84.4$ & $45.8^\ddagger$ \\
        
        \makecell[l]{\textsc{EquiFM} \\{\scriptsize\citet{equifm}}}
          & $-$ & $84.1$ & $-$ \\
        
        \makecell[l]{\textsc{END} \\{\scriptsize\citet{end}}}
          & $100$  & $87.2_{\text{\tiny $\pm$.1}}$ & $73.7_{\text{\tiny $\pm$.4}}$ \\
        
        \makecell[l]{\textsc{EDM*} \\{\scriptsize\citet{end}}}
          & $250$  & $85.4_{\text{\tiny $\pm$.0}}$ & $61.4_{\text{\tiny $\pm$.6}}$ \\
        
        \makecell[l]{\textsc{GeoBFN} \\{\scriptsize\citet{geobfn}}}
          & $1000$ & $85.6$ & -- \\
        
        \makecell[l]{\textsc{VEDA-E} \\{\scriptsize\citet{veda}}}
          & $100$ & $90.0$ & -- \\
        \midrule
        
        \oursacro
            & $100$  & $\mathbf{95.0_{\scriptscriptstyle \pm.0}}$ & $\mathbf{81.2_{\scriptscriptstyle \pm.3}}$ \\
        \bottomrule
        \end{tabular}
    \end{minipage}%
    \hfill%
    \begin{minipage}[t]{0.55\textwidth}
        \vspace{0pt} %
        \centering
        \small
        \setlength{\tabcolsep}{1.0pt}
        \renewcommand{\arraystretch}{1.789} 
        \caption{Conditional generation results on \textsc{QM9}.}
        \label{cond_gen_table}
        \begin{tabular}{@{} l c c c @{}}
        \toprule
        \multirow{2}{*}{Method} & \multirow{2}{*}{Steps} & \multicolumn{1}{c}{\textsc{Composition}} & \multicolumn{1}{c}{\textsc{Substructure}} \\ 
        \cmidrule(lr){3-3}\cmidrule(l){4-4}
         & & Matching, $\%$ ($\uparrow$) & Tanimoto Sim. ($\uparrow$) \\
        \midrule
        
        \makecell[l]{\textsc{cEDM} \\ \scriptsize\citet{eegsde}}
          & $1000$ & -- & $.671_{\text{\tiny $\pm$.004}}$ \\
        
        \makecell[l]{\textsc{EEGSDE} \\ \scriptsize\citet{eegsde}}
          & $1000$ & -- & $.750_{\text{\tiny $\pm$.003}}$ \\
        
        \multirow{2}{*}{\makecell[l]{\textsc{cEDM}* \\ \scriptsize\citet{end}}}
          & $500$  & $76.2_{\text{\tiny $\pm$0.6}}$ & $.669_{\text{\tiny $\pm$.001}}$ \\
          & $1000$ & $75.5_{\text{\tiny $\pm$0.5}}$ & $.673_{\text{\tiny $\pm$.002}}$ \\
        
        \multirow{2}{*}{\makecell[l]{\textsc{cEND} \\ \scriptsize\citet{end}}}
          & $500$  & $91.5_{\text{\tiny $\pm$0.8}}$ & $.825_{\text{\tiny $\pm$.001}}$ \\
          & $1000$ & $91.0_{\text{\tiny $\pm$0.9}}$ & $.828_{\text{\tiny $\pm$.001}}$ \\
        \midrule
        
        \makecell[l]{\textsc{c}\oursacro}
          & $100$  & $\mathbf{95.4_{\scriptscriptstyle \pm0.5}}$ & $\mathbf{.862_{\scriptscriptstyle \pm.002}}$ \\
        \bottomrule
        \end{tabular}
    \end{minipage}
    \vspace{-6mm}
\end{table}
\paragraph{Tasks and Datasets} In Section \ref{Unconditional Generation}, we consider the task of unconditional generation on two common benchmarks, \textsc{QM9} \citep{qm9} and \textsc{GEOM-Drugs} \citep{geom}. The former dataset contains $134$k small organic molecules with up to $29$ atoms in total, with a maximum of $9$ heavy atoms. The latter, larger-scale dataset of molecular conformers comprises $430$k medium-sized molecules, with up to $181$ atoms and an average of $44.4$ atoms per molecule. For both datasets, we follow the same setup as in previous works \citep{edm, end, geodiff}. We then proceed to the two conditional experiments on \textsc{QM9} in Section \ref{Conditional Generation}, demonstrating our method's ability to generate molecules with desired properties. We conclude by studying the proposed rigid-motif fragmentation strategy and its modifications on the two datasets containing larger molecules, which is the principal use case of our method:  \textsc{GEOM-Drugs} and \textsc{QMugs} \citep{qmugs}. As \textsc{QMugs}, which contains large drug-like molecules with up to $100$ heavy atoms, lacks established generative baselines, we introduce its $300$k-conformation subset in our ablations to provide a reproducible benchmark and serve as a baseline for future works.
\vspace{-3mm} 
\paragraph{Baselines} We compare the performance of \oursacro to models within the same paradigm of inferring bonds from interatomic distances \citep{edm, edm_bridge, geoldm, equifm, end, geobfn}, including the concurrent \textsc{VEDA-E} \citep{veda}. For reference, we also provide training data statistics obtained under this bond lookup strategy. For experimental details and extended results, see Appendix \ref{Experimental Details} and \ref{Extended Results}.
\subsection{Unconditional Generation} \label{Unconditional Generation}
For this task, we follow \citet{end} and sample $10^4$ molecules across $3$ seeds, reporting the mean and standard deviation. For each baseline with fixed-step solvers, we provide the step configuration that yielded the best molecular stability for \textsc{QM9} and atom stability for \textsc{GEOM-Drugs}, among those reported in the respective papers.
\vspace{-3mm} 
\paragraph{Metrics} \looseness=-1 Following the established practice \citep{edm, equifm}, we report the percentages of stable atoms A and molecules M, as well as valid V and valid and unique V$\times$U molecules computed on \texttt{RDKit} \citep{rdkit}. Since for larger \textsc{GEOM-Drugs}, the connectivity of the generated molecules is commonly reported to be more challenging, while practically all samples are unique, we follow \citet{end} and report the percentage of valid and connected V$\times$C molecules instead. In line with the baselines, we also do not report molecular stability for \textsc{GEOM-Drugs}, as it was found to be non-informative when bonds are inferred from interatomic distances \citep{edm, equifm}.
\vspace{-3mm} 
\paragraph{\textsc{QM9}} Main results are provided in Table \ref{qm9_uncond}. The \textsc{QM9} benchmark is fairly saturated on unconditional generation and thus mainly reported for completeness. Overall, \oursacro performs on par with methods that use $10\times$ the number of generation steps: we obtain higher molecular stability and lower uniqueness, both being the consequence of using rigid motifs as larger building blocks on small molecules; this trade-off is controllable via sampling temperature in discrete flow and fragmentation granularity (see Section \ref{Ablations}; Table \ref{tab:qm9_temp}). The concurrent \textsc{VEDA-E} \citep{veda}, however, surpasses \oursacro across all metrics on this benchmark; Appendix \ref{app:uniqueness} traces the lower uniqueness to repetitions in motif multisets, an issue specific to fragmented smaller molecules, which are outside of our method's primary focus on complex, more realistic drug-like compounds. 
\vspace{-3mm}
\paragraph{\textsc{GEOM-Drugs}} \looseness=-1 Main results\footnote{\citet{end} have conflicting V$\times$C results on \textsc{GEOM-Drugs} for their \textsc{END} model reported in the appendix and main body of their paper; we resort to results provided in the main body.} are presented in Table \ref{geom_uncond}. On this dataset of larger, more realistic, and challenging molecules, \oursacro significantly outperforms the baselines. We note that our use of rigid motifs as larger building blocks allows us to ameliorate errors in atom valencies that the true atom-level data exhibits when assessed by the bond lookup of \citet{edm}.

\begin{figure}[t] %
    \centering
    \begin{minipage}[b]{0.48\linewidth}
        \centering 
        \includegraphics[width=\linewidth]{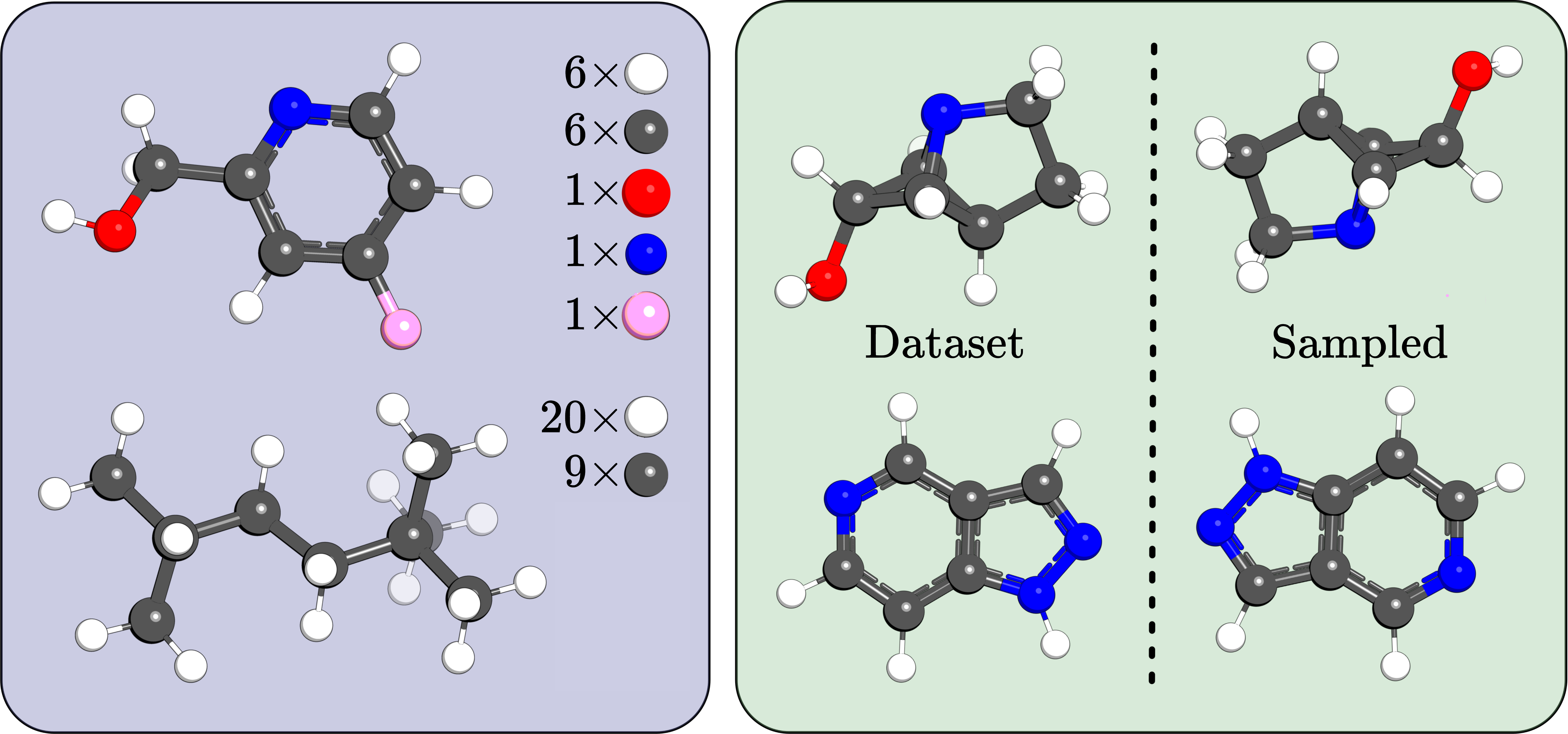}
        \captionof{figure}{Examples of results for the conditional tasks on \textsc{QM9}: atom composition (\textit{left}: samples with target atom counts) and substructure (\textit{right}: samples conditioned on dataset fingerprints).}
        \label{cond_gen_pic}
    \end{minipage}
    \hfill %
    \begin{minipage}[b]{0.48\linewidth}
        \centering 
        \setlength{\abovecaptionskip}{7.96pt}
        \includegraphics[width=\linewidth]{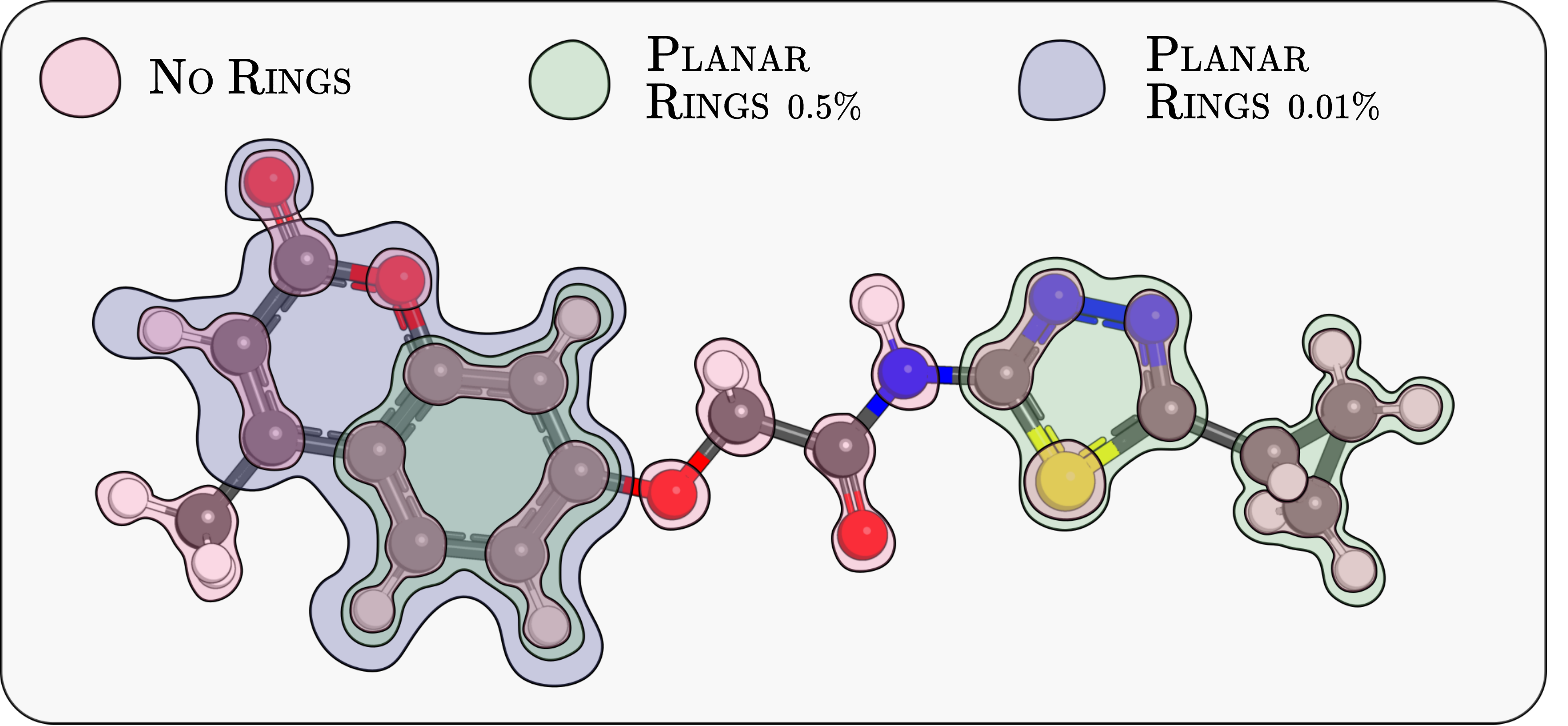}
        \caption{Resulting sets of rigid motifs for a \textsc{GEOM-Drugs} molecule $\text{C}_{17}\text{H}_{15}\text{N}_{3}\text{O}_{4}\text{S}$ under different fragmentation strategies. Each shaded region is a single rigid motif.}
        \label{decomp_various}
    \end{minipage}
    \vspace{-3mm}
\end{figure}

\subsection{Conditional Generation} \label{Conditional Generation}

With the conditional experiments, we seek to answer two primary questions that arise naturally from the proposed change in molecular representations:
\begin{enumerate}[label=(\roman*), noitemsep, topsep=0pt, parsep=0pt, partopsep=0pt, wide=0pt]
  \item \textit{Fine-grained conditioning}: does the motif-based representation perform competitively at generation conditioned on the \textit{atom-level} information, despite only modelling it implicitly?
  \item \textit{Coarse-grained conditioning}: do fragments lead to better generation of desired \textit{substructures}?
\end{enumerate}

We adopt the tasks for both scenarios from the previous works \citep{end, eegsde}. For the fine-grained level, we condition on the \textit{atom composition} $\mathbf{c} = (c_1, \dots, c_{|\mathcal{V}_a|}) \in \mathbb{R}^{|\mathcal{V}_a|}$, where $c_j$ is the number of atoms of the type $j$ that the desired sampled molecule is required to have. Following \citet{end}, we generate $10$ samples for each unique target atom composition from the validation and test sets across $3$ seeds, and report the percentage of matched compositions. For the coarse-grained level, we condition on the \textit{substructural features}: concretely, we use a molecular fingerprint $\mathbf{c} = (c_1, \dots,c_{F}) \in \{0, 1\}^F$, each entry of which indicates the presence or absence of a certain substructure in the molecule. We use \texttt{OpenBabel} \citep{openbabel} to generate the fingerprints for the test set, and evaluate the generation by computing Tanimoto similarity between the fingerprints of generated molecules and those from the test set, which are injected as conditions to the model. To ensure a fair comparison with the baselines in both tasks, we follow the strategy of \citet{end}: while many techniques for guiding flow models exist \citep{cfg, discrete_cfg}, in this section, we directly use the conditional model $v_\theta(\bm{\mathcal{M}}_t, t, \mathbf{c})$ by adding the conditioning information $\mathbf{c}$ directly to the input. We denote this version as \textsc{c}\oursacro. The results are in Table \ref{cond_gen_table}, with extended results and qualitative examples in Appendix \ref{Extended Results}. The tasks are summarised in Figure \ref{cond_gen_pic}. \textsc{c}\oursacro outperforms the baselines, achieving better adherence to conditions at both levels of molecular granularity while requiring fewer generation steps.

\begin{table}[t] %
    \centering
    \begin{minipage}[t]{0.42\textwidth}
        \vspace{0pt} 
        \centering
        \includegraphics[width=\linewidth]{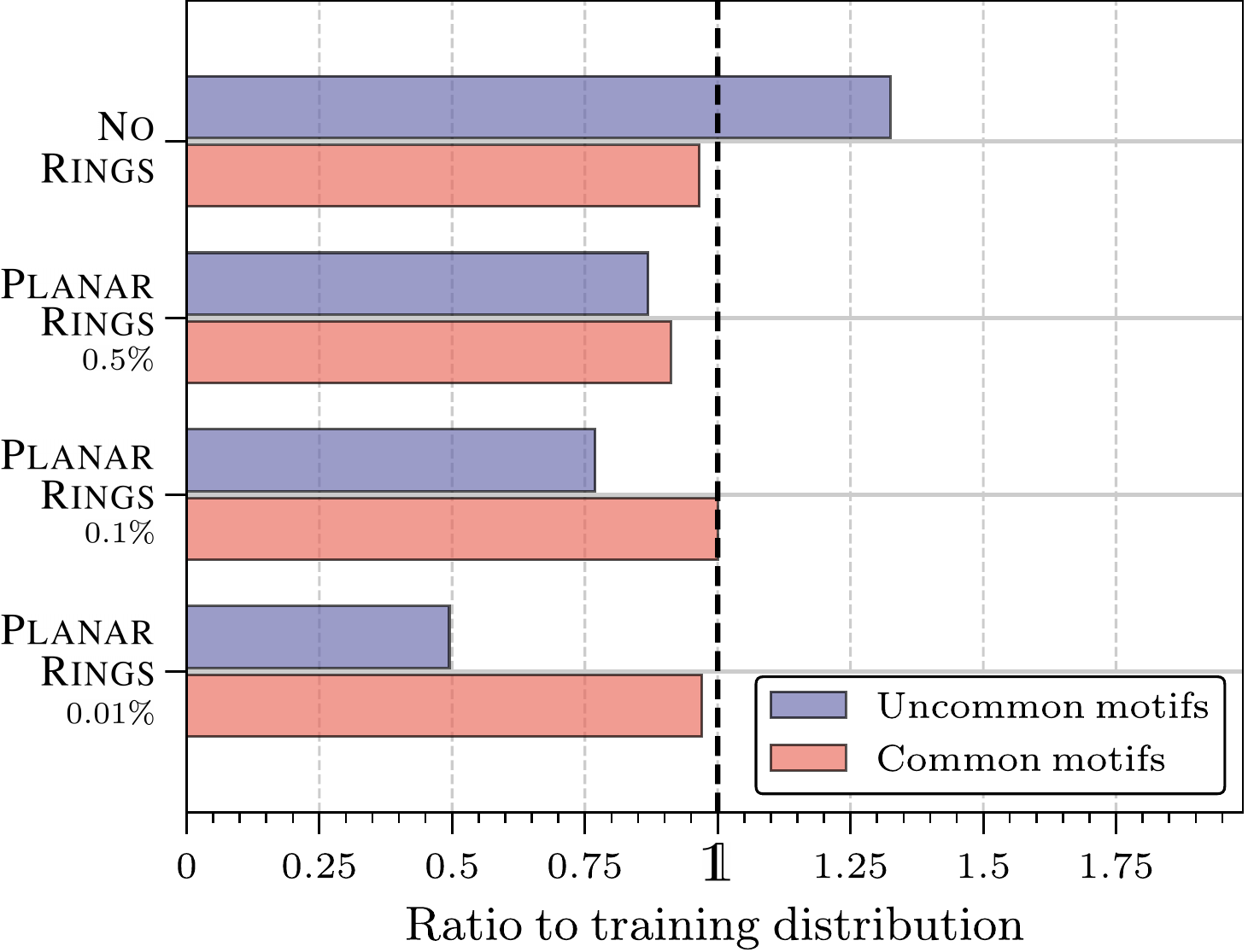}
        \captionof{figure}{Comparison of sampled motif types to frequencies in the training distribution of \textsc{QMugs}. A ratio of 1 is optimal.}
        \label{motif_distr_qmugs}
    \end{minipage}
    \hfill
    \begin{minipage}[t]{0.55\textwidth}
        \vspace{0pt}
        \vspace{-7.68pt}
        \centering
        \captionof{table}{Ablation results for fragmentation strategies. Metrics are reported for sampling with $100$ steps.}
        \label{ablation}
        \vspace{2.5mm}
        \footnotesize
        \setlength{\tabcolsep}{1.0pt}
        \renewcommand{\arraystretch}{1.50}
        \begin{tabular}{@{} l ccc ccc @{}}
        \toprule
        \multirow{3}{*}{\makecell[l]{Strategy}} & \multicolumn{3}{c}{\textsc{QMugs}} & \multicolumn{3}{c}{\textsc{GEOM-Drugs}} \\ 
        \cmidrule(lr){2-4} \cmidrule(l){5-7}
         & \multirow{2}{*}{$|\mathcal{V}_m|$} & Stability & V $\times$ C & \multirow{2}{*}{$|\mathcal{V}_m|$} & Stability & V $\times$ C \\
         & & A, \% & \% & & A, \% & \% \\
        \midrule
        \textsc{No Rings} & 49 & $85.5_{\pm1.8}$ & $\mathbf{85.6_{\pm.3}}$ & 39 & $86.7_{\pm1.8}$ & $\mathbf{82.9_{\pm.4}}$ \\
        \makecell[l]{\textsc{Planar} \\ \textsc{Rings} \scriptsize $0.5\%$} & 96 & $95.5_{\pm0.1}$ & $84.0_{\pm.5}$ & 81 & $95.2_{\pm0.1}$ & $82.1_{\pm.3}$ \\
        \makecell[l]{\textsc{Planar} \\ \textsc{Rings} \scriptsize $0.1\%$} & 253 & $\mathbf{96.1_{\pm0.0}}$ & $83.1_{\pm.4}$ & 202 & $95.0_{\pm0.0}$ & $81.2_{\pm.3}$ \\
        \makecell[l]{\textsc{Planar} \\ \textsc{Rings} \scriptsize $0.01\%$} & 1184 & $95.3_{\pm0.0}$ & $80.4_{\pm.5}$ & 867 & $\mathbf{96.3_{\pm0.0}}$ & $82.4_{\pm.3}$ \\
        \bottomrule
        \end{tabular}
    \end{minipage}
\vspace{-5mm}
\end{table}

\subsection{Ablations} \label{Ablations}
The purpose of this section is to further study the effects of the rigid-motif decomposition on 3D molecular generation. As noted by existing literature (e.g., \citealp{magnet}), the motif-based molecular graph decomposition offers a trade-off: while many larger functional groups are present in the data, adding larger motifs to the vocabulary leads, on average, to less frequent classes observed during training, which could make generalisation to uncommon substructures challenging. 

\looseness=-1 Yet we hypothesise that in 3D, larger rigid motifs, if sufficiently frequent, bring about the benefits of the fragment-based generation, which were empirically established in Sections \ref{Unconditional Generation} and \ref{Conditional Generation}. To verify this, we consider several variations of the rigid-motif decomposition. In the \textsc{No Rings} variant, we only preserve bonds to hydrogen atoms, as well as double and triple bonds. This configuration allows us to evaluate the performance of the finer-grained vocabulary with fewer rare fragments. With all other components fixed, it is the closest controlled proxy for an atom-level representation under our architecture, albeit not a complete atom-versus-motif comparison. On the opposite side of the spectrum, we consider our main decomposition introduced in Section \ref{Rigid-Motif Decomposition}, which preserves approximately planar rings and 
fused ring systems. For this \textsc{Planar Rings} fragmentation, we consider three thresholds (including our base one, $\alpha=0.1\%$) for the minimal frequency of a fragment relative to the dataset size; i.e., a smaller threshold corresponds to a larger vocabulary with more rare motifs, while with a larger threshold, they are decomposed further into finer components. We note that $\alpha$ is the only tunable parameter in fragmentation, and we use the same values across datasets. An example of a molecule from \textsc{GEOM-Drugs} under different rigid-motif decompositions is provided in Figure \ref{decomp_various}. Detailed information on the setup is in Appendix \ref{Fragmentation Ablation Details}.

First, we ablate the performance of each fragmentation strategy at the unconditional generation on \textsc{GEOM-Drugs} and \textsc{QMugs}; as it is for the main experiments, we evaluate the metrics based on $10^4$ samples obtained with $3$ seeds; results are in Table \ref{ablation}. Our hypothesis is largely confirmed: all rigid-motif vocabularies that treat planar rings as distinct classes yield superior atom stability compared to finer fragmentation, with the two smaller threshold configurations scoring best.

\looseness=-1 To assess the extent to which uncommon motifs are purposefully sampled during generation, we analyse the motif set of generated molecules. We first formally define \textit{common} and \textit{uncommon} motifs as those that exceed the base occurrence threshold $\alpha=0.1\%$ and those that do not reach its frequency in the data, respectively. For each group, we then compute the total counts of its constituent motifs, normalised to the total number of molecules, independently for the training and generated sets. The ratio of these normalised counts, if the model accurately represents the underlying distribution of motifs, should be close to $1$ for both common and uncommon motifs. Figure \ref{motif_distr_qmugs} shows that, while common motifs are consistently generated with close to the true occurrence, the uncommon motifs are notably oversampled for the finer \textsc{No Rings} strategy. Further, we observe an escalating undersampling behaviour for \textsc{Planar Rings} strategy with the decrease in threshold $\alpha$, confirming the expected trade-off between the vocabulary size and uncommon motif coverage \citep{magnet}. We thus conclude that frequency-based fragmentation with planar ring systems at a relatively high $\alpha = 0.1\%$ offers the optimal trade-off, significantly improving generation over conventional atom-based baselines and other rigid decomposition strategies.

%% file: contents/conclusion.tex
\section{Limitations, Future Work, and Broader Impact}
\label{Limitations, Future Work, and Broader Impact}
\looseness=-1 This method, while extending \textit{motif-based} molecular graph generation to the 3D space, also inherits its limitations. Most notably, the reliance on a predefined vocabulary introduces a trade-off between vocabulary size and generalisation: as the vocabulary grows to capture larger motifs, the frequency of individual classes drops, leading to lower coverage of uncommon substructures during sampling. While rare substructures remain representable via recursive decomposition into more frequent motifs, this does not guarantee generalisation to chemistry beyond the training distribution. An alternative, \textit{scaffold-based} generation \citep{magnet}, however, is non-trivial in 3D due to the vastly different spatial geometries of fragments of the same shape, and thus its integration into our method constitutes an exciting direction for future work. Of potential interest could also be an extension of our method to models that jointly generate 3D and 2D molecular structures (e.g., \citealp{codesign}), thereby bridging motif-based generation with explicit modelling of all bonds and enabling a controlled comparison with bond-explicit methods. Beyond the vocabulary, the benefit of \oursacro depends on the molecular regime: it is largest for medium-to-large molecules with recurring rigid substructures, and smaller for small or highly flexible ones, where little structure can be compressed. Finally, while the \textsc{No Rings} ablation (Section \ref{Ablations}) isolates the effect of rigid motifs under a fixed architecture, it does not fully disentangle the representation from model capacity, since an atom-level baseline cannot share our rigid-frame architecture.
\paragraph{Broader Impact}
Generative models for molecules have the potential to accelerate \textit{in silico} discovery and the design of novel drugs and materials, but they also carry potential dangers, as such models could be misused for designing chemicals with socially adverse properties. As our model is trained on benchmarks of drug-like molecules, we do not consider it to pose such a risk at its current stage.

\section{Conclusion}
\label{Conclusion}
\looseness=-1 In this work, we introduced \oursacro, a novel generative framework for 3D molecules that operates on \textit{rigid motifs} rather than individual atoms. By combining the rigid-motif decomposition strategy with the multimodal flow matching objective, we generate drug-like molecules via flows on the $\mathrm{SE(3)}$ manifold coupled with discrete categorical flows. Our empirical evaluation on standard benchmarks demonstrates that this higher-level representation yields improved stability on larger molecules compared to established bond-implicit all-atom methods, while offering more concise molecular representations as well as significant advantages in sampling efficiency and conditional generation.

%% file: contents/appendix.tex
\newpage
\appendix
\onecolumn

\section{Supplementary Material}

\subsection{Flow Matching in Euclidean Spaces}
\label{Euclidean Flow}
The task of generative modelling can be seen as transporting a sample $\mathbf{x}_0 \sim p(\mathbf{x})$ in $\mathbb{R}^d$ from a tractable prior distribution $p$ to a data sample $\mathbf{x}_1 \sim p^*(\mathbf{x})$ of the unknown data distribution $p^*$. In this formulation, the goal is thus to construct a \textit{probability density path} $p_t$, $t\in[0,1]$, such that $p_0 \approx p$, $p_1 \approx p^*$. 
\textit{Flow matching} \citep{fm, fm1, fm2} approaches it by introducing a \textit{flow}, i.e., an \textit{interpolant}, $\phi_t(\mathbf{x}_0)$, such that $\mathbf{x}_t = \phi_t(\mathbf{x}_0) \sim p_t$ if $\mathbf{x}_0 \sim p_0$. One can model $\phi_t(\mathbf{x}_0)$ as a solution of an ordinary differential equation (ODE) with a time-varying \textit{vector field} $u_t(\mathbf{x}_t)\colon \mathbb{R}^d \times [0, 1] \rightarrow \mathbb{R}^d$ and the initial condition $\phi_0(\mathbf{x}_0) = \mathbf{x}_0$:
\begin{equation}\label{eq:ode}
\frac{d\mathbf{x}_t}{dt} = u_t(\mathbf{x}_t) \quad \text{and} \quad \phi_1(\mathbf{x}_0) = \mathbf{x}_0 + \int_0^1 u_t(\mathbf{x}_t) \, dt.
\end{equation}
The training objective arises from regressing a neural network $v_\theta\colon \mathbb{R}^d \times [0, 1] \rightarrow \mathbb{R}^d$ to the vector field $u_t$ in \eqref{eq:ode}:
\begin{equation}\label{eq:loss_fm}
    \mathcal{L}_{\text{FM}}(\theta) = \mathbb{E}_{\substack{\mathbf{x}_t \sim p_t(\mathbf{x}) \\ t \sim \mathcal{U}(0, 1)}} \|v_\theta(\mathbf{x}_t, t) - u_t(\mathbf{x}_t)\|^2.
\end{equation}
Computing this loss in \eqref{eq:loss_fm} requires access to intractable $u_t(\mathbf{x})$ and $p_t$. It can be shown (e.g., \citealp{fm}) that optimising $\mathcal{L}_{\text{FM}}$ is equivalent in expectation to optimising the \textit{conditional flow matching} (CFM) objective $\mathcal{L}_{\text{CFM}}$:
\begin{equation}\label{eq:loss_cfm}
    \mathcal{L}_{\text{CFM}}(\theta) = \mathbb{E}_{\substack{\mathbf{z} \sim p(\mathbf{z}), \,\mathbf{x}_t \sim p_t(\mathbf{x} \mid \mathbf{z}) \\ t \sim \mathcal{U}(0, 1)}} \|v_\theta(\mathbf{x}_t, t) - u_t(\mathbf{x}_t \mid \mathbf{z})\|^2,
\end{equation}
where, in \eqref{eq:loss_cfm}, $u_t(\mathbf{x} \mid \mathbf{z})\colon \mathbb{R}^d \times [0, 1] \rightarrow \mathbb{R}^d$ is a conditional vector field that generates a conditional probability path $p_t(\mathbf{x} \mid \mathbf{z})$. That is, instead of regressing over the marginal vector field, we regress over conditional vector fields. The conditioning is commonly done on the data point $\mathbf{z} = \mathbf{x}_1$ or on a source-target coupling $\mathbf{z} = (\mathbf{x}_0, \mathbf{x}_1)$ \citep{ot_cfm}.

\subsection{Flow Matching on Riemannian Manifolds}
\label{Riemannian Flow}
Riemannian flow matching \citep{rfm} generalises the above framework to a smooth manifold $\mathcal{X}$ equipped with a Riemannian metric $g$. A time-dependent vector field $u_t\colon \mathcal{X} \times [0, 1] \rightarrow \mathcal{T}\mathcal{X}$ assigns a tangent vector $u_t(\mathbf{x}) \in \mathcal{T}_{\mathbf{x}}\mathcal{X}$ to each point $\mathbf{x} \in \mathcal{X}$ and defines a flow on $\mathcal{X}$ via $\frac{d\mathbf{x}_t}{dt} = u_t(\mathbf{x}_t)$. For manifolds with closed-form geodesics, the conditional probability path can be constructed by the geodesic interpolant between $\mathbf{x}_0 \sim p_0$ and $\mathbf{x}_1 \sim p^*$, which is generated by the conditional vector field
\begin{equation}
\label{eq:geodesic}
\mathbf{x}_t = \exp_{\mathbf{x}_0}\left(t \log_{\mathbf{x}_0}(\mathbf{x}_1)\right), \qquad u_t(\mathbf{x}_t \mid \mathbf{x}_1) = \frac{1}{1-t}\log_{\mathbf{x}_t}(\mathbf{x}_1),
\end{equation}
where $\exp$ and $\log$ in \eqref{eq:geodesic} denote the Riemannian exponential and logarithmic maps. Analogous to \eqref{eq:loss_cfm}, the Riemannian conditional flow matching objective in \eqref{eq:loss_rcfm} regresses the neural vector field onto $u_t$ with respect to the norm induced by $g$:
\begin{equation}
\label{eq:loss_rcfm}
\mathcal{L}_{\text{RCFM}}(\theta) = \mathbb{E}_{\substack{\mathbf{x}_0 \sim p_0, \, \mathbf{x}_1 \sim p^* \\ t \sim \mathcal{U}(0, 1)}} \left\|v_\theta(\mathbf{x}_t, t) - u_t(\mathbf{x}_t \mid \mathbf{x}_1)\right\|_g^2.
\end{equation}
For $\mathrm{SO}(3)$, \citet{foldflow} use the uniform prior (with a reversed time convention) and evaluate both maps in closed form in the Lie algebra $\mathfrak{so}(3)$ of skew-symmetric matrices. Identifying $\boldsymbol{\omega} \in \mathbb{R}^3$ with $[\boldsymbol{\omega}]_\times \in \mathfrak{so}(3)$ and denoting $\omega = \|\boldsymbol{\omega}\|$, Rodrigues' formula gives
\begin{equation}
\label{eq:rodrigues}
\exp\left([\boldsymbol{\omega}]_\times\right) = \mathbf{I} + \frac{\sin\omega}{\omega}[\boldsymbol{\omega}]_\times + \frac{1-\cos\omega}{\omega^2}[\boldsymbol{\omega}]_\times^2,
\end{equation}
while, inverting \eqref{eq:rodrigues}, for $\omega \in (0, \pi)$ the logarithm of $\mathbf{R} \in \mathrm{SO}(3)$ is $\log(\mathbf{R}) = \frac{\omega}{2\sin\omega}(\mathbf{R}-\mathbf{R}^\top)$ with $\omega = \arccos\left(\frac{1}{2}(\mathrm{Tr}(\mathbf{R})-1)\right)$. The maps at an arbitrary point $\mathbf{R}_t$ follow by left translation, e.g., $\log_{\mathbf{R}_t}(\mathbf{R}_1) = \mathbf{R}_t \log(\mathbf{R}_t^\top \mathbf{R}_1)$, as described in Section \ref{SE(3) Flow}.

\subsection{Discrete Flow Matching}
\label{app:discrete_flows}
Discrete flow matching \citep{discrete_flows} transports a prior over a finite state space $\mathcal{V}_m$ to the data distribution via a CTMC with rate matrix $\mathbf{Q}_t$ (Section \ref{Discrete Flows}). Generation is simulated with Euler steps of size $\Delta t$,
\begin{equation}
\label{eq:ctmc_euler}
m_{t+\Delta t} \sim \mathrm{Cat}\left(\delta_{m_t} + \mathbf{Q}^\theta_t(m_t, \cdot)\,\Delta t\right),
\end{equation}
where the diagonal of $\mathbf{Q}^\theta_t$ is set such that each row sums to zero. For a conditional probability path $p_t(\cdot \mid m_1)$, \citet{discrete_flows} show that a valid conditional rate matrix generating it is given, for $j \neq k$ with $p_t(j \mid m_1), p_t(k \mid m_1) > 0$ (and zero otherwise), by
\begin{equation}
\label{eq:rate_star}
\mathbf{Q}^*_t(j, k \mid m_1) = \frac{\mathrm{ReLU}\left(\partial_t p_t(k \mid m_1) - \partial_t p_t(j \mid m_1)\right)}{\mathcal{Z}_t \, p_t(j \mid m_1)},
\end{equation}
where $\mathcal{Z}_t$ is the number of states with non-zero probability under $p_t(\cdot \mid m_1)$. For the masking path in Section \ref{Multimodal Flow}, \eqref{eq:rate_star} recovers the unmasking rate $(1-t)^{-1}$. The rate matrix is, however, not unique: any $\mathbf{Q}^{\mathrm{DB}}_t$ satisfying detailed balance with respect to $p_t(\cdot \mid m_1)$ can be added without changing the probability path,
\begin{equation}
\label{eq:rate_eta}
\mathbf{Q}^\eta_t(\cdot \mid m_1) = \mathbf{Q}^*_t(\cdot \mid m_1) + \eta \, \mathbf{Q}^{\mathrm{DB}}_t(\cdot \mid m_1), \qquad \eta \geq 0,
\end{equation}
where $\eta$ in \eqref{eq:rate_eta} controls the stochasticity of sampling. For the masking path, \citet{discrete_flows} take $\mathbf{Q}^{\mathrm{DB}}_t(m_1, [\text{MASK}]) = 1$ and $\mathbf{Q}^{\mathrm{DB}}_t([\text{MASK}], m_1) = \frac{t}{1-t}$: unmasked motifs are \textit{remasked} at rate $\eta$ (i.e., with probability $\eta\Delta t$ per step of \eqref{eq:ctmc_euler}), while the unmasking rate rises to $\frac{1+\eta t}{1-t}$, which allows the model to revise earlier decisions. We refer to $\eta$ as the remasking rate.

\subsection{Implementation Details} \label{Implementation Details}

\paragraph{Geometric Matching and Canonicalisation}
To ensure geometric robustness, we quantify the dimensionality of motifs using singular value decomposition (SVD) on the centred coordinate matrix $\mathbf{P} \in \mathbb{R}^{N \times 3}$. Let $\sigma_1 \ge \sigma_2 \ge \sigma_3$ denote the singular values of $\mathbf{P}$. We classify a motif as \textit{linear} if $\sigma_2 < 0.05 \cdot \sigma_1$, and \textit{planar} if $\sigma_3 < 0.05 \cdot \sigma_1$. This check determines whether dummy atoms are required to lock the coordinate frame. During vocabulary construction, a candidate fragment instance matches an existing vocabulary class if its structural similarity falls within a specified tolerance. We align the instance to the exemplar using the Kabsch algorithm with atom-wise weights $w$, setting $w=1.0$ for heavy atoms and $w=0.1$ for hydrogens and dummy atoms. This weighting ensures the alignment is driven by the rigid heavy-atom scaffold, while still utilising lighter atoms to uniquely define the orientation of frames with collinear or planar heavy-atom cores. A match is accepted if the weighted root mean square deviation (RMSD) is $\le 1.0$\,\AA\ and the maximum single-atom deviation is $\le 1.0$\,\AA. Since we pre-select naturally rigid chemical moieties, these rather loose thresholds are not sensitive to minor structural noise (e.g., bond length vibrations); rather, they ensure that, e.g., distinct \textit{geometric stereoisomers}, such as the \textit{(E)}- and \textit{(Z)}-isomers of alkenes that share the same graph topology but differ in rigid 3D arrangement, are correctly assigned to separate vocabulary entries.

\paragraph{Exact Kabsch Alignment via Automorphisms}
As described in Section \ref{Canonicalisation and Vocabulary}, establishing the orientation of a motif instance requires aligning it to a canonical pose. In generic 3D point cloud tasks, aligning unordered sets of $N$ points requires exploring $O(N!)$ permutations, which is computationally intractable even for small molecules. However, because our preprocessing explicitly treats molecules as graphs with chemical elements, we restrict the permutation space strictly to valid graph automorphisms $\pi \in \Pi_i$ that preserve topology and chemical identities. By pre-computing these automorphisms, we guarantee that the subsequent Kabsch alignment is both exact and computationally efficient, bounding the symmetry group to tractable sizes.

\paragraph{Architecture}
Our model architecture is based on the \textsc{FoldFlow-Base} backbone \citep{foldflow}, which utilises Invariant Point Attention (IPA) \citep{alphafold2} to process 3D rigid frames. To adapt this protein-specific architecture for general rigid-motif molecular generation, we make several key modifications. We reduce the hidden dimension per attention head to $64$ for each of $8$ heads, resulting in a compact model size of $12.4$ million parameters for the unconditional model, compared to $17$ million in the original \textsc{FoldFlow-Base}. The conditional variant, \textsc{c}\oursacro, utilises $13.7$ million parameters due to the additional projection layers required for processing conditioning signals.

\paragraph{Embeddings and Input Processing}
The network accepts a set of noisy rigid frames $\mathbf{T}_t$ and discrete motif tokens $m_t$. The node embeddings are initialised by projecting the motif token embeddings, sinusoidal timestep embeddings, and, during training, with $50\%$ probability, the self-conditioning embeddings derived from the previous step's prediction.

Unlike protein models that rely on residue index offsets for edge initialisation, our motifs have no intrinsic linear ordering. Consequently, we replace sequence-based edge biases with purely geometric information. We compute pairwise Euclidean distances between the centres of rigid motifs and encode them using Gaussian Radial Basis Functions (RBFs) with $64$ basis functions. These geometric features are concatenated with the cross-concatenated node features to form the initial edge embeddings. The point cloud of fragments is treated as a fully connected graph, i.e., each pair of fragments is assigned an edge embedding.

\paragraph{Backbone and Updates}
The backbone consists of 4 blocks, each containing an IPA layer, transition modules, and 2 transformer encoder layers. To better capture the global geometric constraints of an unordered molecule, we augment the standard IPA blocks with triangle multiplicative updates \citep{alphafold2}. Specifically, we apply both outgoing and incoming triangle updates to the edge representations before the attention mechanism in every block.

For the geometry updates, we employ a split-head design: the rotational update is predicted via a linear layer initialised to zero, while the translational update is predicted via a small 2-layer MLP.

\paragraph{Conditioning}
For conditional generation tasks, we follow the conditioning strategy of \citet{end}. The conditioning signal $\mathbf{c}$, i.e., either atom composition or substructure fingerprint, is encoded into a global context vector via a dedicated MLP. For composition conditioning, we learn embeddings for each atom type and compute a weighted sum based on the target counts before passing it to the MLP. This global context is injected into the network at three points: 
\begin{enumerate*}[label=(\roman*)]
    \item concatenated to the initial node embeddings,
    \item added as a bias to the node representations within each IPA block, and
    \item concatenated to the final node representations before the discrete readout head.
\end{enumerate*}

\paragraph{Training}
We train the model using the flow matching objective described in Section \ref{Method}, \eqref{eq:loss_total}, where $\mathcal{L}_{\text{DFM}}$ is evaluated at masked positions, and $\mathcal{L}_{\text{trans}}$ and $\mathcal{L}_{\text{rot}}$ are averaged over all motifs. In each experiment, we use a learning rate of $10^{-4}$.

\paragraph{Sampling} 
For the rotational component of the $\mathrm{SE}(3)$ flow, we use the exponential rate scheduler \citep{foldflow} with the factor of 10, following \citet{frameflow}.

\subsection{Extended Discussion on Motif-Based Generation}
\label{app:extended_discussion}

\paragraph{Prior Work on Fragments and Synthesisability}
While atom-level generation has dominated 3D molecular machine learning, the concept of building molecules from fragments has a rich history, particularly in 2D graph generation and reinforcement learning. A motif is typically considered \textit{chemically meaningful} when it corresponds to stable functional groups, common ring systems, or established synthetic building blocks. Approaches utilising reaction templates, such as the framework by \citet{gottipati2020learning}, have shown that assembling molecules from predefined building blocks can offer advantages in navigating towards synthetically accessible chemical space.

Recent literature has increasingly leveraged this hierarchical abstraction across various generative tasks. For instance, \citet{levy2023molecular} and \citet{frag_fm} explored fragment-based diffusion and discrete flow matching models, respectively, for assembling molecular graphs from predefined structural vocabularies. In the realm of 3D geometries, \citet{park2024equivariant} proposed equivariant blurring diffusion for hierarchical conformer generation, predicting 3D geometries for a known 2D graph from coarse-grained down to fine-grained scales. Our work extends this motif-based philosophy to the distinct task of \textit{de novo} 3D generation, treating molecules as collections of natively rigid geometric frames. This enforces local geometric validity (e.g., planar aromatic rings) by construction, rather than relying on the generative model to implicitly learn it from atomic coordinates.

\paragraph{Comparison with Prior Fragment-Based Methods}\label{app:comparison}
The individual generative components of \oursacro are established: the continuous component builds on $\mathrm{SE}(3)$ flow matching \citep{foldflow}, and the categorical component builds on discrete flow matching \citep{discrete_flows}. Our contribution is the representation and the machinery required to make these components operate jointly for \textit{de novo} 3D molecule generation. Concretely, \oursacro
\begin{enumerate*}[label=(\roman*)]
  \item represents a molecule as an unordered set of discrete motif identities and full $\mathrm{SE}(3)$ poses;
  \item develops data-driven rigid fragmentation and vocabulary construction for arbitrary molecular topologies;
  \item canonicalises each retained motif while accounting for its discrete symmetries in 3D;
  \item defines frames for otherwise underdetermined single-atom or collinear fragments using a fixed dummy-atom construction, making the method applicable to any non-collinear molecule composed of supported atom types; and
  \item lifts generated motifs back to atoms by a fixed rigid transformation, without sequential autoregressive assembly or a learned atom-level decoder.
\end{enumerate*}
Table \ref{tab:fragment_methods} contrasts \oursacro with prior fragment-based approaches. \textsc{HierDiff} \citep{hierdiff}, the closest existing non-autoregressive \textit{de novo} 3D fragment method, instead decodes coarse fragment nodes through a learned fine-grained model and iterative refinement. \textsc{AutoFragDiff} \citep{autofragdiff} and \textsc{AutoDiff} \citep{autodiff} perform pocket-conditioned, fragment-wise autoregressive generation and do not provide direct set-valued generation with a deterministic rigid lift. \textsc{SigmaDock} \citep{sigmadock}, by contrast, is not a \textit{de novo} method: it performs molecular docking and takes the ligand graph as input. Finally, \textsc{DecompDiff} \citep{decompdiff} and \textsc{DecompOpt} \citep{decompopt} use molecular decomposition to structure priors and conditioning, while the generation of the \textit{de novo} target molecule remains inherently atom-level. The novelty of \oursacro is thus not merely placing motifs inside an existing $\mathrm{SE}(3)$ flow matching framework, but making this representation well-defined for general topologies, which permits direct motif-based generation in 3D without autoregressive assembly, learned atom-level decoding, or iterative refinement.

\paragraph{Representation Efficiency and Architectural Scalability}\label{app:scalability}
A recent trend in molecular and protein structure modelling is to replace specialised equivariant layers with generic, highly scalable architectures, as exemplified by the transition from \textsc{AlphaFold2} \citep{alphafold2} to \textsc{AlphaFold3} \citep{alphafold3}. Indeed, IPA layers \citep{alphafold2}, used in the \textsc{FoldFlow-Base} architecture \citep{foldflow} that \oursacro builds upon, are more expensive per token and scale less favourably than generic transformers. It is, however, important to separate this architectural scalability from the representation efficiency of \oursacro. By representing frequently occurring rigid substructures as single units, we substantially reduce the number of states describing a molecule (Figure \ref{qmugs_distr}). At the scale of \textsc{FoldFlow-Base}, this deterministically reconstructible compression offsets the overhead of the specialised architecture: \oursacro samples faster than atom-level baselines across various generative frameworks, which are also equivariant (Table \ref{qm9_walltime_results}). This establishes practical end-to-end efficiency at the evaluated scale rather than a controlled comparison against an architecture-agnostic model, and does not imply that IPA-based architectures remain optimal at very large parameter or molecular scales. Importantly, the rigid-motif representation itself is not tied to strictly $\mathrm{SE}(3)$-equivariant architectures: e.g., \textsc{MOFFlow-2} \citep{mofflow2} uses a non-equivariant transformer to predict building-block translations and rotations under Riemannian flow matching. Coupling the rigid-motif parametrisation with such architectures thus constitutes a promising direction for future work.

\begin{table}[ht]
\centering
\small
\caption{Comparison of \oursacro with prior fragment-based approaches to molecule generation.}
\label{tab:fragment_methods}
\begin{tabular}{@{} l c c @{}}
\toprule
Approach & \makecell{\textit{De novo}\\fragment-based\\generation in 3D} & \makecell{Deterministic\\fragment-to-atom\\lift in 3D} \\
\midrule
\makecell[l]{Fragment-based graph generation \\ \scriptsize\citet{jtvae, hiervae, moler, micam}} & \no & \no \\
\makecell[l]{\textsc{HierDiff} \\ \scriptsize\citet{hierdiff}} & \yes & \no \\
\makecell[l]{\textsc{SigmaDock} \\ \scriptsize\citet{sigmadock}} & \no & \yes \\
\makecell[l]{Pocket-conditioned autoregressive methods \\ \scriptsize\citet{autofragdiff, autodiff}} & \yes & \no \\
\makecell[l]{Decomposition-guided pocket methods \\ \scriptsize\citet{decompdiff, decompopt}} & \no & \no \\
\midrule
\oursacro & \yes & \yes \\
\bottomrule
\end{tabular}
\end{table}

\paragraph{Compatibility with Acceleration Techniques}
Because our method is built upon standard $\mathrm{SE}(3)$ and categorical flow matching formulations (where the latter admits a continuous-flow perspective), it natively supports recent speed-up techniques, such as distillation and flow maps \citep{roos2026categorical, potaptchik2026discrete, davis2026generalised}. Crucially, the representational shift from atom space to rigid-motif space is orthogonal to these acceleration methods. Consequently, \oursacro can be freely coupled with them to achieve further advantages in efficiency via the one-step fragment-based 3D generation without requiring fundamental changes to the underlying framework.

\subsection{Experimental Details} \label{Experimental Details} 

\subsubsection{Baseline Details} \label{Baseline Details}

In this section, we briefly overview the baseline methods used for comparison in our experiments. All baselines are atom-based generative models that generate 3D molecules by determining the type and coordinates of individual atoms.

\textbf{\textsc{EDM} \citep{edm}} is a seminal work that introduces a score-based generative model operating directly on the continuous coordinates and discrete atom types of the molecule. It utilises an $\mathrm{E}(3)$-equivariant graph neural network \citep{engnn} to learn a denoising process that reverses a diffusion process, which transforms data into Gaussian noise. The model ensures that the generated likelihood is invariant to rotations and translations.

\textbf{\textsc{EDM-Bridge} \citep{edm_bridge}} improves upon the \textsc{EDM} framework by replacing the standard Gaussian prior with an informative prior that contains structural or physical information. It learns a bridge process that connects this informative prior to the data distribution, utilising Lyapunov functions to guide the generation and improve stability and validity.

\textbf{\textsc{GeoLDM} \citep{geoldm}} is a latent diffusion framework for 3D molecules. Unlike \textsc{EDM}, which operates in the data space, \textsc{GeoLDM} first compresses molecular structures into a low-dimensional, continuous latent space using an $\mathrm{E}(3)$-invariant autoencoder. A diffusion model is then trained in this latent space, enabling more efficient sampling and the capture of high-level geometric features.

\textbf{\textsc{EquiFM} \citep{equifm}} applies the flow matching framework to 3D molecular generation. It proposes a hybrid probability transport path: it uses optimal transport for continuous atomic coordinates to generate straight trajectories, and a conditional probability path for discrete atom types. This formulation allows significantly faster sampling speeds than diffusion-based equivalents while maintaining $\mathrm{SE}(3)$ equivariance.

\textbf{\textsc{END} \citep{end}} is a recent diffusion-based method that introduces a learnable forward process. Unlike standard diffusion models that use a fixed, predefined corruption process (e.g., adding Gaussian noise), END parametrises the forward process with a time- and data-dependent transformation that is equivariant to rigid transformations. This flexibility allows the model to learn a more optimal degradation and restoration path for molecular geometries.

\textbf{\textsc{GeoBFN} \citep{geobfn}} adapts the Bayesian flow network \citep{bfn} framework to 3D geometry. Instead of adding noise to the data directly, \textsc{GeoBFN} operates on the parameters of the data distribution (e.g., mean and variance for coordinates, probabilities for atom types). It updates these parameters iteratively using Bayesian inference and a neural network, unifying the generation of discrete and continuous modalities in a differentiable parameter space.

\textbf{\textsc{VEDA} \citep{veda}} is an $\mathrm{SE}(3)$-equivariant variance-exploding diffusion model for 3D molecules. It employs a noise schedule that enables noise injection analogous to simulated annealing, a preconditioning scheme reconciling coordinate-predicting equivariant networks with a residual-based diffusion objective, and an arcsine-based scheduler concentrating sampling steps in critical intervals of the signal-to-noise ratio, which allows for generation with a reduced number of steps. We compare against its bond-implicit variant, \textsc{VEDA-E}, in which bonds are not generated but inferred post hoc from interatomic distances and chemical valence rules, which matches the bond-inference protocol of \oursacro and the other baselines. Results for \textsc{VEDA-E} are taken from \citet{veda}; for the wall-clock comparison, we reproduced its published architecture and sampler.

\subsubsection{Evaluation Metrics} \label{Evaluation Metrics}

We adopt the evaluation framework and metric definitions from previous works \citep{frag_fm, end, hierdiff}. All metrics are implemented using \texttt{RDKit} \citep{rdkit}. Consistent with the reference methodology \citep{frag_fm, end}, unless otherwise noted, raw generated coordinate samples are converted into molecular objects using \texttt{OpenBabel} \citep{openbabel} and property metrics are calculated exclusively on samples that are both valid and connected.

\begin{itemize}
    \item \textbf{Atom and Molecule Stability}: atom stability is determined by valency constraints; an atom is stable if its charge is 0. Molecular stability requires every atom in the structure to be stable. Bond types are inferred from pairwise atomic distances using the lookup table method from \citet{edm}.

    \item \textbf{Validity and Connectivity}: \textit{validity} measures the percentage of generated samples that successfully pass \texttt{RDKit}'s parsing and sanitisation checks. However, standard sanitisation does not penalise disconnected molecules. To address this, particularly for larger molecules in \textsc{GEOM-Drugs} and \textsc{QMugs} datasets, we report \textit{connectivity}, which verifies that a valid molecule consists of a single connected component.

    \item \textbf{Uniqueness}: in line with all baselines, this metric reports the percentage of valid samples that possess a unique canonical SMILES string \citep{smiles}. Following standard practice for larger molecules in \textsc{GEOM-Drugs} and \textsc{QMugs} datasets, we omit uniqueness from those results as generated samples are almost invariably unique.

    \item \textbf{Total Variation (TV)}: to measure the distributional fidelity of atom and bond types, we compute the total variation. This is defined as the mean absolute error between the marginal distributions of atom and bond types in the training set and those in the generated samples.

    \item \textbf{Strain Energy}: this metric assesses the geometric quality of the molecules. It is calculated as the difference between the energy of the generated structure and the energy of its relaxed conformation. The relaxation is performed using the MMFF force field within \texttt{RDKit} \citep{rdkit}.

    \item \textbf{Chemical Properties (SA, QED)}: we report a couple of standard molecular descriptors:
    \begin{itemize}
        \item \textbf{SA}: the Synthetic Accessibility score \citep{sa_score} estimates the difficulty of synthesising a molecule. We follow \citet{end} and normalise this score to the interval $[0, 1]$, where 1 indicates high synthesisability (easy to synthesise) and 0 indicates low synthesisability.
        \item \textbf{QED}: the Quantitative Estimation of Drug-likeness.
    \end{itemize}
\end{itemize}

\subsection{Extended Experimental Results} \label{Extended Results}

\subsubsection{Unconditional Generation}
\paragraph{\textsc{QM9}}
We provide the extended results in Table \ref{qm9_uncond_extended}.
\renewcommand{\arraystretch}{1.15}
\begin{table}[htbp]
\centering
\small
\setlength{\tabcolsep}{1.6pt}
\caption{Extended results at unconditional generation on \textsc{QM9} with additional metrics from \citet{hierdiff}. $^\dagger$\citet{equifm} use Dopri5 adaptive solver. Bond-explicit methods generate bonds jointly with the 3D structure, and their stability is computed on the generated bonds; results are taken from the original publications. $^\S$Connected validity. \textsc{FlowMol-CTMC} is referred to as \textsc{FlowMol2} in later work \citep{veda}.}
\label{qm9_uncond_extended}
\begin{tabular}{l c c c c c c c c c c}
\toprule
\multirow{2}{*}{Method} & \multirow{2}{*}{Steps} & \multicolumn{2}{c}{Stability ($\uparrow$)} & \multicolumn{2}{c}{Val. / Uniq. ($\uparrow$)} & \multicolumn{2}{c}{TV ($\downarrow$)} & \multicolumn{1}{c}{Str. En. ($\downarrow$)} & \multirow{2}{*}{SA ($\uparrow$)} & \multirow{2}{*}{QED ($\uparrow$)} \\ 
\cmidrule(l){3-8}
 & & A, \% & M, \% & V, \% & V $\times$ U, \% & A$\left[10^{-2}\right]$ & B$\left[10^{-3}\right]$ & $\Delta \text{E} \left[\frac{\text{kcal}}{\text{mol}}\right]$ & & \\
\midrule
\makecell[l]{Training Data} & -- & $99.0$ & $95.2$ & $97.7$ & $97.7$ & -- & -- & $7.7$ & $0.63$ & $0.46$ \\
\specialrule{0.7pt}{3pt}{3pt}
\rowcolor{bondgray}\multicolumn{11}{l}{\textit{Bond-explicit methods (for reference; not directly comparable)}} \\
\rowcolor{bondgray}\begin{tabular}[c]{@{}l@{}}\textsc{SemlaFlow} \\ \scriptsize\citet{semlaflow}\end{tabular} & $100$ & $99.9$ & $99.7$ & $99.4$ & -- & -- & -- & -- & -- & -- \\
\rowcolor{bondgray}\begin{tabular}[c]{@{}l@{}}\textsc{FlowMol-} \\ \textsc{CTMC} \\ \scriptsize\citeauthor{flowmol2} \\ \scriptsize[\citeyear{flowmol2}]\end{tabular} & $100$ & -- & $99.3$ & $99.3$ & -- & -- & -- & -- & -- & -- \\
\rowcolor{bondgray}\begin{tabular}[c]{@{}l@{}}\textsc{EQGAT-diff} \\ \scriptsize\citet{eqgatdiff}\end{tabular} & $500$ & $99.9$ & $98.7$ & $99.0$ & -- & -- & -- & -- & -- & -- \\
\rowcolor{bondgray}\begin{tabular}[c]{@{}l@{}}\textsc{JODO} \\ \scriptsize\citet{jodo}\end{tabular} & -- & $99.9$ & $98.8$ & $99.0^\S$ & $96.0$ & -- & -- & -- & -- & -- \\
\rowcolor{bondgray}\begin{tabular}[c]{@{}l@{}}\textsc{Megalodon} \\ \scriptsize\citeauthor{codesign} \\ \scriptsize[\citeyear{codesign}]\end{tabular} & $500$ & $99.9$ & $98.6$ & $98.7^\S$ & -- & -- & -- & -- & -- & -- \\
\specialrule{0.7pt}{3pt}{3pt}

\makecell[l]{\textsc{EDM} \\ \scriptsize\citeauthor{edm} \\ \scriptsize[\citeyear{edm}]}
  & $1000$ & $98.7$ & $82.0$ & $91.9$ & $90.7$ & -- & -- & -- & -- & -- \\
\midrule

\makecell[l]{\textsc{EDM-Bridge} \\ \scriptsize\citet{edm_bridge}}
  & $1000$ & $98.8$ & $84.6$ & $92.0$ & $90.7$ & -- & -- & -- & -- & -- \\
\midrule

\makecell[l]{\textsc{GeoLDM} \\ \scriptsize\citet{geoldm}}
  & $1000$ & $98.9_{\text{\tiny $\pm$.1}}$ & $89.4_{\text{\tiny $\pm$.5}}$ & $93.8_{\text{\tiny $\pm$.4}}$ & $92.7_{\text{\tiny $\pm$.5}}$ & $1.6$ & $1.3$ & $10.4$ & -- & -- \\
\midrule

\makecell[l]{\textsc{EquiFM} \\ \scriptsize\citet{equifm}}
  & $-^{\dagger}$ & 
  $98.9_{\text{\tiny $\pm$.1}}$ & $88.3_{\text{\tiny $\pm$.3}}$ & 
  $94.7_{\text{\tiny $\pm$.4}}$ & $93.5_{\text{\tiny $\pm$.3}}$ &
  -- & -- & -- & -- & -- \\
\midrule

\multirow{5}{*}{\makecell[l]{\textsc{END} \\ \scriptsize\citet{end}}} 
  & $50$   & $98.6_{\text{\tiny $\pm$.0}}$ & $84.6_{\text{\tiny $\pm$.1}}$ & $92.7_{\text{\tiny $\pm$.1}}$ & $91.4_{\text{\tiny $\pm$.1}}$ & $1.5_{\text{\tiny $\pm$.1}}$ & $1.9_{\text{\tiny $\pm$.4}}$ & $12.1_{\text{\tiny $\pm$.3}}$ & $0.60_{\text{\tiny $\pm$.00}}$ & $0.46_{\text{\tiny $\pm$.00}}$ \\
  & $100$  & $98.8_{\text{\tiny $\pm$.0}}$ & $87.4_{\text{\tiny $\pm$.2}}$ & $94.1_{\text{\tiny $\pm$.0}}$ & $92.3_{\text{\tiny $\pm$.2}}$ & $1.3_{\text{\tiny $\pm$.0}}$ & $1.8_{\text{\tiny $\pm$.3}}$ & $10.6_{\text{\tiny $\pm$.2}}$ & $0.61_{\text{\tiny $\pm$.00}}$ & $0.46_{\text{\tiny $\pm$.00}}$ \\
  & $250$  & $98.9_{\text{\tiny $\pm$.1}}$ & $88.8_{\text{\tiny $\pm$.5}}$ & $94.7_{\text{\tiny $\pm$.2}}$ & $92.6_{\text{\tiny $\pm$.1}}$ & $1.2_{\text{\tiny $\pm$.1}}$ & $0.8_{\text{\tiny $\pm$.2}}$ & $9.6_{\text{\tiny $\pm$.2}}$ & $0.62_{\text{\tiny $\pm$.00}}$ & $0.46_{\text{\tiny $\pm$.00}}$ \\
  & $500$  & $98.9_{\text{\tiny $\pm$.0}}$ & $88.8_{\text{\tiny $\pm$.4}}$ & $94.8_{\text{\tiny $\pm$.2}}$ & $92.8_{\text{\tiny $\pm$.2}}$ & $1.2_{\text{\tiny $\pm$.1}}$ & $0.8_{\text{\tiny $\pm$.5}}$ & $9.5_{\text{\tiny $\pm$.1}}$ & $0.62_{\text{\tiny $\pm$.00}}$ & $0.46_{\text{\tiny $\pm$.00}}$ \\
  & $1000$ & $98.9_{\text{\tiny $\pm$.0}}$ & $89.1_{\text{\tiny $\pm$.1}}$ & $94.8_{\text{\tiny $\pm$.1}}$ & $92.6_{\text{\tiny $\pm$.2}}$ & $1.2_{\text{\tiny $\pm$.1}}$ & $0.8_{\text{\tiny $\pm$.5}}$ & $9.3_{\text{\tiny $\pm$.1}}$ & $0.62_{\text{\tiny $\pm$.00}}$ & $0.46_{\text{\tiny $\pm$.00}}$ \\
\midrule

\multirow{5}{*}{\makecell[l]{\textsc{EDM*} \\ \scriptsize\citet{end}}} 
  & $50$   & $97.6_{\text{\tiny $\pm$.0}}$ & $77.6_{\text{\tiny $\pm$.5}}$ & $90.2_{\text{\tiny $\pm$.2}}$ & $89.2_{\text{\tiny $\pm$.2}}$ & $4.6_{\text{\tiny $\pm$.1}}$ & $1.7_{\text{\tiny $\pm$.5}}$ & $16.4_{\text{\tiny $\pm$.2}}$ & $0.61_{\text{\tiny $\pm$.00}}$ & $0.46_{\text{\tiny $\pm$.00}}$ \\
  & $100$  & $98.1_{\text{\tiny $\pm$.0}}$ & $81.9_{\text{\tiny $\pm$.4}}$ & $92.1_{\text{\tiny $\pm$.2}}$ & $90.9_{\text{\tiny $\pm$.2}}$ & $3.5_{\text{\tiny $\pm$.1}}$ & $1.4_{\text{\tiny $\pm$.3}}$ & $13.5_{\text{\tiny $\pm$.1}}$ & $0.61_{\text{\tiny $\pm$.00}}$ & $0.46_{\text{\tiny $\pm$.00}}$ \\
  & $250$  & $98.3_{\text{\tiny $\pm$.0}}$ & $84.3_{\text{\tiny $\pm$.1}}$ & $93.2_{\text{\tiny $\pm$.4}}$ & $91.7_{\text{\tiny $\pm$.3}}$ & $2.8_{\text{\tiny $\pm$.2}}$ & $1.3_{\text{\tiny $\pm$.4}}$ & $12.3_{\text{\tiny $\pm$.4}}$ & $0.62_{\text{\tiny $\pm$.00}}$ & $0.46_{\text{\tiny $\pm$.00}}$ \\
  & $500$  & $98.4_{\text{\tiny $\pm$.0}}$ & $85.2_{\text{\tiny $\pm$.5}}$ & $93.5_{\text{\tiny $\pm$.2}}$ & $92.2_{\text{\tiny $\pm$.3}}$ & $2.6_{\text{\tiny $\pm$.2}}$ & $1.3_{\text{\tiny $\pm$.4}}$ & $11.7_{\text{\tiny $\pm$.1}}$ & $0.62_{\text{\tiny $\pm$.00}}$ & $0.46_{\text{\tiny $\pm$.00}}$ \\
  & $1000$ & $98.4_{\text{\tiny $\pm$.0}}$ & $85.3_{\text{\tiny $\pm$.3}}$ & $93.5_{\text{\tiny $\pm$.1}}$ & $91.9_{\text{\tiny $\pm$.1}}$ & $2.5_{\text{\tiny $\pm$.1}}$ & $1.4_{\text{\tiny $\pm$.4}}$ & $11.3_{\text{\tiny $\pm$.1}}$ & $0.62_{\text{\tiny $\pm$.00}}$ & $0.46_{\text{\tiny $\pm$.00}}$ \\
\midrule

\multirow{4}{*}{\makecell[l]{\textsc{GeoBFN} \\ \scriptsize\citet{geobfn}}} 
  & $50$   & $98.3_{\text{\tiny $\pm$.1}}$ & $85.1_{\text{\tiny $\pm$.5}}$ & $92.3_{\text{\tiny $\pm$.4}}$ & $90.7_{\text{\tiny $\pm$.3}}$ & -- & -- & -- & -- & -- \\
  & $100$  & $98.6_{\text{\tiny $\pm$.1}}$ & $87.2_{\text{\tiny $\pm$.3}}$ & $93.0_{\text{\tiny $\pm$.3}}$ & $91.5_{\text{\tiny $\pm$.3}}$ & -- & -- & -- & -- & -- \\
  & $500$  & $98.8_{\text{\tiny $\pm$.8}}$ & $88.4_{\text{\tiny $\pm$.2}}$ & $93.4_{\text{\tiny $\pm$.2}}$ & $91.8_{\text{\tiny $\pm$.2}}$ & -- & -- & -- & -- & -- \\
  & $1000$ & $99.1_{\text{\tiny $\pm$.1}}$ & $90.9_{\text{\tiny $\pm$.2}}$ & $95.3_{\text{\tiny $\pm$.1}}$ & $93.0_{\text{\tiny $\pm$.1}}$ & -- & -- & -- & -- & -- \\
\midrule

\multirow{2}{*}{\makecell[l]{\textsc{VEDA-E} \\ \scriptsize\citet{veda}}} 
  & $30$ & $99.2$ & $91.5$ & $97.0$ & $96.8$ & -- & -- & -- & -- & -- \\
  & $50$ & $99.4$ & $93.7$ & $98.1$ & $97.9$ & -- & -- & -- & -- & -- \\
\midrule

\multirow{2}{*}{\makecell[l]{\textbf{\oursacro}}}
  & $50$   & $99.1_{\text{\tiny $\pm$.0}}$ & $92.3_{\text{\tiny $\pm$.3}}$ & $96.6_{\text{\tiny $\pm$.1}}$ & $88.4_{\text{\tiny $\pm$.2}}$ & $2.41_{\text{\tiny $\pm$.08}}$ & $2.2_{\text{\tiny $\pm$.3}}$ & $13.6_{\text{\tiny $\pm$.5}}$ & $0.70_{\text{\tiny $\pm$.00}}$ & $0.47_{\text{\tiny $\pm$.00}}$ \\
  & $100$  & $99.1_{\text{\tiny $\pm$.1}}$ & $92.6_{\text{\tiny $\pm$.5}}$ & $95.3_{\text{\tiny $\pm$.6}}$ & $86.3_{\text{\tiny $\pm$.9}}$ & $2.24_{\text{\tiny $\pm$.10}}$ & $2.7_{\text{\tiny $\pm$.4}}$ & $10.3_{\text{\tiny $\pm$.4}}$ & $0.70_{\text{\tiny $\pm$.00}}$ & $0.47_{\text{\tiny $\pm$.00}}$ \\
\bottomrule
\end{tabular}
\end{table}
\renewcommand{\arraystretch}{1.0}

We observed no improvement in generation when setting more than $100$ steps in the sampling process, consistent with the evidence that flow models generally require fewer steps than diffusion models \citep{fm}. We used a linear schedule for sampling temperature in the discrete flow, with a minimum of $1.0$ and a maximum of $1.5$. For all \textsc{QM9} experiments, we set the remasking rate $\eta$ \citep{discrete_flows} (Appendix \ref{app:discrete_flows}) in the discrete flow to zero, thus disabling it.

On \textsc{QM9}, we train all models for $1000$ epochs.

\paragraph{\textsc{GEOM-Drugs}} We follow \citet{edm} and opt not to report the molecular stability, neither inferring it via their lookup tables nor via \texttt{OpenBabel}, since the latter introduces significant bias and errors \citep{codesign}. We used a fixed sampling temperature of $1$ in the discrete flow. For \textsc{GEOM-Drugs}, we set the remasking rate in the discrete flow to $\eta = 1.5$.

We train the model for $135$ epochs. The extended results are provided in Table \ref{geom_uncond_extended}.
\renewcommand{\arraystretch}{1.15}
\begin{table}[htbp]
\centering
\small
\setlength{\tabcolsep}{10.2pt}
\caption{Extended results at unconditional generation on \textsc{GEOM-Drugs}. Results for \textsc{VEDA-E} are taken from \citet{veda}, which does not report the remaining metrics. $^\dagger$\citet{equifm} use Dopri5 adaptive solver. Results obtained by \citet{end} on the model of \citet{geoldm} are denoted with $\ddagger$. Bond-explicit methods generate bonds jointly with the 3D structure, and their stability is computed on the generated bonds; results are taken from the original publications, and \textsc{FlowMol-CTMC} \citep{flowmol2} reports neither metric on this dataset.}
\label{geom_uncond_extended}
\begin{tabular}{l c c c c c c}
\toprule
\multirow{2}{*}{Method} & \multirow{2}{*}{Steps} & \multicolumn{1}{c}{Stability} & \multicolumn{1}{c}{V $\times$ C} & \multicolumn{1}{c}{TV ($\downarrow$)} & \multirow{2}{*}{SA ($\uparrow$)} & \multirow{2}{*}{QED ($\uparrow$)} \\ 
\cmidrule(lr){3-3} \cmidrule(lr){4-4} \cmidrule(lr){5-5}
 & & A, \% ($\uparrow$) & \% ($\uparrow$) & A$\left[10^{-2}\right]$ & & \\
\midrule
\makecell[l]{Training Data} & -- & $86.5$ & $99.0$ & -- & $0.83$ & $0.67$ \\
\specialrule{0.7pt}{3pt}{3pt}
\rowcolor{bondgray}\multicolumn{7}{l}{\textit{Bond-explicit methods (for reference; not directly comparable)}} \\
\rowcolor{bondgray}\begin{tabular}[c]{@{}l@{}}\textsc{SemlaFlow} \\ \scriptsize\citet{semlaflow}\end{tabular} & $100$ & $99.8$ & -- & -- & -- & -- \\
\rowcolor{bondgray}\begin{tabular}[c]{@{}l@{}}\textsc{EQGAT-diff} \\ \scriptsize\citet{eqgatdiff}\end{tabular} & $500$ & $99.8$ & -- & -- & -- & -- \\
\rowcolor{bondgray}\begin{tabular}[c]{@{}l@{}}\textsc{JODO} \\ \scriptsize\citet{jodo}\end{tabular} & -- & $100.0$ & $87.4$ & -- & -- & -- \\
\rowcolor{bondgray}\begin{tabular}[c]{@{}l@{}}\textsc{Megalodon} \\ \scriptsize\citet{codesign}\end{tabular} & $500$ & $99.9$ & $92.7$ & -- & -- & -- \\
\specialrule{0.7pt}{3pt}{3pt}

\makecell[l]{\textsc{EDM} \\ \scriptsize\citeauthor{edm} \\ \scriptsize[\citeyear{edm}]}
  & $1000$ & $81.3$ & -- & -- & -- & -- \\
\midrule

\makecell[l]{\textsc{EDM-Bridge} \\ \scriptsize\citet{edm_bridge}}
  & $1000$ & $82.4$ & -- & -- & -- & -- \\
\midrule

\makecell[l]{\textsc{GeoLDM} \\ \scriptsize\citet{geoldm}}
  & $1000$ & $84.4$ & $45.8^{\ddagger}$  & $10.6$ & -- & -- \\
\midrule

\makecell[l]{\textsc{EquiFM} \\ \scriptsize\citet{equifm}}
  & $-^{\dagger}$ & 
  $84.1$ & -- & 
  -- & -- & -- \\
\midrule

\multirow{5}{*}{\makecell[l]{\textsc{END} \\ \scriptsize\citet{end}}} 
  & $50$   & $87.1_{\text{\tiny $\pm$.1}}$ & $66.0_{\text{\tiny $\pm$.4}}$ & $5.9_{\text{\tiny $\pm$.1}}$ &  $0.62_{\text{\tiny $\pm$.00}}$ & $0.48_{\text{\tiny $\pm$.00}}$ \\
  & $100$  & $87.2_{\text{\tiny $\pm$.1}}$ & $73.7_{\text{\tiny $\pm$.4}}$ & $4.5_{\text{\tiny $\pm$.1}}$ &  $0.66_{\text{\tiny $\pm$.00}}$ & $0.55_{\text{\tiny $\pm$.00}}$ \\
  & $250$  & $87.1_{\text{\tiny $\pm$.1}}$ & $77.4_{\text{\tiny $\pm$.4}}$ & $3.5_{\text{\tiny $\pm$.0}}$ &  $0.69_{\text{\tiny $\pm$.00}}$ & $0.58_{\text{\tiny $\pm$.00}}$ \\
  & $500$  & $87.0_{\text{\tiny $\pm$.0}}$ & $78.6_{\text{\tiny $\pm$.3}}$ & $3.3_{\text{\tiny $\pm$.0}}$ &  $0.70_{\text{\tiny $\pm$.00}}$ & $0.59_{\text{\tiny $\pm$.00}}$ \\
  & $1000$ & $87.0_{\text{\tiny $\pm$.0}}$ & $79.4_{\text{\tiny $\pm$.4}}$ & $3.0_{\text{\tiny $\pm$.0}}$ &  $0.70_{\text{\tiny $\pm$.00}}$ & $0.59_{\text{\tiny $\pm$.00}}$ \\
\midrule

\multirow{5}{*}{\makecell[l]{\textsc{EDM*} \\ \scriptsize\citet{end}}} 
  & $50$   & $84.7_{\text{\tiny $\pm$.0}}$ & $46.6_{\text{\tiny $\pm$.3}}$ & $10.5_{\text{\tiny $\pm$.1}}$ &  $0.59_{\text{\tiny $\pm$.00}}$ & $0.48_{\text{\tiny $\pm$.00}}$ \\
  & $100$  & $85.2_{\text{\tiny $\pm$.1}}$ & $56.2_{\text{\tiny $\pm$.4}}$ & $8.0_{\text{\tiny $\pm$.1}}$ &  $0.61_{\text{\tiny $\pm$.00}}$ & $0.53_{\text{\tiny $\pm$.00}}$ \\
  & $250$  & $85.4_{\text{\tiny $\pm$.0}}$ & $61.4_{\text{\tiny $\pm$.6}}$ & $6.7_{\text{\tiny $\pm$.1}}$ &  $0.63_{\text{\tiny $\pm$.00}}$ & $0.56_{\text{\tiny $\pm$.00}}$ \\
  & $500$  & $85.4_{\text{\tiny $\pm$.0}}$ & $63.4_{\text{\tiny $\pm$.1}}$ & $6.4_{\text{\tiny $\pm$.1}}$ &  $0.64_{\text{\tiny $\pm$.00}}$ & $0.57_{\text{\tiny $\pm$.00}}$ \\
  & $1000$ & $85.3_{\text{\tiny $\pm$.1}}$ & $64.2_{\text{\tiny $\pm$.6}}$ & $6.2_{\text{\tiny $\pm$.0}}$ &  $0.64_{\text{\tiny $\pm$.00}}$ & $0.57_{\text{\tiny $\pm$.00}}$ \\
\midrule

\multirow{4}{*}{\makecell[l]{\textsc{GeoBFN} \\ \scriptsize\citet{geobfn}}} 
  & $50$ & $75.1$ & -- & -- & -- & -- \\
  & $100$  & $78.9$ & -- & -- & -- & -- \\
  & $500$  & $81.4$ & -- & -- & -- & -- \\
  & $1000$ & $85.6$ & -- & -- & -- & -- \\
\midrule
\multirow{2}{*}{\makecell[l]{\textsc{VEDA-E} \\ \scriptsize\citet{veda}}} 
  & $50$ & $88.7$ & -- & -- & -- & -- \\
  & $100$  & $90.0$ & -- & -- & -- & -- \\
\midrule
\multirow{2}{*}{\makecell[l]{\textbf{\oursacro}}}
  & $50$ & $94.2_{\text{\tiny $\pm$.0}}$ & $81.3_{\text{\tiny $\pm$.2}}$  & $8.1_{\text{\tiny $\pm$.1}}$ & $0.69_{\text{\tiny $\pm$.00}}$ & $0.72_{\text{\tiny $\pm$.00}}$ \\
  & $100$ & $95.0_{\text{\tiny $\pm$.0}}$ & $81.2_{\text{\tiny $\pm$.3}}$  & $7.4_{\text{\tiny $\pm$.1}}$ & $0.69_{\text{\tiny $\pm$.00}}$ & $0.72_{\text{\tiny $\pm$.00}}$ \\
\bottomrule
\end{tabular}
\end{table}
\renewcommand{\arraystretch}{1.0}

\begin{figure}[ht]
    \centering
    \includegraphics[width=\textwidth]{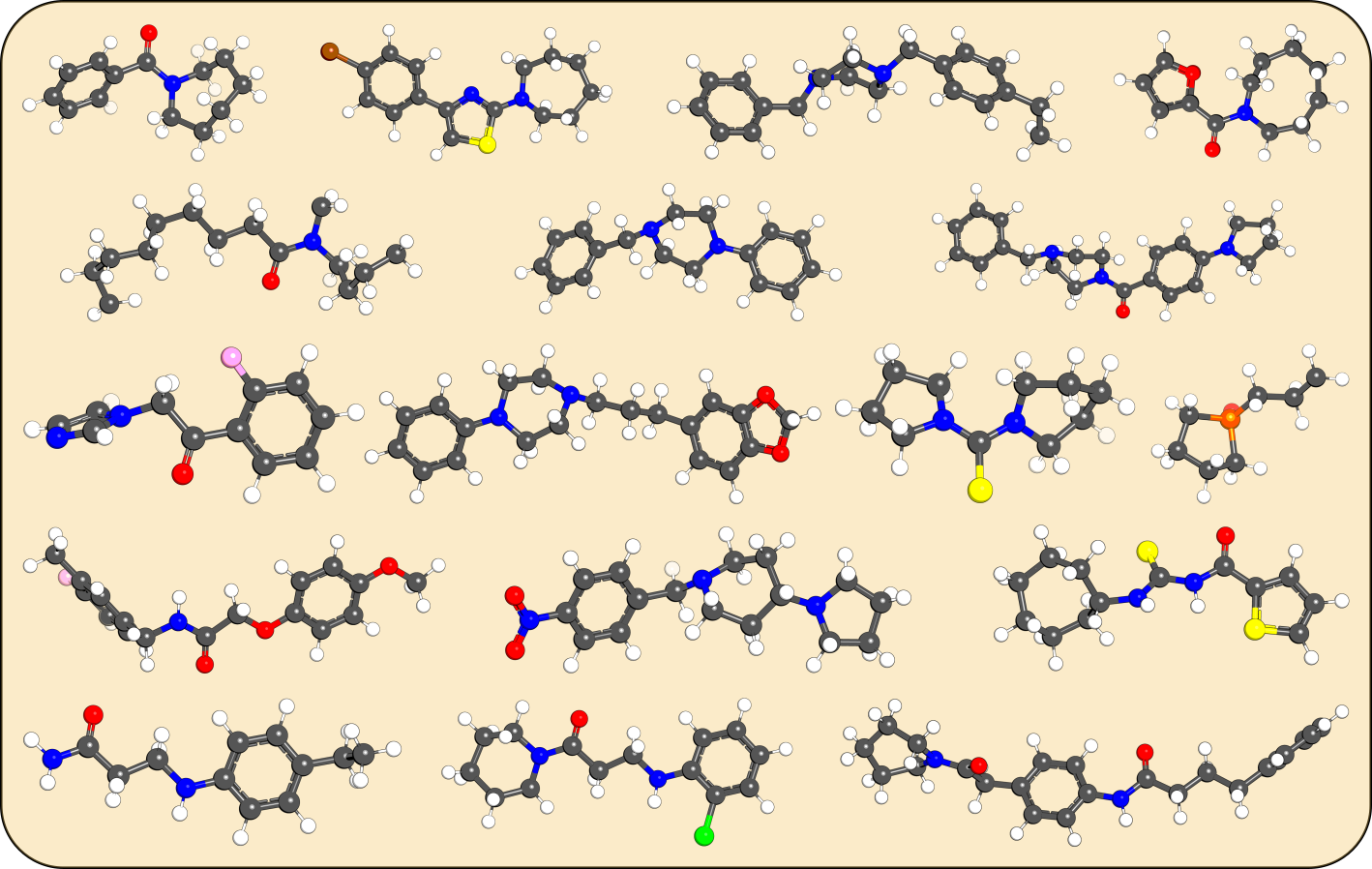} 
    \caption{Randomly selected molecules that are sampled unconditionally from \oursacro trained on \textsc{GEOM-Drugs}.}
    \label{uncond_geom_appendix}
    \vspace{5mm}
    \includegraphics[width=\textwidth]{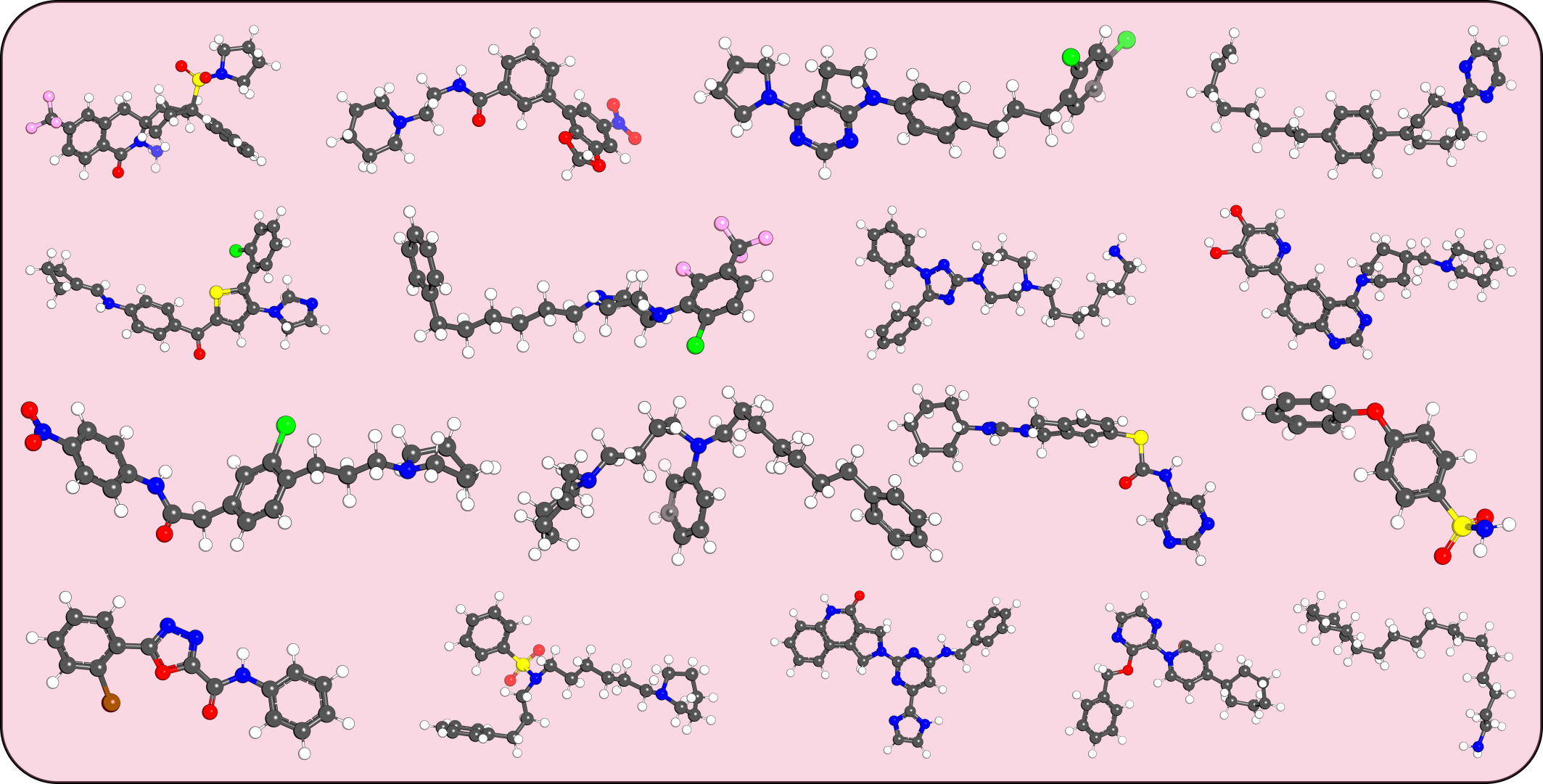} 
    \caption{Randomly selected molecules that are sampled unconditionally from \oursacro trained on \textsc{QMugs}.}
    \label{uncond_qmugs_appendix}
\end{figure}

\paragraph{Uniqueness and Temperature Scheduling}\label{app:uniqueness}
As shown in the main text, \oursacro achieves high molecular stability on \textsc{QM9} but exhibits a drop in uniqueness. This represents a controllable trade-off governed by the sampling temperature in the discrete flow. We performed additional experiments varying the temperature annealing schedule (linear decrease from $T_{\text{start}} \rightarrow T_{\text{end}}$). As shown in Table \ref{tab:qm9_temp}, by adjusting the schedule to $2.0 \rightarrow 1.0$ without changing the granularity of fragmentation ($\alpha = 0.1\%$), \oursacro recovers a highly competitive validity $\times$ uniqueness ($V \times U$) of $91.2 \pm 0.1\%$ while maintaining a molecular stability of $91.4 \pm 0.3\%$, still surpassing all baselines except the concurrent \textsc{VEDA-E} \citep{veda}.

To identify the source of the reduced uniqueness, we grouped the valid molecules from all three runs by their Bemis--Murcko scaffolds \citep{bemis_murcko}. Scaffold uniqueness is $29.7$--$30.5\%$ in the generated samples and $28.7\%$ in reference samples drawn from the training data, while the ten most frequent scaffolds account for $23.1$--$23.9\%$ of generated molecules, compared with $27.4\%$ in the training data; the most frequent generated scaffold is the empty scaffold shared by all acyclic molecules. The generated set therefore retains the scaffold diversity of \textsc{QM9}, and the lost uniqueness stems from repeated complete molecular graphs within this scaffold distribution. The source of these repetitions is visible in the categorical predictions: the most common motif, \texttt{[H][C][H]}, accounts for $22.0\%$ of training motifs, and the model increases its frequency to $31.0$--$31.3\%$. Representing each molecule by the multiset of its motif classes, the probability that two randomly selected molecules share the same multiset is $0.0089\%$ in the training data and $0.0154$--$0.0163\%$ in the generated sets, and $93.2$--$94.8\%$ of the generated molecules whose graph duplicates an earlier sample also share its motif multiset. The categorical model thus overproduces common motif classes. Although a multiset does not uniquely determine a molecular graph, as the bonds between motifs are determined by the generated geometry, with only $6.53$ motifs per \textsc{QM9} molecule on average (compared with $13.1$ on \textsc{GEOM-Drugs} and $15.9$ on \textsc{QMugs}), some of the repeated multisets resolve into identical graphs. Consistent with this explanation, flattening the categorical distribution via the temperature schedule ($1.5 \rightarrow 1.0$ to $2.0 \rightarrow 1.0$) raises uniqueness among valid samples from $90.6\%$ to $96.4\%$ without changing the learned geometric model.

Crucially, this stability-uniqueness trade-off does not apply to medium- and large-scale molecules. On the \textsc{GEOM-Drugs} dataset, \oursacro, similar to all baselines, generates highly diverse structures, achieving uniqueness of $99.92 \pm 0.01\%$.

\begin{table}[ht]
    \centering
    \small
    \caption{Effect of discrete flow temperature scheduling on \textsc{QM9} unconditional generation.}
    \label{tab:qm9_temp}
    \begin{tabular}{@{} l cccc @{}}
    \toprule
    Temperature Schedule & Atom Stable, \% & Molecule Stable, \% & Valid, \% & $V \times U$, \% \\
    \midrule
    $1.0$ (Constant) & $\mathbf{99.3_{\scriptscriptstyle \pm 0.1}}$ & $\mathbf{94.2_{\scriptscriptstyle \pm 0.7}}$ & $\mathbf{96.3_{\scriptscriptstyle \pm 0.4}}$ & $87.2_{\text{\tiny $\pm$0.2}}$ \\
    $1.5 \rightarrow 1.0$ (Original) & $99.1_{\text{\tiny $\pm$0.1}}$ & $92.6_{\text{\tiny $\pm$0.5}}$ & $95.3_{\text{\tiny $\pm$0.6}}$ & $86.3_{\text{\tiny $\pm$0.9}}$ \\
    $2.0 \rightarrow 1.0$ & $99.0_{\text{\tiny $\pm$0.0}}$ & $91.4_{\text{\tiny $\pm$0.3}}$ & $94.6_{\text{\tiny $\pm$0.3}}$ & $\mathbf{91.2_{\scriptscriptstyle \pm 0.1}}$ \\
    $3.0 \rightarrow 1.0$ & $98.5_{\text{\tiny $\pm$0.0}}$ & $87.5_{\text{\tiny $\pm$0.3}}$ & $92.7_{\text{\tiny $\pm$0.5}}$ & $90.4_{\text{\tiny $\pm$0.7}}$ \\
    \bottomrule
    \end{tabular}
\end{table}

\subsubsection{Wall-clock Sampling Time}

To assess how the compressed molecular representation and reduced sampling steps translate into inference speed, we track the wall-clock time required to sample $1024$ molecules from \textsc{QM9} in batches of size $128$. All methods were evaluated on the same \texttt{GTX 1080} GPU. We sampled the molecular size from the marginal distribution, either at the atom level for the baselines or at the fragment level for \textsc{MotiFlow}. Crucially, the measured time for \textsc{MotiFlow} explicitly includes the reconstruction to the atom level, ensuring a fair comparison with atom-based methods. We compare the generation-step configurations reported in the respective papers. The results are in Table \ref{qm9_walltime_results}.

\textsc{EDM} \citep{edm} and \textsc{GeoLDM} \citep{geoldm}, which are discrete-time diffusion models with fixed step configurations, perform comparably. In contrast, \textsc{END} \citep{end} incurs a higher computational cost due to its learnable forward process. The concurrent \textsc{VEDA-E} \citep{veda}, for which we reproduced the published architecture and sampler, is $3.5\times$ slower than \oursacro at $50$ steps, and even with $30$ steps slower than \oursacro with $100$. Overall, \textsc{MotiFlow} achieves the lowest sampling time, with \textsc{GeoBFN} \citep{geobfn} being the closest competitor. However, the performance gap widens for larger molecules: e.g., on \textsc{GEOM-Drugs}, sampling from \textsc{GeoBFN} requires $213.3$ and $437.9$ seconds for $50$ and $100$ steps, respectively. This is more than double the time required by \textsc{MotiFlow}, which completes the task in $100.7$ and $196.2$ seconds, respectively. These results further highlight the efficiency of utilising a rigid-body motif representation for 3D molecular generation.

\renewcommand{\arraystretch}{1.15}
\begin{table}[ht]
\centering
\small
\setlength{\tabcolsep}{15pt} 
\caption{Wall-clock time in seconds for sampling $1024$ molecules on the \textsc{QM9} dataset. $^\dagger$\citet{equifm} use Dopri5 adaptive solver.}
\label{qm9_walltime_results}
\begin{tabular}{@{} l c c @{}}
\toprule
Method & Steps & Time, s ($\downarrow$) \\ 
\midrule

\makecell[l]{\textsc{EDM} \\ \scriptsize\citet{edm}}
  & $1000$ & $1875.9$ \\
\midrule

\multirow{5}{*}{\makecell[l]{\textsc{END} \\ \scriptsize\citet{end}}} 
  & $50$   & $155.2$ \\
  & $100$  & $304.1$ \\
  & $250$  & $784.1$ \\
  & $500$  & $1541.3$ \\
  & $1000$  & $3156.3$ \\
\midrule

\makecell[l]{\textsc{GeoLDM} \\ \scriptsize\citet{geoldm}}
  & $1000$ & $1819.0$ \\
\midrule

\makecell[l]{\textsc{EquiFM} \\ \scriptsize\citet{equifm}}
  & $-^{\dagger}$ & $476.5$ \\
\midrule

\multirow{5}{*}{\makecell[l]{\textsc{GeoBFN} \\ \scriptsize\citet{geobfn}}} 
  & $50$   & $33.9$ \\
  & $100$  & $68.0$ \\
  & $250$  & $171.3$ \\
  & $500$  & $357.4$ \\
  & $1000$ & $685.5$ \\
\midrule

\multirow{2}{*}{\makecell[l]{\textsc{VEDA-E} \\ \scriptsize\citet{veda}}} 
  & $30$   & $62.6$ \\
  & $50$   & $102.3$ \\
\midrule

\multirow{2}{*}{\makecell[l]{\textsc{MotiFlow}}}
  & $50$  & $29.4$ \\
  & $100$ & $56.1$ \\
\bottomrule
\end{tabular}
\end{table}
\renewcommand{\arraystretch}{1.0}

\subsubsection{Modality Scheduling Analysis}
\label{app:modality_scheduling}

To examine the interplay between the discrete and $\mathrm{SE}(3)$ flows, we performed additional experiments decoupling the two components during training. Our baseline \oursacro model evolves both modalities concurrently, using a uniformly linear time $t$ directly in the Euclidean and geodesic paths and the discrete masking path $p_t(m_t \mid m_1)$.

To explicitly schedule the modalities, we follow recent mixed discrete-continuous models \citep{midi, flowmol} and replace $t$ in the interpolants with a modality-specific temporal schedule $\alpha_t$ of the form:
\begin{equation}\label{eq:modality_schedule}
    \alpha_t(\nu) = 1 - \cos^2 \left(\frac{\pi}{2}t^\nu\right) = \sin^2 \left(\frac{\pi}{2}t^\nu\right).
\end{equation}
Here, $\nu > 0$ in \eqref{eq:modality_schedule} controls the transition speed. A larger $\nu$ (e.g., $\nu=2$) causes $t^\nu$ to grow more slowly initially, meaning the transition from the prior to the data distribution is delayed, and the state remains noisy or masked for a longer duration. We evaluate two decoupled configurations on \textsc{GEOM-Drugs} ($100$ sampling steps). The results are presented in Table \ref{tab:modality_schedule}.
\begin{table}[ht]
    \centering
    \small
    \caption{Effect of decoupled modality scheduling on \textsc{GEOM-Drugs} ($100$ steps).}
    \label{tab:modality_schedule}
    \begin{tabular}{@{} l cccc @{}}
    \toprule
    Schedule & Atom Stability, \% & Validity $V$, \% & Connectivity $C$, \% & $V \times C$, \% \\
    \midrule
    (1) $\{\nu_{\text{motif}} = 1, \nu_{\mathrm{SE}(3)} = 2\}$ & $\mathbf{97.3_{\scriptscriptstyle \pm .0}}$ & $86.2_{\text{\tiny $\pm$.3}}$ & $72.4_{\text{\tiny $\pm$.9}}$ & $62.4_{\text{\tiny $\pm$1.0}}$ \\
    (2) $\{\nu_{\text{motif}} = 2, \nu_{\mathrm{SE}(3)} = 1\}$ & $92.7_{\text{\tiny $\pm$.0}}$ & $77.9_{\text{\tiny $\pm$.3}}$ & $\mathbf{96.0_{\scriptscriptstyle \pm .2}}$ & $74.8_{\text{\tiny $\pm$0.4}}$ \\
    \midrule
    \oursacro (Baseline) & $95.0_{\text{\tiny $\pm$.0}}$ & $\mathbf{87.2_{\scriptscriptstyle \pm .3}}$ & $93.1_{\text{\tiny $\pm$.2}}$ & $\mathbf{81.2_{\scriptscriptstyle \pm 0.3}}$ \\
    \bottomrule
    \end{tabular}
\end{table}
In Schedule (1), motif identities resolve faster than the geometry. Because they are established early, they yield highly stable local chemistry; however, the delayed $\mathrm{SE}(3)$ flow subsequently struggles to merge these fixed, disparate rigid bodies into a single molecule, resulting in low connectivity. Conversely, in Schedule (2), the geometry resolves faster. The $\mathrm{SE}(3)$ flow successfully connects abstract masked shapes early (high connectivity), but the categorical flow is then forced to assign discrete classes to a pre-existing 3D layout, causing valency mismatches and spatial clashes (lower stability and validity).

Our concurrent, uniformly linear evolution in \oursacro avoids both failure modes via self-conditioning \citep{harm_self_cond}: motif geometry is continuously guided by the current estimate of discrete identities at each step, allowing local chemistry and global topology to mutually adapt.

\subsubsection{Conditional Generation}

\begin{figure}[ht]
    \centering
    \includegraphics[width=0.65\columnwidth]{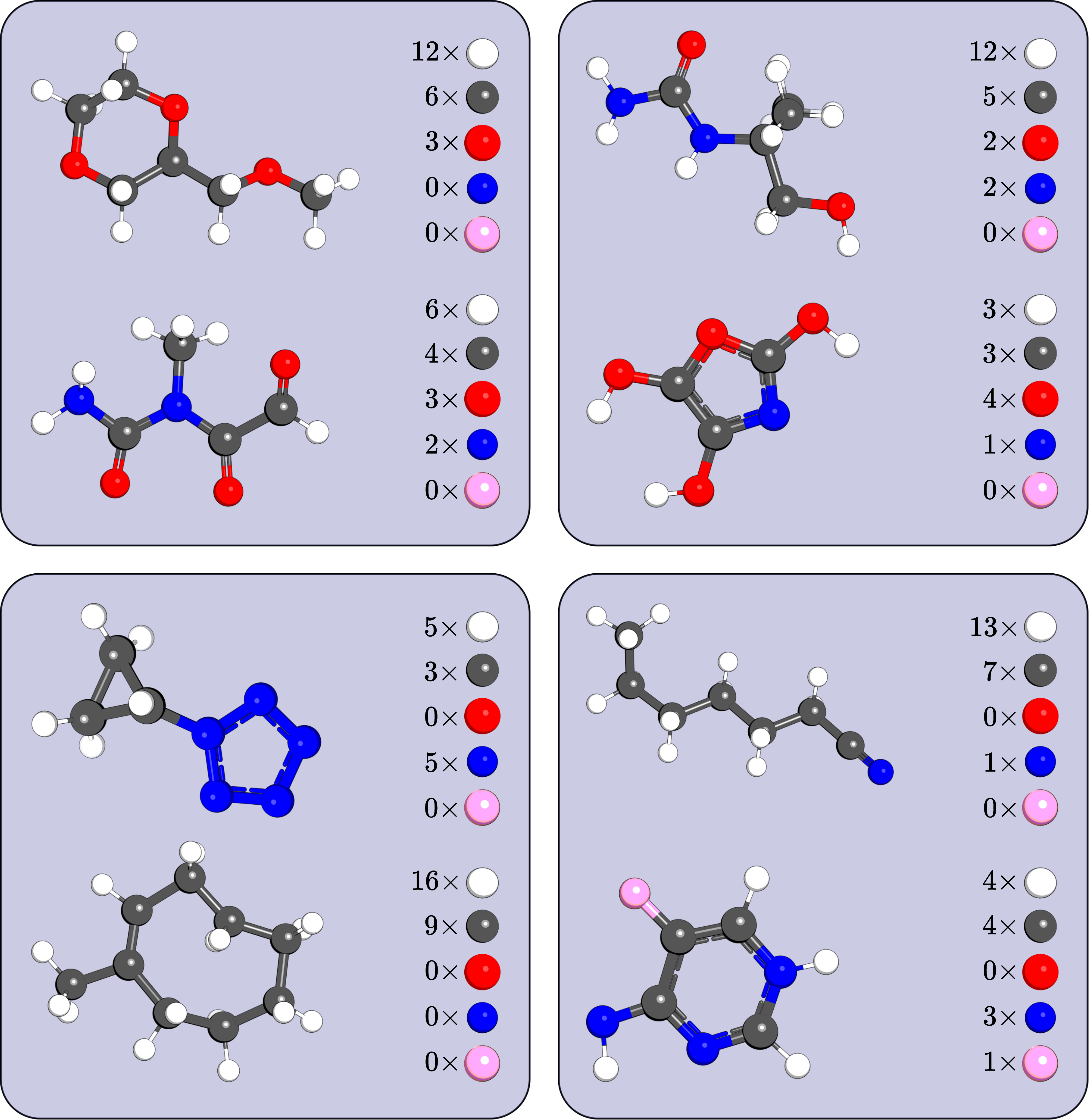}
    \caption{Randomly chosen samples from the atom composition conditional task.}
    \label{composition_appendix}
    
    \vspace{2mm}
    
    \includegraphics[width=0.65\columnwidth]{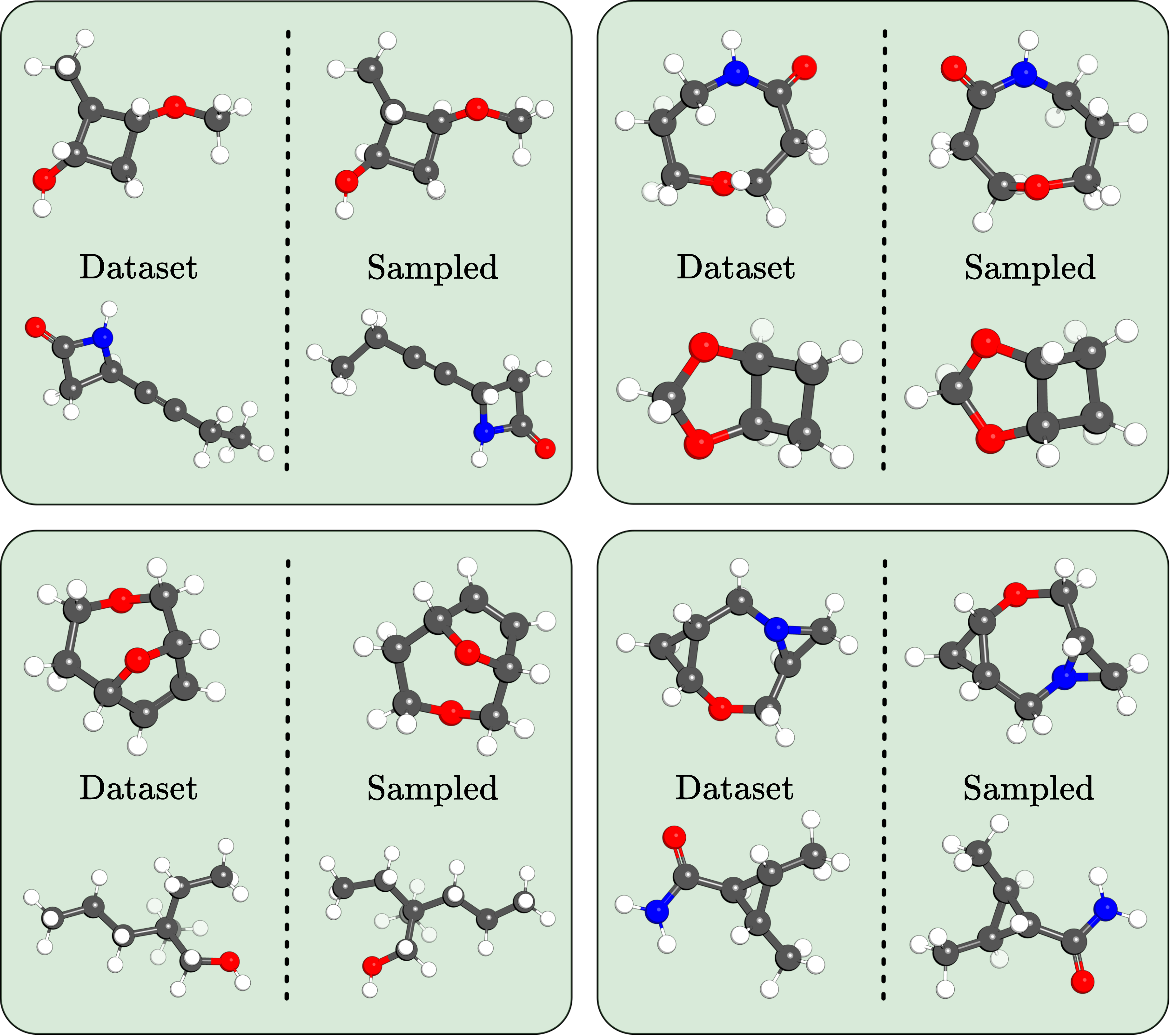}
    \caption{Randomly chosen samples from the fingerprint substructure conditional task.}
    \label{substructure_appendix}
\end{figure}

Results on the atom composition and substructural features conditioning with additional generation steps configurations are reflected in Table \ref{cond_gen_table_extended}. Figures \ref{composition_appendix} and \ref{substructure_appendix} present instances of conditionally sampled molecules for both tasks.

\renewcommand{\arraystretch}{1.15}
\begin{table}[ht]
\centering
\small
\setlength{\tabcolsep}{6pt}
\caption{Results on atom composition and substructural features conditional generation tasks.}
\label{cond_gen_table_extended}
\begin{tabular}{@{} l c c c @{}}
\toprule
\multirow{2}{*}{Method} & \multirow{2}{*}{Steps} & \multicolumn{1}{c}{\textsc{Composition}} & \multicolumn{1}{c}{\textsc{Substructure}} \\ 
\cmidrule(lr){3-3} \cmidrule(lr){4-4}
 & & Matching, $\%$ ($\uparrow$) & Tanimoto Sim. ($\uparrow$) \\
\midrule

\makecell[l]{\textsc{cEDM} \\ \scriptsize\citet{eegsde}}
  & $1000$ & -- & $.671_{\text{\tiny $\pm$.004}}$ \\
\midrule

\makecell[l]{\textsc{EEGSDE} \\ \scriptsize\citet{eegsde}}
  & $1000$ & -- & $.750_{\text{\tiny $\pm$.003}}$ \\
\midrule

\multirow{5}{*}{\makecell[l]{\textsc{cEDM}* \\ \scriptsize\citet{end}}} 
  & $50$   & $69.6_{\text{\tiny $\pm$0.6}}$ & $.601_{\text{\tiny $\pm$.000}}$ \\
  & $100$  & $73.0_{\text{\tiny $\pm$0.6}}$ & $.640_{\text{\tiny $\pm$.002}}$ \\
  & $250$  & $74.1_{\text{\tiny $\pm$1.4}}$ & $.663_{\text{\tiny $\pm$.002}}$ \\
  & $500$  & $76.2_{\text{\tiny $\pm$0.6}}$ & $.669_{\text{\tiny $\pm$.001}}$ \\
  & $1000$ & $75.5_{\text{\tiny $\pm$0.5}}$ & $.673_{\text{\tiny $\pm$.002}}$ \\
\midrule

\multirow{5}{*}{\makecell[l]{\textsc{cEND} \\ \scriptsize\citet{end}}} 
  & $50$   & $89.2_{\text{\tiny $\pm$0.8}}$ & $.783_{\text{\tiny $\pm$.001}}$ \\
  & $100$  & $90.1_{\text{\tiny $\pm$1.0}}$ & $.807_{\text{\tiny $\pm$.001}}$ \\
  & $250$  & $91.2_{\text{\tiny $\pm$0.8}}$ & $.819_{\text{\tiny $\pm$.001}}$ \\
  & $500$  & $91.5_{\text{\tiny $\pm$0.8}}$ & $.825_{\text{\tiny $\pm$.001}}$ \\
  & $1000$ & $91.0_{\text{\tiny $\pm$0.9}}$ & $.828_{\text{\tiny $\pm$.001}}$ \\
\midrule

\multirow{2}{*}{\makecell[l]{\textsc{c}\oursacro}}
  & $50$   & $92.2_{\text{\tiny $\pm$0.4}}$ & $.830_{\text{\tiny $\pm$.001}}$ \\
  & $100$  & $95.4_{\text{\tiny $\pm$0.5}}$ & $.862_{\text{\tiny $\pm$.002}}$ \\
\bottomrule
\end{tabular}
\end{table}
\renewcommand{\arraystretch}{1.0}

\subsubsection{Fragmentation Ablation Details} \label{Fragmentation Ablation Details}

\paragraph{Fragmentation statistics} In Section \ref{Ablations}, we compare our proposed \textsc{Planar Rings} decomposition strategy against a fine-grained \textsc{No Rings} variant and across different occurrence thresholds $\alpha$. Here, we provide detailed statistics on these decompositions for the training subsets of \textsc{GEOM-Drugs} and \textsc{QMugs}.

\textbf{Dataset properties}: the general statistics of the molecules in the training splits used for the ablation studies are summarised in Table \ref{tab:dataset_stats}. \textsc{QMugs} contains generally larger and heavier molecules than \textsc{GEOM-Drugs}.

\begin{table}[ht]
    \centering
    \small
    \caption{Statistics of the training subsets of the datasets used in the ablation study.}
    \label{tab:dataset_stats}
    \begin{tabular}{@{}l ccc ccc@{}}
    \toprule
    \multirow{2}{*}{Dataset} & \multicolumn{3}{c}{Total Atoms} & \multicolumn{3}{c}{Heavy Atoms} \\
    \cmidrule(lr){2-4} \cmidrule(l){5-7}
     & Mean & Median & Max & Mean & Median & Max \\
    \midrule
    \textsc{GEOM-Drugs} & $45.2$ & $45.0$ & $143$ & $25.3$ & $25.0$ & $77$ \\
    \textsc{QMugs}      & $53.6$ & $52.0$ & $178$ & $29.9$ & $29.0$ & $96$ \\
    \bottomrule
    \end{tabular}
\end{table}

\textbf{Fragmentation settings}: the \textsc{No Rings} strategy decomposes molecules by cutting all bonds except those to hydrogens and double/triple bonds, effectively breaking rings. For the \textsc{Planar Rings} strategy, we vary the hyperparameter $\alpha$, reported as a percentage of the total dataset size. This threshold determines the minimum absolute occurrence count required for a motif to be retained in the vocabulary $\mathcal{V}_m$; motifs appearing less frequently are recursively decomposed. 

Table \ref{tab:frag_stats} details the resulting fragment statistics. We observe that lower $\alpha$ thresholds (e.g., $0.01\%$) lead to a larger vocabulary with larger motifs up to 23 atoms and fewer fragments per molecule. Conversely, the \textsc{No Rings} baseline yields the highest number of fragments per molecule with the smallest maximum fragment size. Note that the fragment size includes necessary dummy atoms added to lock the coordinate frames of collinear or single-atom motifs. Across all configurations and datasets, the maximum size of the discrete symmetry group $|\mathcal{S}_i|$ encountered for any motif was 12.

\begin{table}[ht]
    \centering
    \small
    \setlength{\tabcolsep}{4.8pt}
    \caption{Detailed statistics of the ablated rigid-motif decomposition strategies. The \textit{Cut-off} column indicates the absolute number of occurrences required for a motif to be preserved in the vocabulary. \textit{Max. size} denotes the maximum number of atoms, including dummy ones, in a single motif.}
    \label{tab:frag_stats}
    \begin{tabular}{@{}l l c ccc c@{}}
    \toprule
    \multirow{2}{*}{Dataset} & \multirow{2}{*}{Fragmentation} & \multirow{2}{*}{Cut-off} & \multicolumn{3}{c}{Fragments per molecule} & \multirow{2}{*}{Max. size} \\
    \cmidrule(lr){4-6}
     & & & Mean & Median & Max & \\
    \midrule
    \multirow{4}{*}{\textsc{GEOM-Drugs}} 
      & \textsc{No Rings} & -- & $16.8$ & $17.0$ & $57$ & $6$ \\
      & \textsc{Planar Rings} \scriptsize $0.5\%$ & $964$ & $13.7$ & $13.0$ & $57$ & $17$ \\
      & \textsc{Planar Rings} \scriptsize $0.1\%$  & $192$ & $13.1$ & $13.0$ & $57$ & $17$ \\
      & \textsc{Planar Rings} \scriptsize $0.01\%$ & $19$ & $12.7$ & $12.0$ & $57$ & $23$ \\
    \midrule
    \multirow{4}{*}{\textsc{QMugs}} 
      & \textsc{No Rings} & -- & $19.9$ & $19.0$ & $80$ & $6$ \\
      & \textsc{Planar Rings} \scriptsize $0.5\%$  & $1419$ & $16.7$ & $16.0$ & $78$ & $17$ \\
      & \textsc{Planar Rings} \scriptsize $0.1\%$  & $283$ & $15.9$ & $15.0$ & $78$ & $17$ \\
      & \textsc{Planar Rings} \scriptsize $0.01\%$ & $28$ & $15.3$ & $15.0$ & $78$ & $23$ \\
    \bottomrule
    \end{tabular}
\end{table}

We provide the extended results of our ablation study on the unconditional performance of different fragmentation methods in Table \ref{ablation_extended}. The training on \textsc{GEOM-Drugs} is done for $70$ epochs, and for $50$ epochs on \textsc{QMugs}.  Qualitative samples are shown in Figures \ref{uncond_geom_appendix} and \ref{uncond_qmugs_appendix}.

\renewcommand{\arraystretch}{1.15}
\begin{table}[ht]
\centering
\small
\setlength{\tabcolsep}{0.9pt}
\caption{Extended ablation results for the unconditional generation on \textsc{GEOM-Drugs} and \textsc{QMugs} for various fragmentation strategies.}
\label{ablation_extended}
\begin{tabular}{@{} l c ccccc ccccc @{}}
\toprule
\multirow{3}{*}{\makecell[l]{Fragment. \\ Strategy}} & \multirow{3}{*}{Steps} & \multicolumn{5}{c}{\textsc{QMugs}} & \multicolumn{5}{c}{\textsc{GEOM-Drugs}} \\ 
\cmidrule(lr){3-7} \cmidrule(l){8-12}
 & & \multicolumn{1}{c}{Stabil.} & \multicolumn{1}{c}{V $\times$ C} & \multirow{2}{*}{SA ($\uparrow$)} & \multirow{2}{*}{QED ($\uparrow$)} & \multicolumn{1}{c}{TV ($\downarrow$)} & \multicolumn{1}{c}{Stabil.} & \multicolumn{1}{c}{V $\times$ C} & \multirow{2}{*}{SA ($\uparrow$)} & \multirow{2}{*}{QED ($\uparrow$)} & \multicolumn{1}{c}{TV ($\downarrow$)} \\
 \cmidrule(lr){3-4} \cmidrule(lr){7-7} \cmidrule(lr){8-9} \cmidrule(l){12-12}
 & & A, \% ($\uparrow$) & \% ($\uparrow$) & & & A $\left[10^{-2}\right]$ & A, \% ($\uparrow$) & \% ($\uparrow$) & & & A$\left[10^{-2}\right]$\\
\midrule

\multirow{2}{*}{{\textsc{No Rings}}}
  & $50$ & $86.5_{\text{\tiny $\pm$0.6}}$ & $83.0_{\text{\tiny $\pm$.2}}$ & $0.74_{\text{\tiny $\pm$.00}}$ & $0.51_{\text{\tiny $\pm$.00}}$ & $5.12_{\text{\tiny $\pm$.06}}$ & $89.1_{\text{\tiny $\pm$0.7}}$ & $80.9_{\text{\tiny $\pm$.2}}$ & $0.71_{\text{\tiny $\pm$.00}}$ & $0.69_{\text{\tiny $\pm$.00}}$ & $3.57_{\text{\tiny $\pm$.05}}$ \\
  & $100$ & $85.5_{\text{\tiny $\pm$1.8}}$ & $85.6_{\text{\tiny $\pm$.3}}$ & $0.74_{\text{\tiny $\pm$.00}}$ & $0.50_{\text{\tiny $\pm$.00}}$ & $4.98_{\text{\tiny $\pm$.06}}$ & $86.7_{\text{\tiny $\pm$1.8}}$ & $82.9_{\text{\tiny $\pm$.4}}$ & $0.71_{\text{\tiny $\pm$.00}}$ & $0.69_{\text{\tiny $\pm$.00}}$ & $3.36_{\text{\tiny $\pm$.08}}$ \\
\midrule

\multirow{2}{*}{\makecell[l]{\textsc{Planar} \\ \textsc{Rings} \scriptsize $0.5\%$}}
  & $50$  & $95.1_{\text{\tiny $\pm$0.1}}$ & $82.2_{\text{\tiny $\pm$.8}}$ & $0.72_{\text{\tiny $\pm$.00}}$ & $0.60_{\text{\tiny $\pm$.00}}$ & $8.56_{\text{\tiny $\pm$.04}}$ & $94.9_{\text{\tiny $\pm$0.0}}$ & $81.3_{\text{\tiny $\pm$.4}}$ & $0.70_{\text{\tiny $\pm$.00}}$ & $0.71_{\text{\tiny $\pm$.00}}$ & $4.47_{\text{\tiny $\pm$.02}}$ \\
  & $100$ & $95.5_{\text{\tiny $\pm$0.1}}$ & $84.0_{\text{\tiny $\pm$.5}}$ & $0.72_{\text{\tiny $\pm$.00}}$ & $0.60_{\text{\tiny $\pm$.00}}$ & $8.08_{\text{\tiny $\pm$.17}}$ & $95.2_{\text{\tiny $\pm$0.1}}$ & $82.1_{\text{\tiny $\pm$.3}}$ & $0.70_{\text{\tiny $\pm$.00}}$ & $0.71_{\text{\tiny $\pm$.00}}$ & $4.24_{\text{\tiny $\pm$.14}}$ \\
\midrule

\multirow{2}{*}{\makecell[l]{\textsc{Planar} \\ \textsc{Rings} \scriptsize $0.1\%$}}
  & $50$  & $95.7_{\text{\tiny $\pm$0.0}}$ & $81.6_{\text{\tiny $\pm$.3}}$ & $0.72_{\text{\tiny $\pm$.00}}$ & $0.58_{\text{\tiny $\pm$.00}}$ & $8.47_{\text{\tiny $\pm$.08}}$ & $94.8_{\text{\tiny $\pm$0.0}}$ & $79.6_{\text{\tiny $\pm$.5}}$ & $0.70_{\text{\tiny $\pm$.00}}$ & $0.68_{\text{\tiny $\pm$.00}}$ & $4.45_{\text{\tiny $\pm$.12}}$ \\
  & $100$ & $96.1_{\text{\tiny $\pm$0.0}}$ & $83.1_{\text{\tiny $\pm$.4}}$ & $0.72_{\text{\tiny $\pm$.00}}$ & $0.58_{\text{\tiny $\pm$.00}}$ & $8.01_{\text{\tiny $\pm$.22}}$ & $95.0_{\text{\tiny $\pm$0.0}}$ & $81.2_{\text{\tiny $\pm$.3}}$ & $0.69_{\text{\tiny $\pm$.00}}$ & $0.68_{\text{\tiny $\pm$.00}}$ & $3.93_{\text{\tiny $\pm$.09}}$ \\
\midrule

\multirow{2}{*}{\makecell[l]{\textsc{Planar} \\ \textsc{Rings} \scriptsize $0.01\%$}}
  & $50$  & $95.1_{\text{\tiny $\pm$0.0}}$ & $78.8_{\text{\tiny $\pm$.3}}$ & $0.72_{\text{\tiny $\pm$.00}}$ & $0.49_{\text{\tiny $\pm$.01}}$ & $11.62_{\text{\tiny $\pm$.11}}$ & $96.2_{\text{\tiny $\pm$0.0}}$ & $81.3_{\text{\tiny $\pm$.2}}$ & $0.69_{\text{\tiny $\pm$.00}}$ & $0.72_{\text{\tiny $\pm$.00}}$ & $8.07_{\text{\tiny $\pm$.06}}$ \\
  & $100$ & $95.3_{\text{\tiny $\pm$0.0}}$ & $80.4_{\text{\tiny $\pm$.5}}$ & $0.73_{\text{\tiny $\pm$.00}}$ & $0.49_{\text{\tiny $\pm$.00}}$ & $11.10_{\text{\tiny $\pm$.03}}$ & $96.3_{\text{\tiny $\pm$0.0}}$ & $82.4_{\text{\tiny $\pm$.3}}$ & $0.69_{\text{\tiny $\pm$.00}}$ & $0.72_{\text{\tiny $\pm$.00}}$ & $7.43_{\text{\tiny $\pm$.05}}$ \\
\bottomrule
\end{tabular}
\end{table}
\renewcommand{\arraystretch}{1.0}

\subsubsection{Vocabulary Statistics and Distributions}
\label{app:vocab_stats}
To further detail the vocabularies extracted by our fragmentation strategy, we analyse the distributions and frequencies of the resulting motifs on the \textsc{GEOM-Drugs} dataset.

\begin{figure}[ht]
    \centering
    \begin{minipage}[t]{0.48\textwidth}
        \vspace{0pt}
        \centering
        \includegraphics[width=\linewidth]{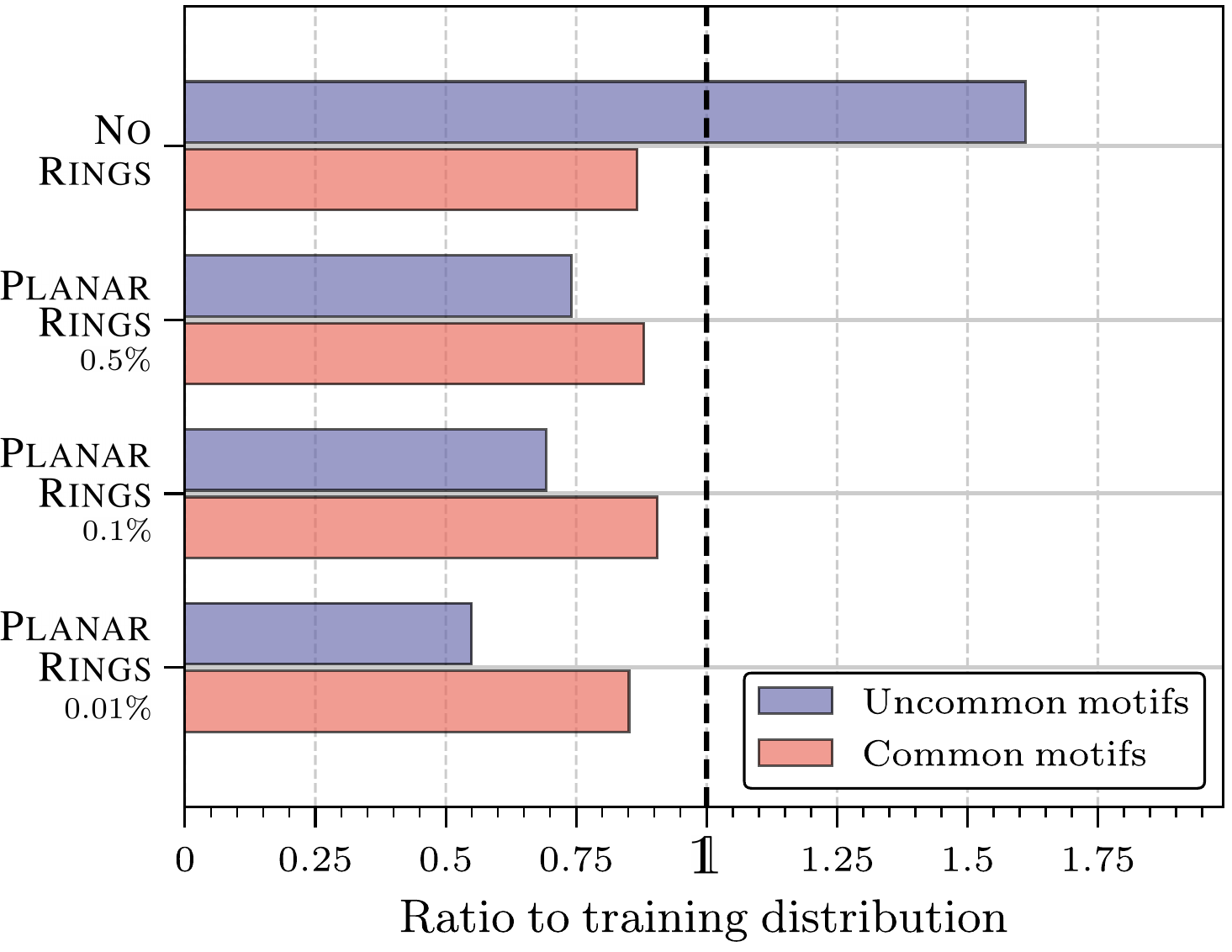} 
        \caption{Comparison of sampled motif types to their frequencies in the training distribution of \textsc{GEOM-Drugs}. A ratio of 1 is optimal.}
        \label{fig:geom_motif_distr}
    \end{minipage}\hfill
    \begin{minipage}[t]{0.48\textwidth}
        \vspace{0pt}
        \centering
        \includegraphics[width=\linewidth]{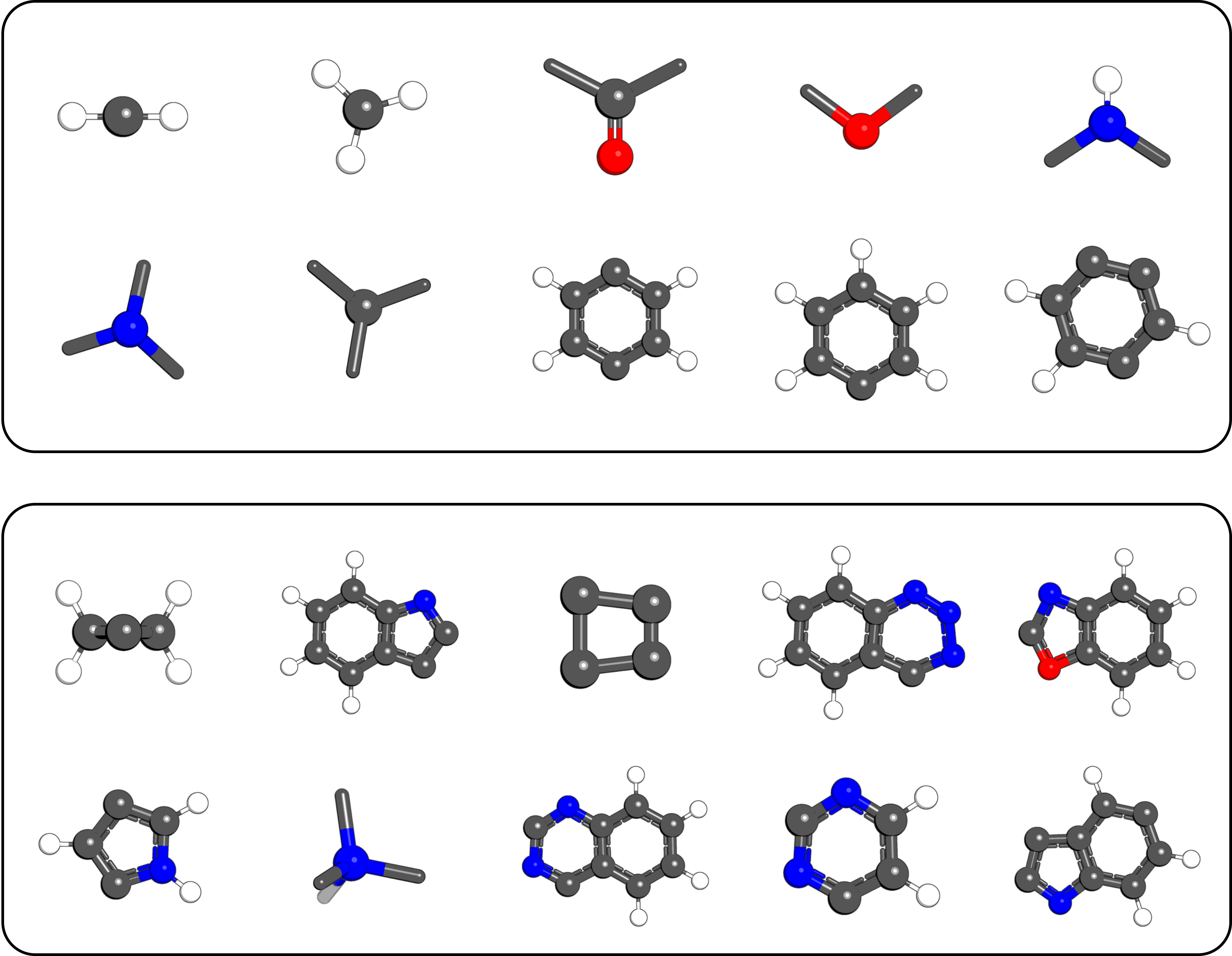}
        \caption{The 10 most frequent (\textit{top}) and rarest (\textit{bottom}) motifs in \textsc{GEOM-Drugs} under the default \textsc{Planar Rings} $\alpha=0.1\%$ fragmentation strategy.}
        \label{fig:geom_motif_occur}
    \end{minipage}
\end{figure}

First, Figure \ref{fig:geom_motif_distr} replicates the motif-sampling analysis from Section \ref{Ablations} (Figure \ref{motif_distr_qmugs}) for \textsc{GEOM-Drugs}. The results mirror the previous findings: common motifs are sampled at close-to-true occurrence (ratios $\approx 1.0$), whereas uncommon motifs are oversampled under the \textsc{No Rings} strategy. As the occurrence threshold $\alpha$ is lowered under the \textsc{Planar Rings} strategy, rare motifs are progressively undersampled. This indicates that the fragmentation trade-off generalises across datasets.

Figure \ref{fig:geom_motif_occur} shows the 10 most and least frequent motifs under the default $\alpha=0.1\%$ setup. Frequent motifs (\textit{top row}) include basic structural units such as methyl groups, carbonyl oxygens, aliphatic links, and benzene rings. Rare motifs (\textit{bottom row}) primarily consist of larger multi-heteroatom fused rings. Treating these structures as predefined rigid frames avoids the need for the continuous flow to reconstruct stable cyclic geometries atom-by-atom.

\begin{figure}[ht]
    \centering
    
    \begin{subfigure}{0.48\textwidth}
        \centering
        \includegraphics[width=\linewidth]{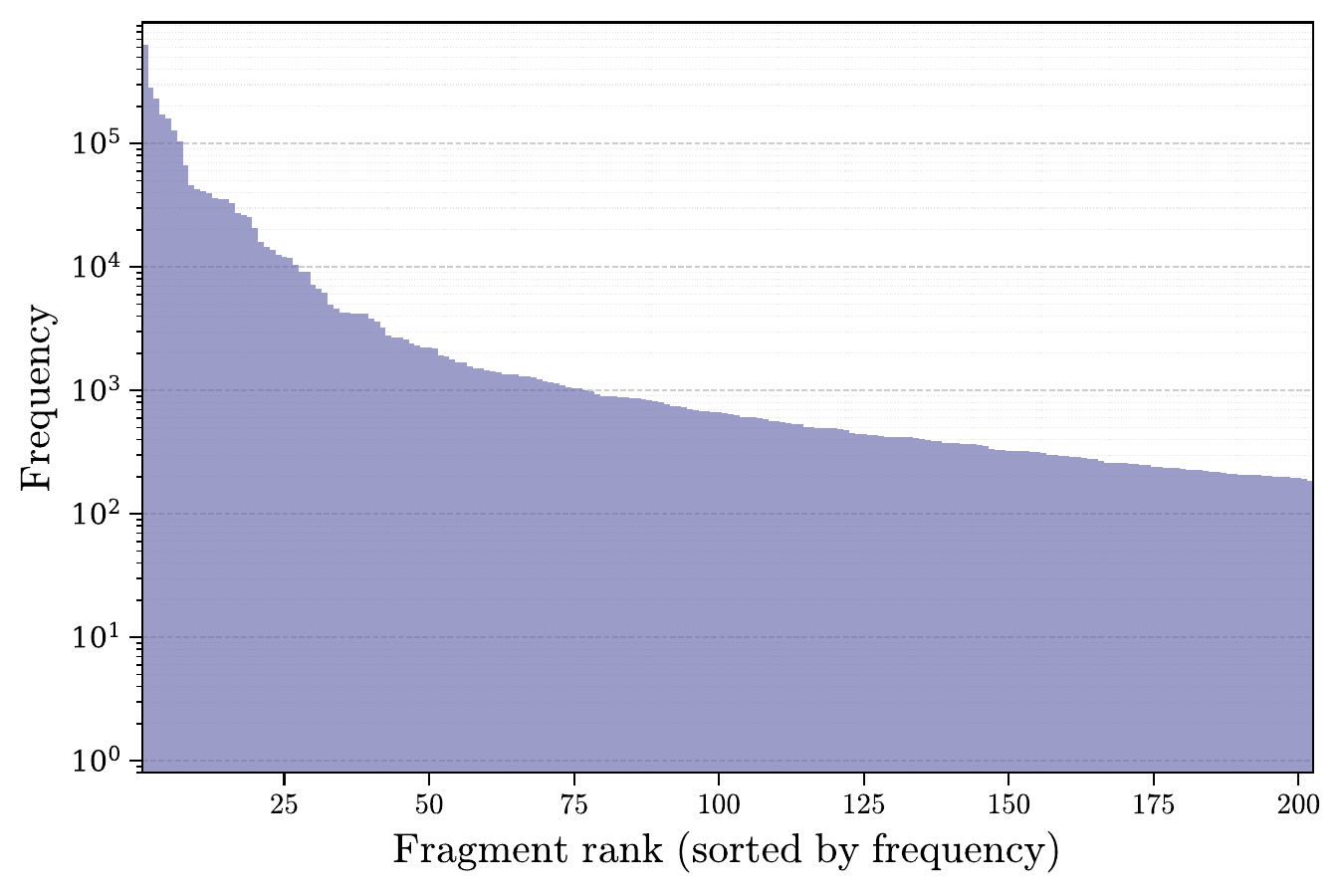} 
        \caption{Frequency Rank: \textsc{Planar Rings} $0.1\%$}
        \label{fig:freq_rank_01}
    \end{subfigure}\hfill
    \begin{subfigure}{0.48\textwidth}
        \centering
        \includegraphics[width=\linewidth]{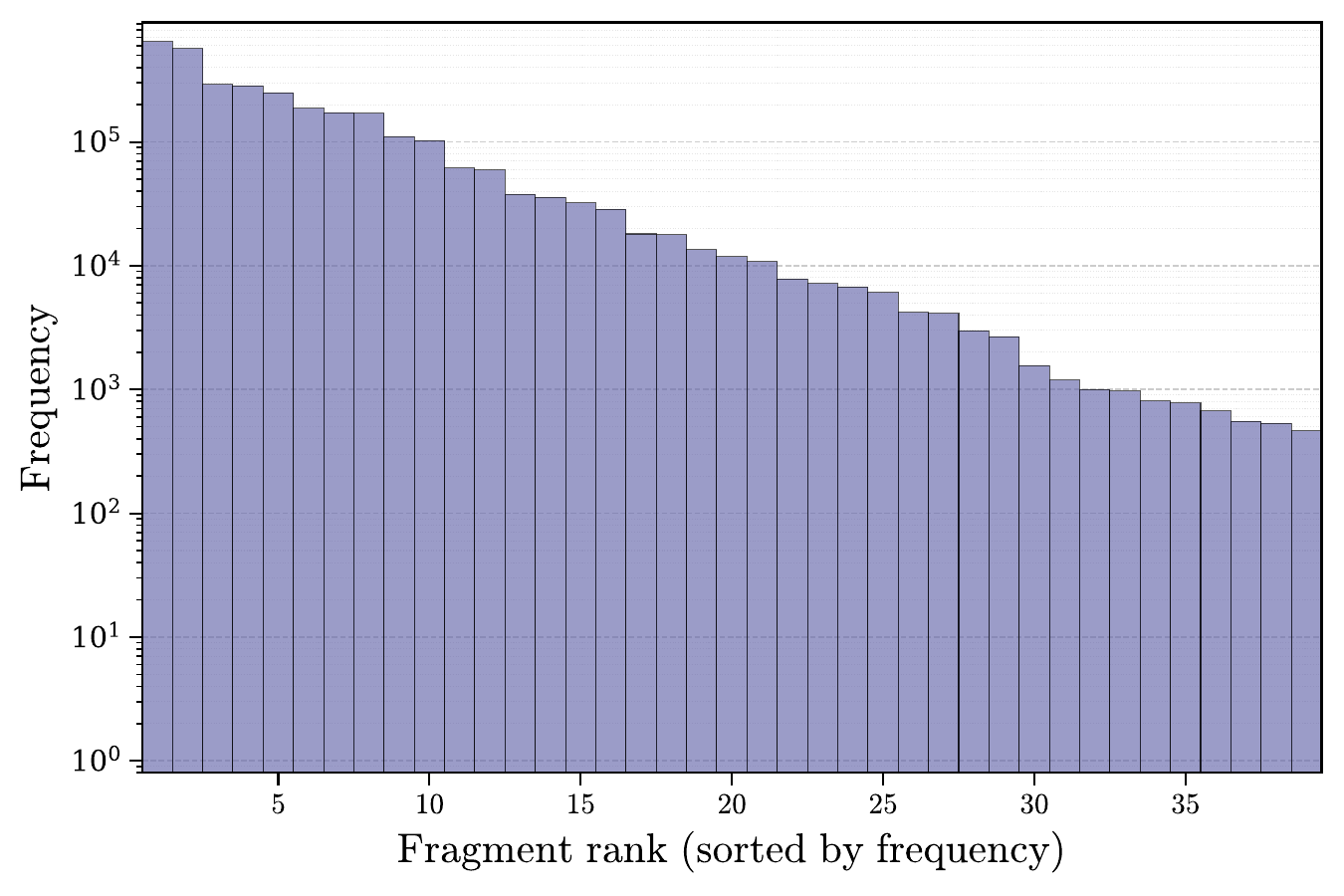} 
        \caption{Frequency Rank: \textsc{No Rings}}
        \label{fig:freq_rank_norings}
    \end{subfigure}
    
    \vspace{4mm}
    
    \begin{subfigure}{0.48\textwidth}
        \centering
        \includegraphics[width=\linewidth]{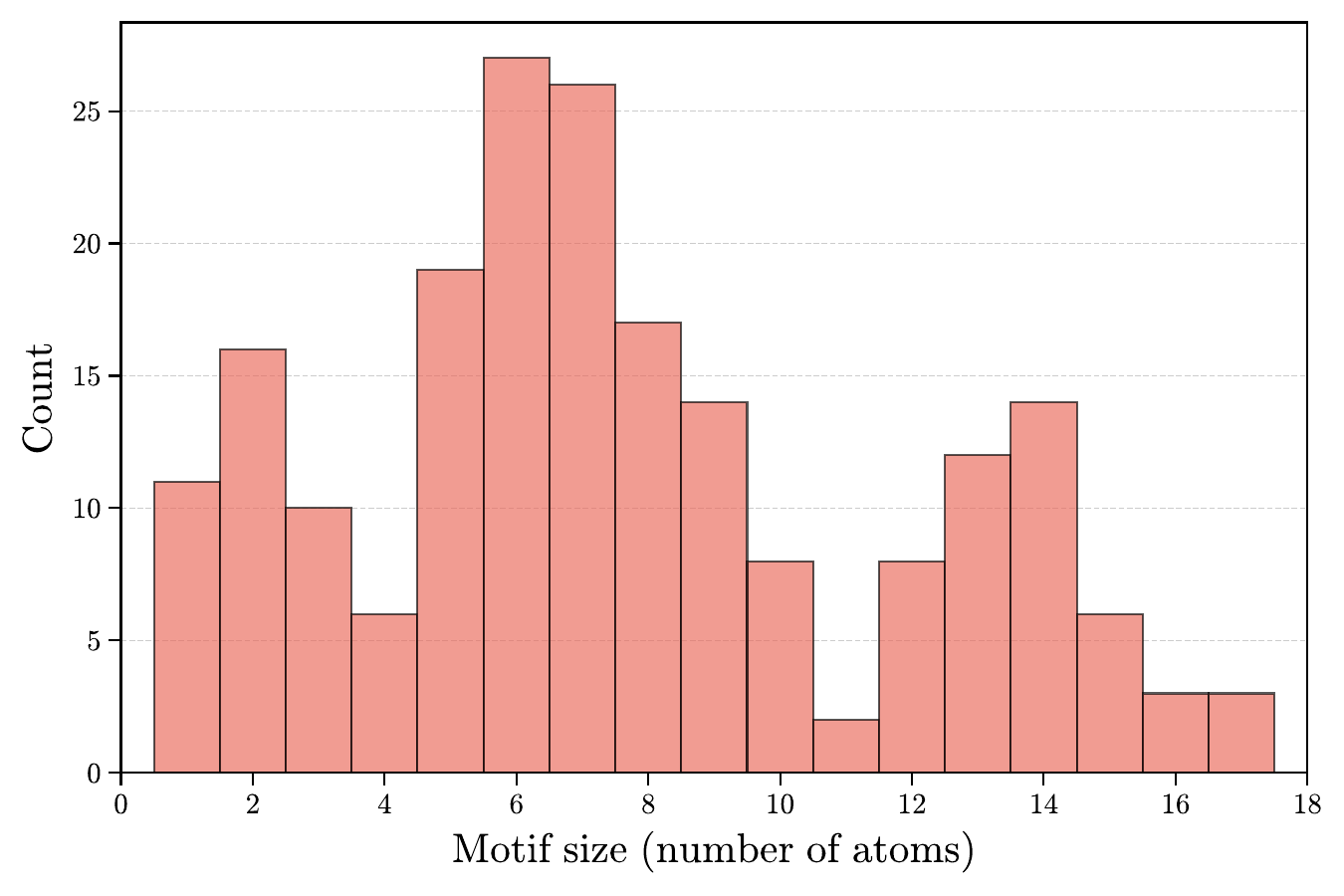} 
        \caption{Motif Sizes: Vocabulary-Based}
        \label{fig:size_vocab}
    \end{subfigure}\hfill
    \begin{subfigure}{0.48\textwidth}
        \centering
        \includegraphics[width=\linewidth]{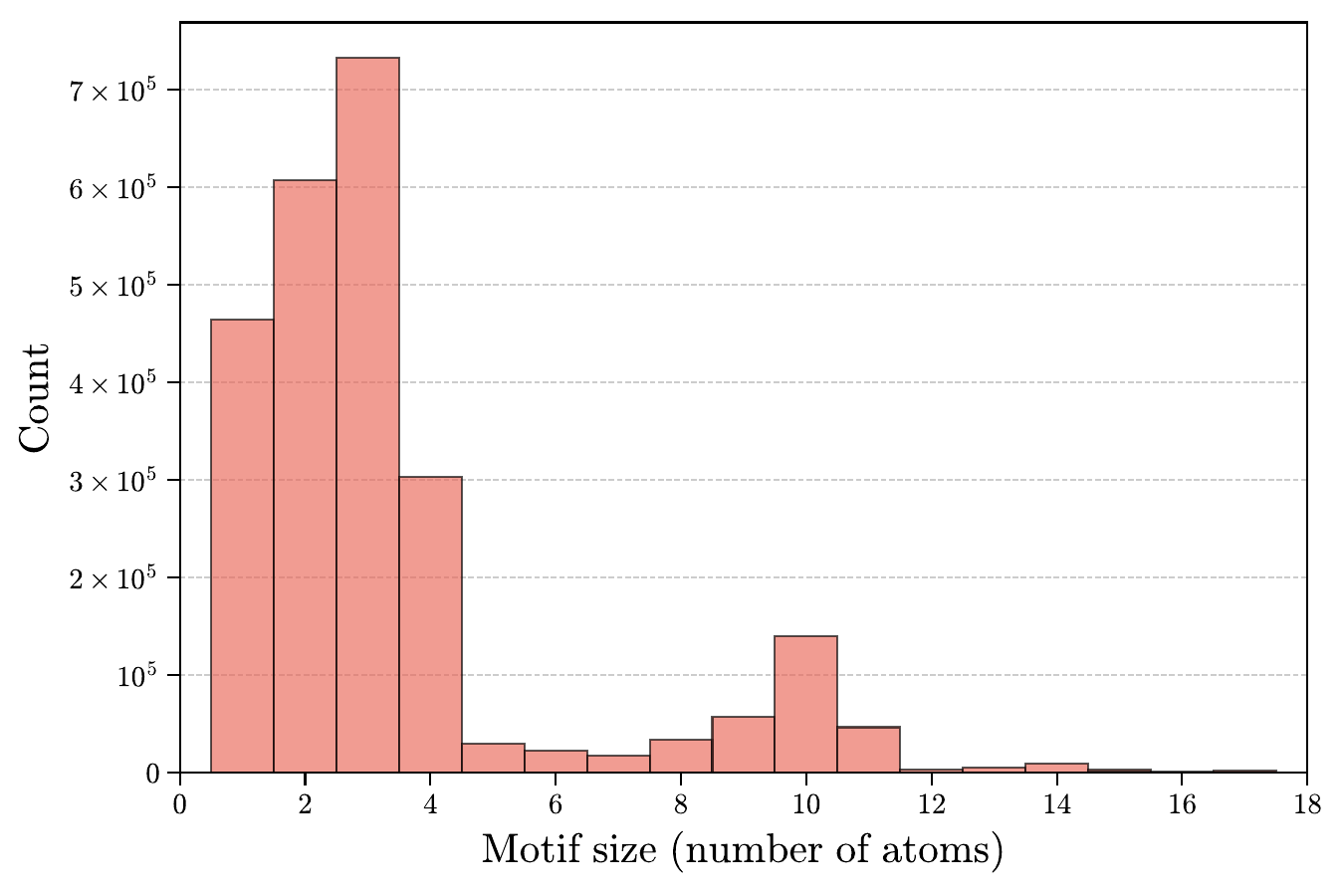} 
        \caption{Motif Sizes: Data-Based}
        \label{fig:size_data}
    \end{subfigure}

    \caption{Detailed statistics of the extracted rigid motifs for \textsc{GEOM-Drugs}. \textbf{Top row:} Histograms comparing the frequencies of motifs versus their rank for the default \textsc{Planar Rings} $0.1\%$ fragmentation (\textit{a}) and the fine-grained \textsc{No Rings} baseline (\textit{b}). \textbf{Bottom row:} Distributions of true chemical motif sizes (excluding dummy atoms) under the default $0.1\%$ strategy, counted uniquely per vocabulary class (\textit{c}) versus weighted by their actual frequency in the training data (\textit{d}).}
    \label{fig:vocab_histograms}
\end{figure}

Figure \ref{fig:vocab_histograms} compares the frequency profiles and sizes of the extracted fragments. Figure \ref{fig:freq_rank_01} shows that the \textsc{Planar Rings} $0.1\%$ strategy yields a long-tailed distribution spanning over $200$ unique classes. On the contrary, the \textsc{No Rings} baseline in Figure \ref{fig:freq_rank_norings} restricts the vocabulary to fewer than $40$ distinct components.

Finally, Figures \ref{fig:size_vocab} and \ref{fig:size_data} contrast the sizes of fragments present in the vocabulary with their actual frequency in the dataset. While the vocabulary itself (\textit{c}) contains many larger structures, peaking around 6-7 and 14 atoms for single and fused rings, respectively, the data-weighted count (\textit{d}) is heavily concentrated on small motifs. This reflects that individual molecules are predominantly composed of smaller functional groups and acyclic links connecting these larger rings.

\subsubsection{Architecture Ablations}
\label{Architecture Ablations}
We ablate the design choices used in our main method presented in Section \ref{Method} on the unconditional generation with \textsc{QM9}. The results are summarised in Table \ref{qm9_architecture_ablation}.

\textbf{Discrete Prior}: we first evaluate the choice of the prior distribution for the discrete flow. While our default method uses a masking prior $m_0 = [\text{MASK}]$, the \textsc{Uniform Prior} variant employs a probability path that interpolates from a uniform categorical distribution over the vocabulary $\mathcal{V}_m$ to the target class.

\textbf{Motif Symmetry Handling}: to validate our dynamic alignment strategy, we compare it against three alternative approaches for handling motif symmetries. First, we consider a naive baseline \textsc{No} $\mathcal{L}_{\textsc{GeoDiff}}$ that ignores symmetry, regressing directly towards the arbitrary orientation fixed during preprocessing. Second, in \textsc{With} $\mathcal{L}_{\textsc{AF3}}$, we test the alignment mechanism from \textsc{AlphaFold3} \citep{alphafold3}, which aligns the target frame to the model's prediction $\hat{\mathbf{R}}_1$ rather than the noisy input $\mathbf{R}_t$. Third, instead of resolving symmetries analytically in the loss, we employ data augmentation in the \textsc{With Augmentation} variant, where we randomly sample a symmetry element $\mathbf{S} \in \mathcal{S}_k$ from the motif's orbit at each training step and apply it to the target: $\tilde{\mathbf{R}}_1^k \leftarrow \mathbf{S}\mathbf{R}_1^k$.

\textbf{Backbone Components}: finally, we assess the contributions of specific architectural features. We evaluate the model performance without the triangle multiplicative updates in \textsc{No} $\Delta$-\textsc{Update}. Additionally, we train a variant with \textsc{No Self-Conditioning} to measure the impact of feeding the model's own discrete predictions of the clean motif types $\hat{m}_1$ back as input.

The results indicate that the most influential factor on the quality of unconditional generation is the removal of the triangle multiplicative update, confirming our hypothesis that it is useful for efficient modelling of unstructured molecular fragments. We did not observe a significant effect from applying data augmentation, since, strictly speaking, the target rotations are already chosen arbitrarily from equivalent orientations during preprocessing. Overall, our base method performs best, and thus reinforces the design choices outlined in Section \ref{Method}.

\begin{table}[ht]
\centering
\small
\setlength{\tabcolsep}{1.1pt}
\caption{Architectural ablations on \textsc{QM9}.}
\label{qm9_architecture_ablation}
\begin{tabular}{@{} l c c c c c c c c c c @{}}
\toprule
\multirow{2}{*}{Method} & \multirow{2}{*}{Steps} & \multicolumn{2}{c}{Stability ($\uparrow$)} & \multicolumn{2}{c}{Val. / Uniq. ($\uparrow$)} & \multicolumn{2}{c}{TV ($\downarrow$)} & \multicolumn{1}{c}{Str. En. ($\downarrow$)} & \multirow{2}{*}{SA ($\uparrow$)} & \multirow{2}{*}{QED ($\uparrow$)} \\ 
\cmidrule(l){3-8}
 & & A, \% & M, \% & V, \% & V $\times$ U, \% & A$\left[10^{-2}\right]$ & B$\left[10^{-3}\right]$ & $\Delta \text{E} \left[\frac{\text{kcal}}{\text{mol}}\right]$ & & \\
\midrule
\makecell[l]{Data} & -- & $99.0$ & $95.2$ & $97.7$ & $97.7$ & -- & -- & $7.7$ & $0.63$ & $0.46$ \\
\specialrule{0.7pt}{3pt}{3pt}

\multirow{2}{*}{\makecell[l]{\textsc{Uniform} \\ \textsc{Prior}}}
  & $50$   & $98.7_{\text{\tiny $\pm$.1}}$ & $90.8_{\text{\tiny $\pm$.1}}$ & $95.2_{\text{\tiny $\pm$.5}}$ & $89.2_{\text{\tiny $\pm$.5}}$ & $1.42_{\text{\tiny $\pm$.06}}$ & $12.2_{\text{\tiny $\pm$.6}}$ & $14.3_{\text{\tiny $\pm$.5}}$ & $0.70_{\text{\tiny $\pm$.00}}$ & $0.47_{\text{\tiny $\pm$.00}}$ \\
  & $100$  & $98.7_{\text{\tiny $\pm$.1}}$ & $90.4_{\text{\tiny $\pm$.4}}$ & $93.5_{\text{\tiny $\pm$.3}}$ & $87.5_{\text{\tiny $\pm$.4}}$ & $1.38_{\text{\tiny $\pm$.09}}$ & $15.2_{\text{\tiny $\pm$.2}}$ & $10.7_{\text{\tiny $\pm$.1}}$ & $0.70_{\text{\tiny $\pm$.00}}$ & $0.47_{\text{\tiny $\pm$.00}}$ \\
\midrule

\multirow{2}{*}{\makecell[l]{\textsc{No} $\mathcal{L}_{\textsc{GeoDiff}}$}}
  & $50$   & $98.6_{\text{\tiny $\pm$.0}}$ & $89.5_{\text{\tiny $\pm$.2}}$ & $95.5_{\text{\tiny $\pm$.1}}$ & $88.7_{\text{\tiny $\pm$.5}}$ & $2.35_{\text{\tiny $\pm$.05}}$ & $4.9_{\text{\tiny $\pm$.5}}$ & $16.5_{\text{\tiny $\pm$.2}}$ & $0.70_{\text{\tiny $\pm$.00}}$ & $0.47_{\text{\tiny $\pm$.00}}$ \\
  & $100$  & $98.9_{\text{\tiny $\pm$.0}}$ & $90.8_{\text{\tiny $\pm$.2}}$ & $94.5_{\text{\tiny $\pm$.1}}$ & $88.0_{\text{\tiny $\pm$.4}}$ & $2.35_{\text{\tiny $\pm$.08}}$ & $6.4_{\text{\tiny $\pm$.5}}$ & $11.8_{\text{\tiny $\pm$.1}}$ & $0.70_{\text{\tiny $\pm$.00}}$ & $0.47_{\text{\tiny $\pm$.00}}$ \\
\midrule

\multirow{2}{*}{\makecell[l]{\textsc{With} $\mathcal{L}_{\textsc{AF3}}$}}
  & $50$   & $98.5_{\text{\tiny $\pm$.0}}$ & $88.7_{\text{\tiny $\pm$.2}}$ & $93.8_{\text{\tiny $\pm$.2}}$ & $85.4_{\text{\tiny $\pm$.4}}$ & $2.81_{\text{\tiny $\pm$.04}}$ & $4.0_{\text{\tiny $\pm$.7}}$ & $16.6_{\text{\tiny $\pm$.6}}$ & $0.70_{\text{\tiny $\pm$.00}}$ & $0.47_{\text{\tiny $\pm$.00}}$ \\
  & $100$  & $98.6_{\text{\tiny $\pm$.1}}$ & $89.4_{\text{\tiny $\pm$.5}}$ & $92.4_{\text{\tiny $\pm$.5}}$ & $83.6_{\text{\tiny $\pm$.4}}$ & $2.87_{\text{\tiny $\pm$.05}}$ & $5.6_{\text{\tiny $\pm$1.5}}$ & $12.1_{\text{\tiny $\pm$.5}}$ & $0.71_{\text{\tiny $\pm$.00}}$ & $0.47_{\text{\tiny $\pm$.00}}$ \\
\midrule

\multirow{2}{*}{\makecell[l]{\textsc{With} \\ \textsc{Augmentation}}}
  & $50$   & $98.8_{\text{\tiny $\pm$.0}}$ & $90.8_{\text{\tiny $\pm$.3}}$ & $95.9_{\text{\tiny $\pm$.1}}$ & $87.9_{\text{\tiny $\pm$.5}}$ & $2.44_{\text{\tiny $\pm$.05}}$ & $4.6_{\text{\tiny $\pm$.8}}$ & $15.4_{\text{\tiny $\pm$.5}}$ & $0.70_{\text{\tiny $\pm$.00}}$ & $0.47_{\text{\tiny $\pm$.00}}$ \\
  & $100$  & $98.8_{\text{\tiny $\pm$.0}}$ & $90.2_{\text{\tiny $\pm$.3}}$ & $93.5_{\text{\tiny $\pm$.3}}$ & $85.0_{\text{\tiny $\pm$.4}}$ & $2.32_{\text{\tiny $\pm$.07}}$ & $8.4_{\text{\tiny $\pm$1.3}}$ & $11.3_{\text{\tiny $\pm$.3}}$ & $0.70_{\text{\tiny $\pm$.00}}$ & $0.47_{\text{\tiny $\pm$.00}}$ \\
\midrule

\multirow{2}{*}{\makecell[l]{\textsc{No} \\ $\Delta$-\textsc{Update}}}
  & $50$   & $98.3_{\text{\tiny $\pm$.1}}$ & $86.2_{\text{\tiny $\pm$.6}}$ & $93.4_{\text{\tiny $\pm$.4}}$ & $85.7_{\text{\tiny $\pm$.5}}$ & $2.28_{\text{\tiny $\pm$.08}}$ & $4.4_{\text{\tiny $\pm$.5}}$ & $16.9_{\text{\tiny $\pm$.5}}$ & $0.70_{\text{\tiny $\pm$.00}}$ & $0.47_{\text{\tiny $\pm$.00}}$ \\
  & $100$  & $98.2_{\text{\tiny $\pm$.1}}$ & $85.7_{\text{\tiny $\pm$1.4}}$ & $90.8_{\text{\tiny $\pm$1.4}}$ & $82.0_{\text{\tiny $\pm$1.3}}$ & $2.51_{\text{\tiny $\pm$.08}}$ & $6.8_{\text{\tiny $\pm$1.8}}$ & $13.3_{\text{\tiny $\pm$.1}}$ & $0.70_{\text{\tiny $\pm$.00}}$ & $0.47_{\text{\tiny $\pm$.00}}$ \\
\midrule

\multirow{2}{*}{\makecell[l]{\textsc{No Self-} \\ \textsc{Conditioning}}}
  & $50$   & $98.9_{\text{\tiny $\pm$.0}}$ & $91.0_{\text{\tiny $\pm$.3}}$ & $96.1_{\text{\tiny $\pm$.3}}$ & $88.3_{\text{\tiny $\pm$.4}}$ & $1.73_{\text{\tiny $\pm$.05}}$ & $4.7_{\text{\tiny $\pm$.4}}$ & $15.5_{\text{\tiny $\pm$.3}}$ & $0.70_{\text{\tiny $\pm$.00}}$ & $0.47_{\text{\tiny $\pm$.00}}$ \\
  & $100$  & $98.8_{\text{\tiny $\pm$.0}}$ & $90.6_{\text{\tiny $\pm$.3}}$ & $93.7_{\text{\tiny $\pm$.3}}$ & $85.2_{\text{\tiny $\pm$.5}}$ & $1.66_{\text{\tiny $\pm$.04}}$ & $7.4_{\text{\tiny $\pm$.1}}$ & $10.9_{\text{\tiny $\pm$.2}}$ & $0.70_{\text{\tiny $\pm$.00}}$ & $0.47_{\text{\tiny $\pm$.00}}$ \\
\midrule

\multirow{2}{*}{\makecell[l]{\oursacro}}
  & $50$   & $99.1_{\text{\tiny $\pm$.0}}$ & $92.3_{\text{\tiny $\pm$.3}}$ & $96.6_{\text{\tiny $\pm$.1}}$ & $88.4_{\text{\tiny $\pm$.2}}$ & $2.41_{\text{\tiny $\pm$.08}}$ & $2.2_{\text{\tiny $\pm$.3}}$ & $13.6_{\text{\tiny $\pm$.5}}$ & $0.70_{\text{\tiny $\pm$.00}}$ & $0.47_{\text{\tiny $\pm$.00}}$ \\
  & $100$  & $99.1_{\text{\tiny $\pm$.1}}$ & $92.6_{\text{\tiny $\pm$.5}}$ & $95.3_{\text{\tiny $\pm$.6}}$ & $86.3_{\text{\tiny $\pm$.9}}$ & $2.24_{\text{\tiny $\pm$.10}}$ & $2.7_{\text{\tiny $\pm$.4}}$ & $10.3_{\text{\tiny $\pm$.4}}$ & $0.70_{\text{\tiny $\pm$.00}}$ & $0.47_{\text{\tiny $\pm$.00}}$ \\
\bottomrule

\end{tabular}
\end{table}
\renewcommand{\arraystretch}{1.0}